\documentclass[runningheads]{llncs}
\usepackage[T1]{fontenc}
\usepackage{graphicx}
\usepackage{color}
\usepackage{booktabs}
\usepackage{subcaption}
\usepackage{amsmath}
\usepackage{amssymb}
\usepackage[pagebackref,breaklinks,colorlinks]{hyperref}

\usepackage{multirow}
\usepackage{cellspace}

\begin{document}

\title{AI-Generated Lecture Slides for Improving Slide Element Detection and Retrieval}
\titlerunning{SynLecSlideGen}

\author{
Suyash Maniyar\thanks{These authors contributed equally.}\inst{1}\orcidID{0009-0000-5882-4377} \and
Vishvesh Trivedi$^{*}$\inst{2}\orcidID{0009-0004-5043-6766} \and
Ajoy Mondal\inst{3}\orcidID{0000-0002-4808-8860} \and
Anand Mishra\inst{1}\orcidID{0000-0002-7806-2557} \and
C.~V.~Jawahar\inst{3}\orcidID{0000-0001-6767-7057}
}
\authorrunning{Maniyar et al.}

\institute{
Indian Institute of Technology, Jodhpur, India \\
\email{suyash.1@alumni.iitj.ac.in, mishra@iitj.ac.in}
\and 
Sardar Vallabhbhai National Institute of Technology, Surat, India \\
\email{u20cs130@coed.svnit.ac.in} 
\and
CVIT, International Institute of Information Technology, Hyderabad, India \\
\email{\{ajoy.mondal, jawahar\}@iiit.ac.in}
}

\maketitle

\begin{abstract}
Lecture slide element detection and retrieval are key problems in slide understanding. Training effective models for these tasks often depends on extensive manual annotation. However, annotating large volumes of lecture slides for supervised training is labor intensive and requires domain expertise. To address this, we propose a large language model (LLM)-guided synthetic lecture slide generation pipeline, SynLecSlideGen, which produces high-quality, coherent and realistic slides. We also create an evaluation benchmark, namely RealSlide by manually annotating 1,050 real lecture slides. To assess the utility of our synthetic slides, we perform few-shot transfer learning on real data using models pre-trained on them. Experimental results show that few-shot transfer learning with pretraining on synthetic slides significantly improves performance compared to training only on real data. This demonstrates that synthetic data can effectively compensate for limited labeled lecture slides. The code and resources of our work are publicly available on our project website:~\url{https://synslidegen.github.io/}.


\end{abstract}

\section{Introduction}\label{sec:introduction}

Presentations are a ubiquitous medium for information exchange across domains such as business, healthcare, law, engineering, and education. Automatically parsing lecture slides and extracting information enables a range of applications, such as automatic summarization, intelligent search, and AI-powered educational assistants. Despite their widespread applications, automating the understanding and analysis of slide content remains a significant challenge primarily due to the absence of a large-scale, densely annotated dataset. Current Vision-Language Models (VLMs) for document understanding~\cite{tang2023unifying,huang2022layoutlmv3,appalaraju2021docformer,da2023vision} are highly dependent on large-scale annotated datasets to adapt pre-trained representations to novel domains. For example, significant improvements in document layout analysis for scientific papers have been achieved by fine-tuning very-large labeled datasets such as PubLayNet~\cite{zhong2019publaynet} and DocBank~\cite{li2020docbank}. However, obtaining similar high-quality annotations for lecture slide images requires substantial manual labor because of their inherently complex layouts and high stylistic variability.

Although previous efforts in lecture slide understanding have achieved incremental progress in tasks such as \textit{Slide Image Segmentation}~\cite{haurilet2019spase,haurilet2019wise}, \textit{Slide Image Retrieval}~\cite{lee2022multimodal,jobin2024semantic}, and \textit{SlideVQA}~\cite{tanaka2023slidevqa}, the persistent challenge of manual data annotation remains a significant bottleneck. Inspired by the growing trend of leveraging synthetic data to enhance training across various domains, we introduce \textbf{SynLecSlideGen} --- an LLM-guided tool to generate coherent, and realistic synthetic lecture slide images and corresponding automatic annotations. We evaluated our generated slide dataset \textbf{SynSlide} on two primary tasks of slide understanding -- \textit{Slide Element Detection (SED)} and \textit{Text-based Slide Image Retrieval (TSIR)} on real lecture slides. To assess the effectiveness of the \textbf{SynSlide}, we conduct few-shot transfer learning on real slides using models pre-trained on synthetic slides. Results show that this approach outperforms training solely on real data, highlighting the value of synthetic slides as a pre-training resource in low-labeled data scenarios. Specifically, pre-training on synthetic slides before fine-tuning with just 50 real images enhances SED performance, achieving a 9.7\% mAP boost on YOLOv9, with significant improvements in low-resource classes such as code snippets (+32.5\% mAP) and natural images (+20.2\% mAP). In the TSIR task, synthetic slide images improved R@1 by 3\% with the CLIP model, showcasing its effectiveness in handling high intraclass variance and rare slide elements. These experiments highlight the potential of synthetic data to enhance performance in low-resource slide image scenarios, providing a scalable solution to understanding lecture presentations.

We summarize our contributions as follows. 

\begin{itemize}
\item \textbf{SynLecSlideGen:} An LLM-based open-source pipeline for generating realistic, coherent, copyright-free lecture slides with automatic annotations, used to create the SynSlide dataset for slide understanding.
\item \textbf{Benchmark Dataset:} A curated RealSlide dataset of 1,050 real lecture slides with manual annotations for slide element detection and query-based retrieval.
\item \textbf{Comprehensive Evaluation:} We conduct extensive experiments to assess the performance of synthetic slides on two main tasks related to lecture-slide understanding, namely slide element detection and query-based slide image retrieval. Experiments show that pre-training on synthetic slides improves few-shot transfer learning on real slides, highlighting their value in low-labeled data settings. The code and resources of our work are publicly available on our project website:~\url{https://synslidegen.github.io/}.

\end{itemize}

\section{Related Work} \label{sec:related-work}

\subsection{Slide Image Understanding}

Slide image understanding is a multidisciplinary field that includes tasks such as slide segmentation, narration, and multi-modal retrieval.

\paragraph{\textbf{Slide Segmentation \& Narration:}}

SPaSe~\cite{haurilet2019spase} provides a multi-label slide segmentation data set with 2,000 pixel-annotated slides from SlideShare1M~\cite{araujo2016large}, addressing overlapping classes, fine-grained labels and supporting multilingual content. WiSe~\cite{haurilet2019wise}, the first wild slide segmentation dataset, includes 1,300 annotated slides reflecting real-world conditions such as noise and lighting. CSNS~\cite{jobin2021classroom} aids visually impaired students by segmenting and narrating slide content in reading order.

While most works focus on pixel-level segmentation, limiting semantic understanding, few explore object-level detection~\cite{peng2024dreamstruct}. These, however, lack benchmarks for open-world, densely structured lecture slides, instead targeting simpler talking presentations.

\paragraph{\textbf{Multi-modal Retrieval:}}

The Lecture Presentations Multi-modal Dataset~\cite{lee2022multimodal} includes 9,000 slide images from 180 hours of lecture videos, annotated with figure bounding boxes and aligned audio transcripts. It introduced \textit{figure-to-text} and \textit{text-to-figure} cross-modal retrieval tasks and PolyViLT, a multimodal transformer trained with multi-instance learning loss, surpassing VLMs like CLIP~\cite{radford2021learning} and PCME~\cite{song2019polysemous} in retrieval tasks. Similarly, Jobin \emph{et al.}~\cite{jobin2024semantic} proposed a semantic labels-aware transformer model for lecture slide retrieval, enabling searches based on natural language and sketch. The proposed model trained on the LecSD dataset of 50,000 annotated slides outperforms existing methods and sets a new benchmark in slide retrieval.

\subsection{Synthetic Data Generation}
Synthetic data generation in computer vision began with statistical methods and evolved through generative modeling to address data scarcity~\cite{sufi2024addressing}, privacy~\cite{osuala2024enhancing}, robustness~\cite{singh2024synthetic}, and domain adaptation~\cite{Shakeri2020}. In document understanding, it enables the creation of articles~\cite{shao2024assisting}, layouts~\cite{he2023diffusion}, handwritten notes~\cite{blanes2018synthetic}, and full documents like scientific papers~\cite{pisaneschi2023automatic} and news reports~\cite{shu2021fact}. Peng~\emph{et al.}~\cite{peng2024dreamstruct} introduced synthetic user interface (UI) and slide generation pipelines for tasks like image captioning and element recognition, assessing slides with FitVid~\cite{kim2022fitvid}, a dataset of screen-captured video presentations. Likewise, Seng \emph{et al.}~\cite{seng2024slidecraft} developed a pipeline to generate lecture slides from Wikipedia, demonstrating that mixed training with both real and synthetic slides enhances performance on FitVid~\cite{kim2022fitvid} and SlideVQA~\cite{tanaka2023slidevqa} for document object detection.

Despite progress in synthetic data generation, fidelity and diversity gaps between synthetic and real slides remain underexplored, often leading to diminishing returns at scale~\cite{seng2024slidecraft,yu2023diversify}. Furthermore, the lack of open, multi-task benchmarks limits the ability to generalize across tasks, with models frequently failing when applied beyond their training objective~\cite{peng2024dreamstruct}.

\section{Synthetic Lecture Slide Generation Approach} \label{sec_ssga}

\begin{figure}[!t]
\centerline{  
\includegraphics[width=1.0\textwidth]{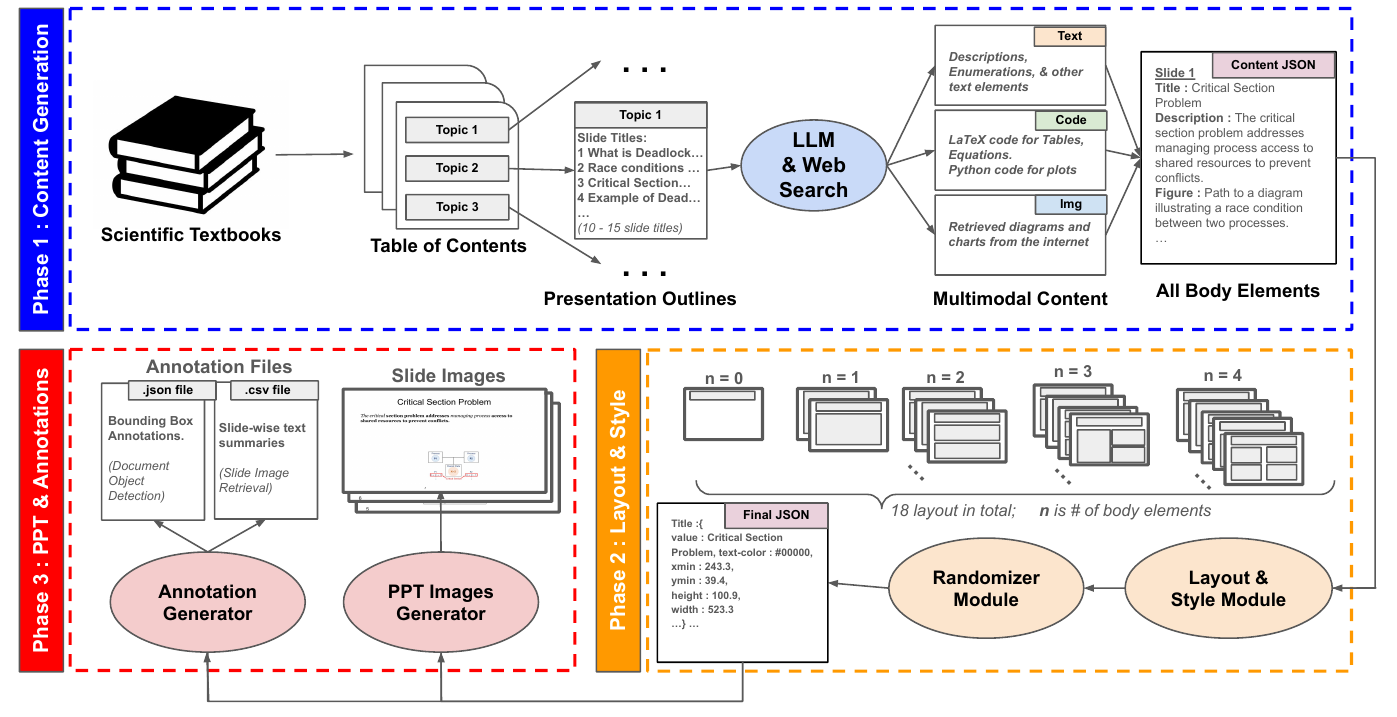}}
\caption{Overview of our Synthetic Lecture Slide Generation \textbf{SynLecSlideGen} pipeline. In Phase I, we generate presentation content using LLMs and Web Image search agents in a multi-step process. In Phase II, we arbitrarily assign layout and style based on each slide's content. Finally, in Phase III, we generate the PPT files, convert them to images, and also generate automatic annotations for multiple downstream tasks from the final JSON file. }\label{fig:generation_phase} 
\end{figure}

In order to reduce the reliance on manual annotation required for training lecture slide-related tasks, we propose a large language model (LLM)-guided Synthetic Lecture Slide Generation pipeline (\textbf{SynLecSlideGen}) consisting of the following three phases. In \textbf{Phase I: Content Generation}, we generate semantically rich multi-modal content using LLM and Web image search. In \textbf{Phase II: Layout and Style Assignment}, a layout and style discriminator aligns layouts with content types, producing a JSON file that includes all content and layout details to generate a presentation. Finally, in \textbf{Phase III: Slide Generation and Annotation}, we create PowerPoint presentation (PPT) files from JSON and convert them to images with automatically generated annotations, eliminating manual effort. Fig.~\ref{fig:generation_phase} illustrates our three-phase process for high visual quality and coherent lecture slide generation.     

\subsection{Phase I: Content Generation for Synthetic Lecture Slides}
The phase evolves in the following five steps.

\paragraph{\textbf{Topic Generation:}}
We create a corpus of lecture topics using indexes from openly available textbooks. We select well-known \textit{STEM} textbooks in \textit{Computer Science, Mathematics, Physics, and Economics} such as \textit{Introduction to Algorithms by Thomas H Cormen, Macroeconomics by Greg Mankiw}. Since large foundational models have been pre-trained on open scientific books, this approach minimizes dependence on local content retrieval as we rely on the model's parametric knowledge~\cite{brown2020language}. For each textbook, an LLM-based topic generation module produces a list of lecture topics based on the index of contents. For example, \textit{Divide-and-Conquer Approach, Growth Theory in Macroeconomics}. These topics are then used as seeds to generate an entire presentation on each topic in the subsequent steps. We provide a one-shot example using one textbook and its expected topics for each of the four domains before generation. On average, the LLM provides between 10 and 15 topics per textbook, providing the seeds for as many presentations as possible.

\paragraph{\textbf{Outline Generation:}} 
Given each topic of the lecture, we prompt the LLM to generate an outline of the presentation, which provides the topic of discussion for each slide of the lecture. The input to this module is the lecture topic (e.g., \textit{Divide-and-Conquer Approach}), and the expected output is a list describing the topic of discussion for each slide (e.g., \textit{Introduction to Divide and Conquer, Motivation behind Divide and Conquer, Pseudo-code for Divide-and-Conquer algorithms}). We utilize three topic-outline pairs from real lectures to elicit in-context learning using few-shot examples. To limit redundancy, we limit the LLM to generate up to 15 slide suggestions per topic. On average, the LLM generates up to 12 to 15 discussion topics per lecture topic, and these are used as \textit{Titles} for each slide in the presentation.

\paragraph{\textbf{Element Type Generation:}} 

This module determines what elements are suitable for each slide based on the title. We use the chain-of-thought~\cite{wei2022chain} prompting technique to instruct the LLM first to determine the type of lecture slide according to the slide title and then suggest the type of elements based on the slide type. Following previous work~\cite{li2019should}, we consider six types of lecture slides -- \textit{Introduction, Definition, Example, Comparison, Conclusion, and References}. Intuitively, an \textit{Introduction} slide may contain text and/or enumeration with an optional figure. A \textit{Comparison} slide should ideally have a table for comparison, etc. We also provide examples of three real lecture slides of \{title, type, elements\} tuples for better suggestions. Hence, the input to this module is the slide title, and the output is a list of tuples where each tuple is \{element, caption\}. For instance, an example slide in the Divide-and-Conquer approach will have `code', `a pseudocode example of Divide and Conquer' as one of the suggestions. As is generally observed in real slides, we ask the LLM to generate up to three body element suggestions per slide. The output is divided into four prompt types: \textit{text} generation, \textit{figure} retrieval, \textit{structure} generation, and \textit{plot} generation. These prompts are provided as separate chains to the LLM to maximize the generated output tokens.

\paragraph{\textbf{Textual Element Generation:}} 

The text generation prompt includes \textit{ description and enumeration} elements where the LLM outputs the text content of each, based on the caption provided by the previous step.

\paragraph{\textbf{Structural Elements Generation and Diagram Retrieval:}}

Structural elements like \textit{Tables, Equations} and plot elements like \textit{Charts} are generated using \textit{LaTeX} and \textit{Python} code, respectively. The code is first stored, then compiled to generate corresponding graphics. Other visual elements like \textit{Diagrams} are challenging to generate using LLMs. Hence, they are retrieved from the Internet by using the \textit{Bing Search API}\footnote{\url{https://www.microsoft.com/en-us/bing/apis/bing-web-search-api}} based on the element caption as a search query. We fetch the top two related diagrams, with one selected randomly for inserting in the slide.

The output of steps 4 and 5 is compiled into a JSON file containing text content and image file hyperlinks for presentation. We use a combination of the GPT-3.5-Turbo and GPT-4 base APIs for the generation of slide content\footnote{Complete details of the LLMs used at each stage, along with the specific prompts, are provided in Appendix A of the supplementary material.}. The top part of Fig.~\ref{fig:generation_phase} illustrates Phase I of the pipeline. 

\subsection{Phase II: Layout and Style Assignment}

This phase assigns slide-level layout and styling properties for each slide in the presentation. It consists of a Python module that assigns one of 18 predefined slide layouts based on the size and count of body elements. We utilize 9 out of the 11 layouts available in the python-pptx library\footnote{\url{https://python-pptx.readthedocs.io/en/latest/dev/analysis/sld-layout.html}} such as \textit{two-column}, \textit{one-row-two-column}, etc. and generate one counterpart for each where \textit{Title} element is absent. In addition, we arbitrarily assign style attributes such as background color, font styling for text, design templates, shadows, and borders. Meta elements, e.g., \textit{slide numbers, footers, natural images, graphics, and logos} are randomly inserted to mimic real slides. The styling of certain elements \textit{ (e.g., font styling for title)} is kept consistent throughout the presentation while allowing slide-wise variations for other elements \textit{(e.g., description text color)}. To introduce variability, we implement two perturbation functions by (i) applying Gaussian noise to slightly shift element positions within layout boundaries, and (ii) modifying elements by their styling. The resulting JSON file now contains all the content, style, and layout information required to generate the presentation. We also use this file to automatically extract the annotations described in the next stage. The middle part of Fig.~\ref{fig:generation_phase} illustrates the steps of Phase II.

\subsection{Phase III: Lecture Slide Generation and Obtaining Automatic Annotation}

This phase constructs the PPT file and generates annotations for downstream tasks using the JSON file obtained from the previous phase. We utilize \textit{python-pptx library}\footnote{\url{https://pypi.org/project/python-pptx/}} that enables generating \textit{".PPT"} files from structured data like XML or JSON. The PPT file is then converted into a sequence of slide images. We use data augmentation techniques, like Gaussian blurs, pixelation, and resizing, to increase the variability of the obtained images. We utilize the data from the JSON file described below to obtain annotations.

\paragraph{\textbf{Slide Element Detection:}} We utilize the bounding-box coordinates for each slide element (e.g., title, table, figure, etc.) from the JSON file and their labels to construct a ground truth file compatible with the COCO~\cite{lin2014microsoft} object detection annotation format for the slide element detection tasks.

\paragraph{\textbf{Slide Image Retrieval:}} We extract content details (e.g., titles, descriptions, enumerations), layout attributes (e.g., element sizes), spatial positions (e.g., top of the slide), and meta elements (e.g., footers, URLs) from the JSON file to generate slide-level summaries for the slide image retrieval task. The output is a CSV file containing image IDs and their corresponding summaries.

The final dataset consists of synthetically generated slide images and annotation files. The leftmost part of the bottom row of Fig.~\ref{fig:generation_phase} illustrates the final phase of the pipeline.

\section{Dataset} 

\begin{figure}[!t]
\centerline{
\fbox{\includegraphics[height=0.12\textwidth,width=0.225\textwidth]{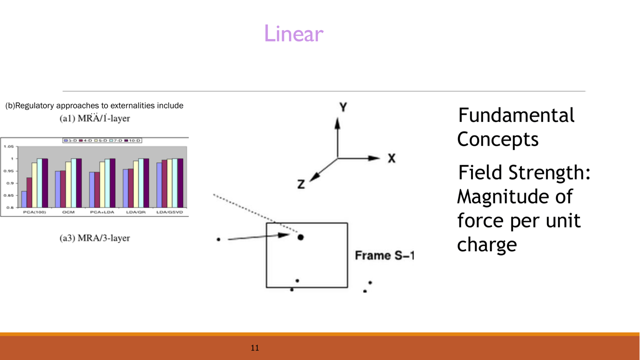}}
\hspace{-0.01\textwidth}
\fbox{\includegraphics[height=0.12\textwidth,width=0.225\textwidth]{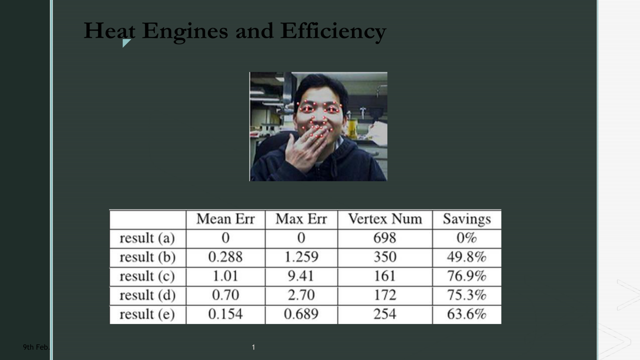}}
\hspace{-0.01\textwidth}
\fbox{\includegraphics[height=0.12\textwidth,width=0.225\textwidth]{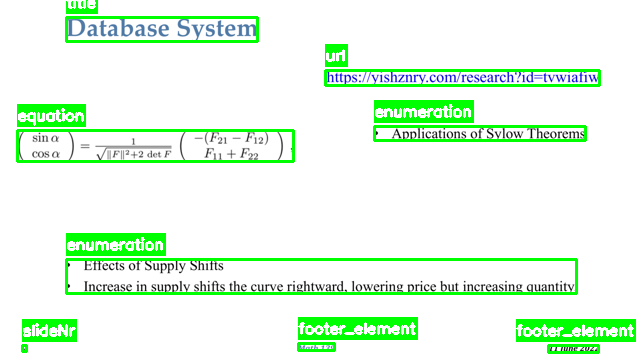}}
\hspace{-0.01\textwidth}
\fbox{\includegraphics[height=0.12\textwidth,width=0.225\textwidth]{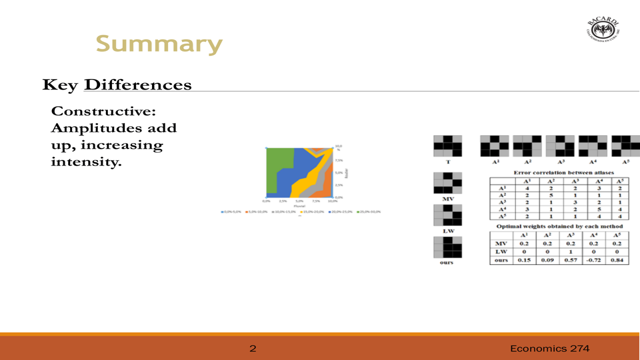}}}
\vspace{0.005\textwidth}
\centerline{
\includegraphics[width=1.015\textwidth]{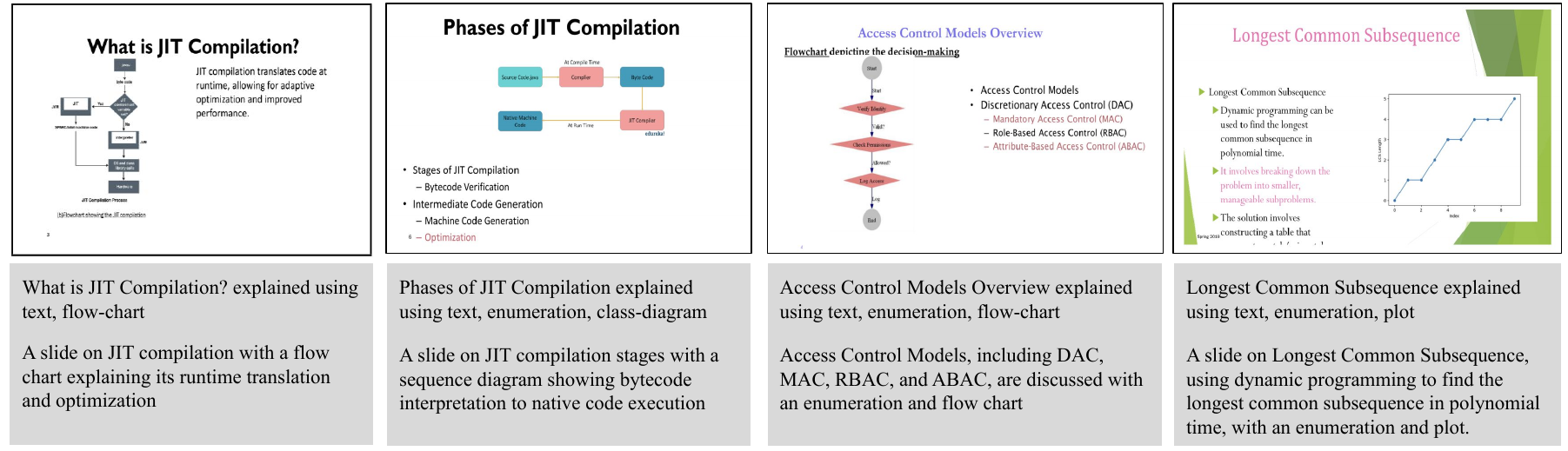}}
\caption{Displays sample slides from our generated \textbf{SynSlide} dataset. The first row presents examples from \textbf{SynDet}, with and without automatic bounding box annotations for elements. The second row showcases coherent slides from \textbf{SynRet}. The third row shows two types of summaries of corresponding slides.}\label{fig:sample}
\end{figure}

\subsection{SynSlide}

Our synthetic slide dataset, \textbf{SynSlide} is created using our pipeline explained in Section~\ref{sec_ssga}. It comprises two subsets: \textbf{SynDet} for document layout analysis and \textbf{SynRet} for text-based slide retrieval. \textbf{SynDet} includes 2,200 high quality slide images with automatic bounding box annotations for 16 element categories namely \textit{(Title, Description, Enumeration, SlideNr, Equation, Table, Logo, Heading, Diagram, Chart, Footer-Element, Code, Figure-Caption, Table-Caption, URL, Natural-Image)}. \textbf{SynRet} contains 2,200 topic coherent slide images with two types of slide summaries for each slide. Both of the synthetic sets have been constructed using the same generation pipeline with some task-specific processing steps as defined in Fig.~\ref{fig:generation_phase}. A few examples from \textbf{SynDet} and \textbf{SynRet} are presented in Fig.~\ref{fig:sample}. 

Most slide image retrieval research~\cite{jobin2024semantic,peng2024dreamstruct} focuses on retrieving images based on slide transcripts. LecSD~\cite{jobin2024semantic} enhances retrieval capabilities by supporting both text and sketch-based queries related to slide content and visuals. We use structured content files generated from synthetic slide creation to produce two types of summaries. The first type follows the \textit{LecSD-style} query format, where slide titles or enumerations are explained using key elements. This format, created via OCR, is sensitive to variations in query phrasing. To overcome this limitation, we introduce an OCR-free semantic summaries generated using a large language model (LLM) that incorporates slide content, type of slide elements, and metadata \textit{like name of instructor in footer, etc.}, thereby making slide retrieval more robust.

\begin{figure}[!ht]
\centerline{
\includegraphics[height=0.3\textwidth,width=1\textwidth]{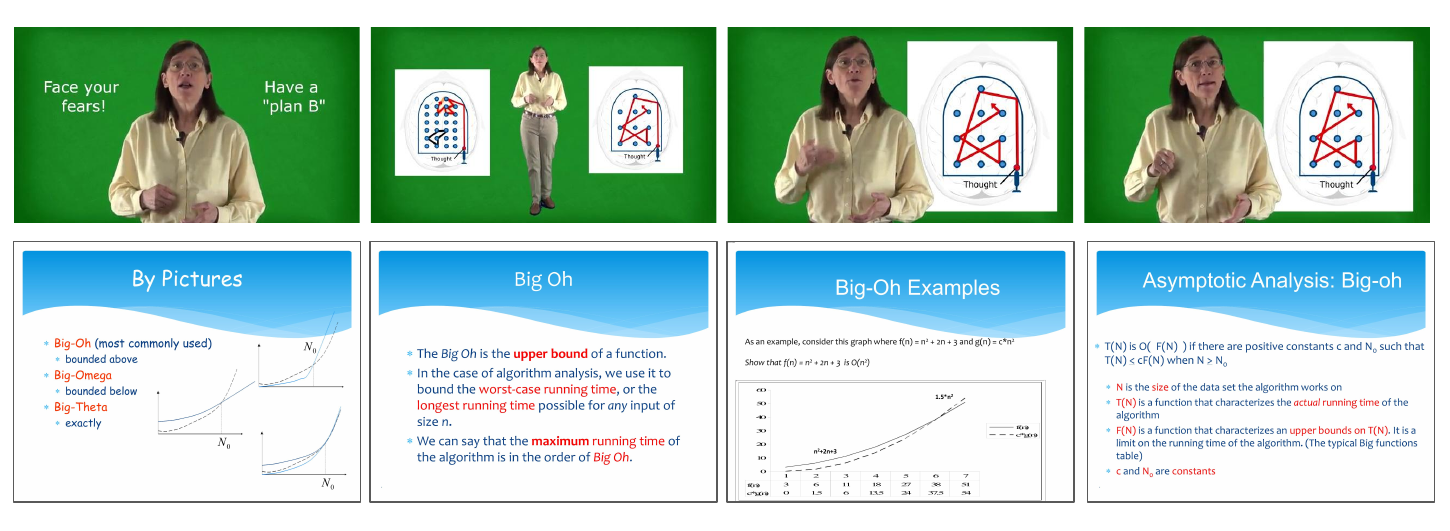}}
\caption{Comparison of four consecutive slides from lecture slide datasets. The top row displays slide images from the existing FitVid~\cite{kim2022fitvid} dataset, while the bottom row presents slide images from our \textbf{RealSlide}.}\label{fig:real_slide} 
\end{figure}

\subsection{RealSlide}

\begin{figure}[!t]
\centerline{
\includegraphics[height=0.35\textwidth,width=1\textwidth]{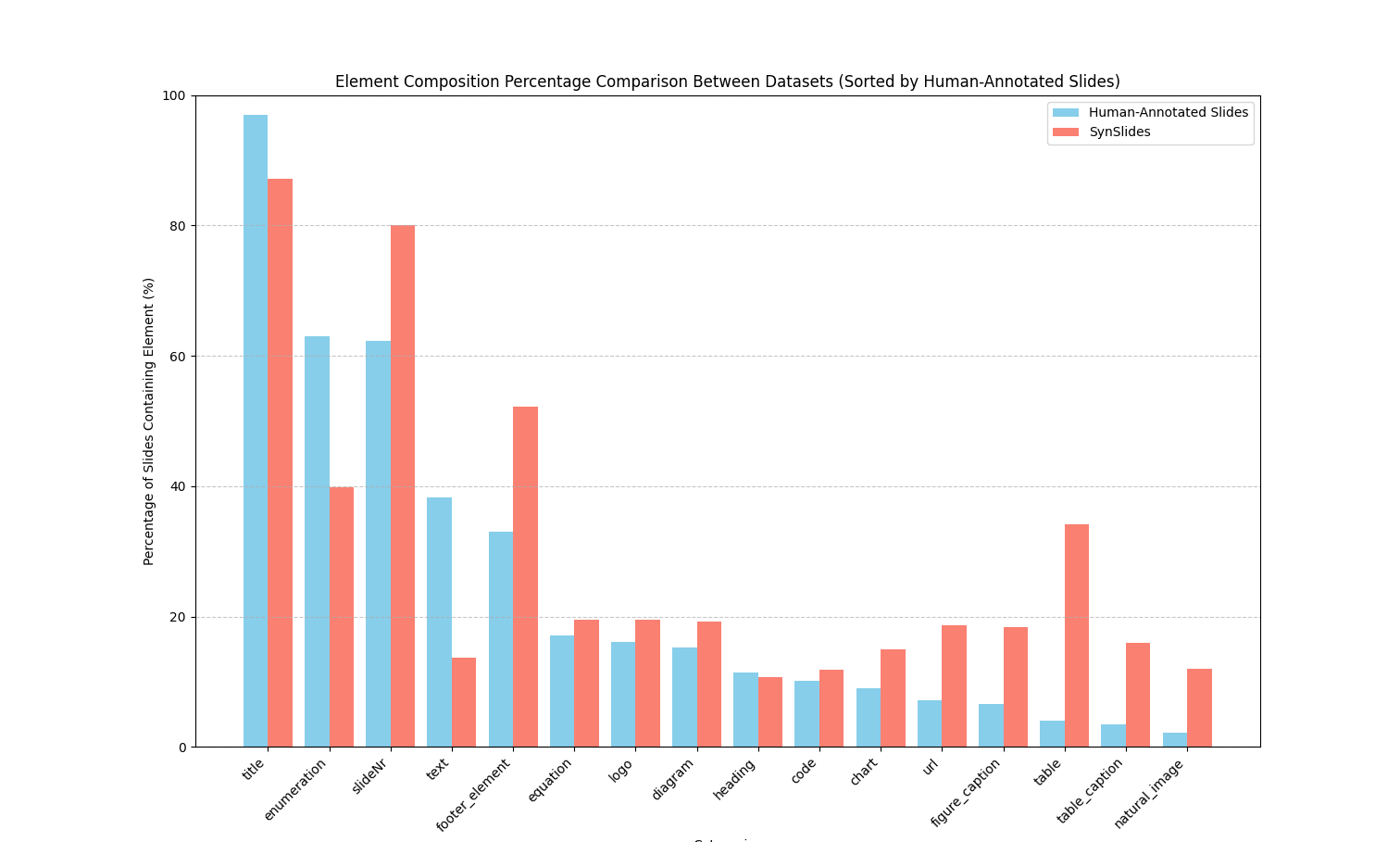}}
\caption{Occurrence of each element in the \textbf{SynDet} (Red) and \textbf{RealSlide} (Blue) datasets. \textbf{SynDet} contains twice as many elements for the five least frequent classes in \textbf{RealSlide}: URL, Figure Caption, Table, Table Caption, and Natural Image.}\label{fig:dataset_stat} 
\end{figure}

\begin{table}[!t]
\centering
\caption{Comparison between \textbf{RealSlide} and \textbf{SynSlide}. Abbreviations: Des. = Description, Ele. = Element, Nat. = Natural, Img. = Image, Cap. = Caption.}\label{tab:data_stat}
\begin{subtable}{0.45\textwidth}
\centering
\begin{tabular}{||l|r|r||}\hline
\textbf{Element} & \textbf{RealS.} & \textbf{SynS.}\\ \hline
\#Slide  &1050  &4400 \\
\#Des. Ele.  &1109 &3518 \\
\#Figures &484 &3862 \\
\#Meta Ele.  &1010 &6640 \\ \hline
Avg. Ele. per Slide &4.77 &5.82 \\ \hline
\end{tabular}
\caption{Key statistics between \textbf{RealSlide} and \textbf{SynSlide} (SynRet + SynDet)}\label{tab:table1_a}
\end{subtable}%
\hfill
\begin{subtable}{0.45\textwidth}
\centering
\begin{tabular}{||l|r|r||}\hline
\textbf{Element} &\textbf{RealS.} &\textbf{SynS.} \\ \hline
Nat. Img. & 33 & 700 \\
Table Cap. &  47 & 602  \\
URL & 61 & 892  \\
Table & 69 & 2012  \\
Heading & 106 & 834 \\ \hline
\end{tabular}
\caption{Comparison of top five scarce elements in \textbf{RealSlide} and \textbf{SynSlide}.}\label{tab:table1_b}
\end{subtable}
\end{table}

Existing benchmarks for slide object detection either provide pixel-level annotations~\cite{haurilet2019spase,haurilet2019wise}, originate from talking presentations~\cite{kim2022fitvid}, or lack detailed semantic annotations --- for instance, grouping multiple bullet points into a single enumeration block or failing to distinguish between distinct figures~\cite{haurilet2019spase,haurilet2019wise,tanaka2023slidevqa}. Moreover, current real datasets only provide annotations for a single task which limits evaluating document models on multiple tasks. Because of this, a new real benchmark for evaluation is needed. For this purpose, we curate a new benchmark, \textbf{RealSlide}, with 1050 human-annotated pre-made lecture slides for detection and retrieval tasks. These slides are curated using real lecture presentations\footnote{Publicly available lecture slides distributed under Creative Commons (CC) license. Links to all presentations are listed in supplementary material.} delivered for university-level courses in the fields of Computer Science (50\%), Economics (20\%), Physics (15\%), and Mathematics (15\%). Fig.~\ref{fig:real_slide} compares samples from an existing lecture slide dataset~\cite{kim2022fitvid} and our \textbf{RealSlide}. \textbf{RealSlide} is used as a validation set to evaluate the performance of \textbf{SynSlide} on two downstream tasks. The dataset is divided into a training set of 300 slides (30\%) for fine-tuning and a validation set of 750 slides (70\%) for evaluation. To prevent data leakage across different sets, we systematically partition the data to ensure that slides from the same presentation are exclusive to training or validation sets. Fig.~\ref{fig:dataset_stat} illustrates the percentage of element occurrences in \textbf{RealSlide} and \textbf{SynSlide}, while Table~\ref{tab:data_stat} compares element-wise statistics between \textbf{RealSlide} and \textbf{SynSlide}

\begin{table*}[!ht]
\caption{Comparison of lecture datasets with various features and availability. (M) indicates manually annotated data, while (A) indicates automatically annotated data. * after pruning classes like handwritten text, maps, etc. $^{\dag}$ only images are available.  }\label{tab:lecture_datasets}
\centering
\begin{tabular}{l|c|c|c|c|c|c|c|c}
\toprule
& \multicolumn{2}{c|}{\textbf{Features}} & \multicolumn{2}{c|}{\textbf{Annotations}} & \multicolumn{2}{c|}{\textbf{Size}} & \textbf{Avail.} \\
\cline{2-8}
& Multi-page & Semantic & Layout & Summary &\#BBox Class & \#Slide &  \\
\midrule
\multicolumn{8}{l}{\textit{Real Lecture Slide Datasets}} \\
\midrule
SPaSe~\cite{haurilet2019spase} & & & \checkmark (M) & & 18* & 2000 & \checkmark \\
WiSe~\cite{haurilet2019wise} & & &
\checkmark (M) & & 18* & 1300 & \checkmark \\
FitVid~\cite{kim2022fitvid} & \checkmark & & \checkmark (A) &  & 12 & 5500 & \checkmark \\
LecSD~\cite{jobin2024semantic} & &  &  & \checkmark (M/A) & - & 54000 &  \checkmark \\
MLP~\cite{lee2022multimodal} & \checkmark & & \checkmark (M/A)* & \checkmark (M/A) & 6 & 9031 &\checkmark  \\
RealSlide (Ours) & \checkmark  & \checkmark & \checkmark (M) &  \checkmark (M) & 16 & 1050 & \checkmark \\
\midrule
\multicolumn{7}{l}{\textit{Synthetic Lecture Slide Datasets}} \\
\midrule
DreamStruct~\cite{peng2024dreamstruct} & & & \checkmark (A) & \checkmark (A) & 12 & 10053 &\checkmark$^{\dag}$ \\
SlideCraft ~\cite{seng2024slidecraft} & \checkmark & \checkmark & \checkmark (A) & & 12 & 25000 & \\
SynSlide (Ours) & \checkmark & \checkmark & \checkmark (A) & \checkmark (A) & 16 & 4400 & \checkmark \\
\bottomrule
\end{tabular}
\end{table*}

\subsection{Comparison with Existing Lecture Slide Datasets}

We compare the generated \textbf{SynSlide} and evaluation set \textbf{RealSlide} with existing lecture slide datasets, focusing on their applicability to real-world tasks such as slide narration and real-time slide generation. Table~\ref{tab:lecture_datasets} compares current real and synthetic benchmarks against our proposed datasets. To assess the quality of our generated \textbf{SynSlide} in comparison to real slides and other synthetically produced slides, such as Dreamstruct, we utilize the Fréchet Inception Distance (FID)~\cite{fid}. The FID score is a technique to quantify the distance between two distributions. A lower score indicates a higher resemblance and vice versa.  Table~\ref{tab:fid_scores} compares the FID scores of two synthetic datasets, DreamStruct\cite{peng2024dreamstruct} and SynSlide against RealSlide. We also compare two equally-sized, presentation-wise split sets of RealSlide as control. The result indicates that SynSlide is closer in distribution to the real benchmark than recently proposed DreamStruct dataset.  

\begin{table}[!ht]
\caption{FID scores for several datasets. }
\label{tab:fid_scores}
\centering
\begin{tabular}{l c c}
\hline
Dataset 1 & Dataset 2  & FID Score $\downarrow$ \\
\hline
RealSlide  &RealSlide & 18.4\\
SynSlide   &RealSlide & 42.5 \\
Dreamstruct &RealSlide & 56.1\\\hline
\end{tabular}
\end{table}

\section{Experiments}

\subsection{Slide Element Detection}

\paragraph{\textbf{Baselines and Implementation Details:}}

We employ three popular models for slide element detection: LayoutLMV3~\cite{huang2022layoutlmv3}, YOLOv9~\cite{wang2024yolov9}, and DETR~\cite{carion2020}. For each of the models, we follow two different fine-tuning strategies --- (i) \textbf{Single Stage (SS):} pre-trained model is fine-tuned with the training set of \textbf{RealSlide} dataset and (ii) \textbf{Two Stage (TS):} pre-trained model is fine-tuned with our synthetic \textbf{SynDet} and again fine-tuned with the training set of \textbf{RealSlide} dataset to adapt real-world slide layouts. For the LayoutLMV3 and YOLOv9 models, we initialize the weights using the pre-trained models from PubLayNet~\cite{zhong2019publaynet} and DocLayNet~\cite{pfitzmann2022doclaynet}, respectively. For the DETR model, we utilize the publicly available COCO pretrained checkpoint . We set the batch size to 16 for all models and follow the standard settings specific to each model keeping all parameters trainable. Additionally, we resize all images to 360$\times$640 pixels and do not apply any image augmentations.

\begin{table}[!ht]
\centering
\caption{Impact of the IoU threshold on slide element detection performance (mAP) for three baseline models under two fine-tuning strategies: (i) \textbf{Single Stage} (\textbf{SS}) and (ii) \textbf{Two Stage} (\textbf{TS}) on the test set of the \textbf{RealSlide} dataset.}\label{tab:iou_ablation}
\begin{tabular}{||l|r r||r r||r r||}\hline
\textbf{IoU} &\multicolumn{2}{c||}{\textbf{YOLOv9}} &\multicolumn{2}{c||}{\textbf{LayoutLMV3}} &\multicolumn{2}{c||}{\textbf{  DETR  }} \\ \cline{2-7}
 &\textbf{SS} &\textbf{TS} &\textbf{SS} &\textbf{TS} &\textbf{SS} &\textbf{TS} \\ \hline
mAP@[0.50]  & 50.3 & 53.4    &38.9 &49.0& 36.6& 40.9\\
mAP@[0.55]  & 49.2 & 51.8   &37.6 &47.9& 33.4& 39.1\\
mAP@[0.60]  & 48.0 & 50.1   &36.2 &46.5& 30.9& 37.7\\
mAP@[0.65]  & 46.2 & 48.4   &34.0 &42.0& 26.1& 35.6\\
mAP@[0.70]  & 43.6 & 46.0   &30.8 &39.2& 22.4& 30.4\\
mAP@[0.75]  & 39.9 & 42.5  &27.2 &33.9& 18.5&
26.9\\
mAP@[0.80]  & 34.6 & 36.3  &23.9 &28.4& 14.6&
20.7\\
mAP@[0.85]  & 26.1 & 29.4  &20.3 &23.6& 10.8&
17.0\\
mAP@[0.90]  & 18.5 & 19.7  &14.4 &17.2& 08.5&
11.8\\
mAP@[0.95]  & 11.3 & 10.5 &08.8 &10.1 & 04.4&
08.5\\
\hline
mAP@[0.5-0.95] &36.8 &38.8 & 27.3& 33.9& 21.2& 27.0 \\ \hline
\end{tabular}
\end{table}

\paragraph{\textbf{Effect of IoU:}} 
To assess the impact of the IoU threshold on slide element detection performance, we compute mAP across multiple IoU thresholds from 0.5 to 0.95. Table~\ref{tab:iou_ablation} presents the performance of slide detection models under these thresholds. 

\paragraph{\textbf{Effect of Real Slide Images:}} 

\begin{figure}[!t]
\centerline{
\includegraphics[width=1.0\textwidth,height=0.25\textwidth]{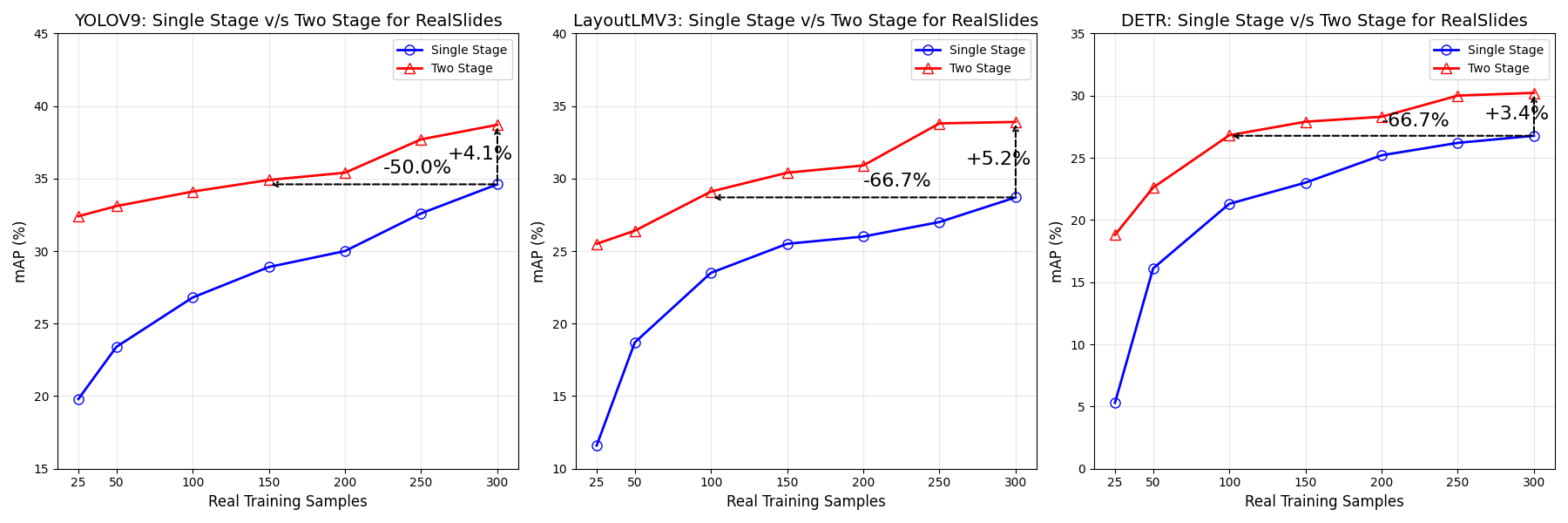}}
\caption{Effect of real slide images (RealSlide) on the performance (mAP@[0.5–0.95]) of three slide element detection models under two training strategies: (i) \textbf{Single Stage} and (ii) \textbf{Two Stage}.}\label{fig:effect_RealSlide}
\end{figure}

\begin{figure}[!t]
\centerline{
\includegraphics[width=1\textwidth,height=0.4\textwidth]{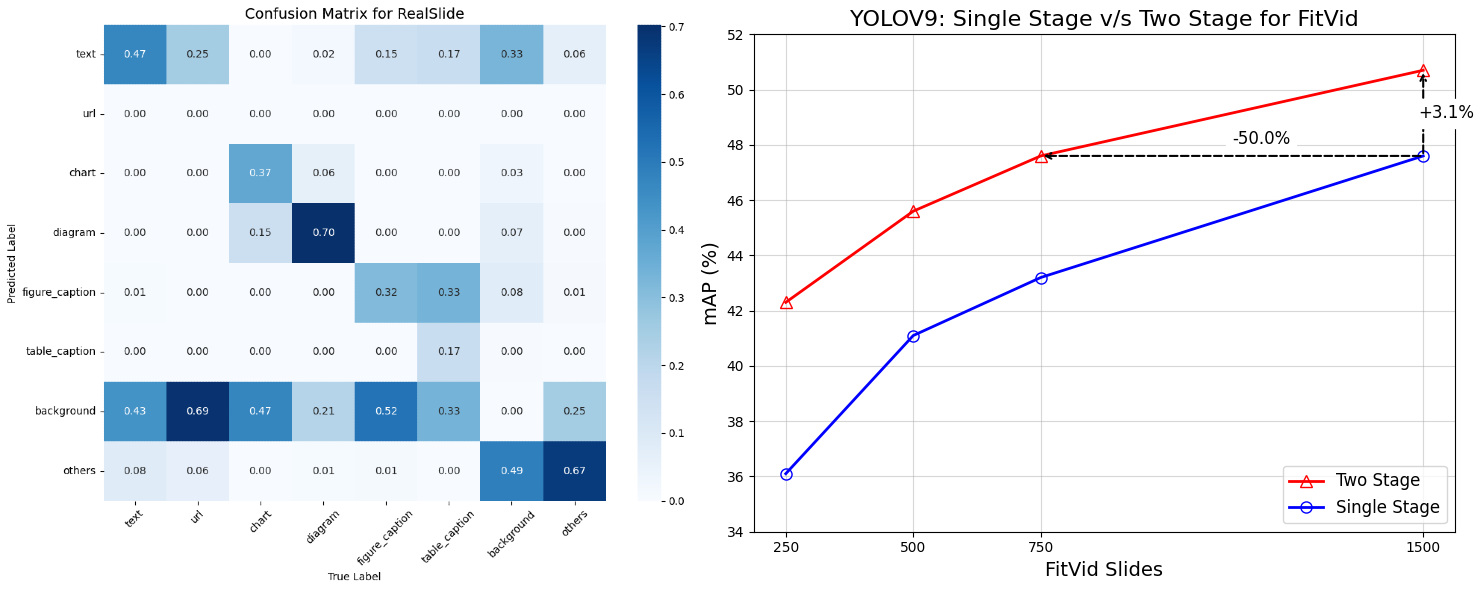}}
\caption{Left: Confusion Matrix for semantically similar classes.  Right: Effect of real slide images (FitVid) on the performance (mAP@[0.5–0.95]) of the best model (YOLOv9) for two training strategies: (i) \textbf{Single Stage} and (ii) \textbf{Two Stage}.}\label{fig:effect_real_slide}
\end{figure}

In real-world scenarios, obtaining a large number of slide images with region bounding box annotations is challenging and expensive. At the same time, annotation using synthetic data is cheap but often lacks enough variance as seen in real world data. Therefore, it is essential to determine the amount of real annotated slide images required to improve the performance of slide element detection models. To explore this, we conduct experiments under two fine-tuning strategies: (i) \textbf{Single Stage} --- the pre-trained model is directly fine-tuned with real annotated slide images and (ii) \textbf{Two Stage} --- the pre-trained model is first fine-tuned with synthetic slide images, followed by fine-tuning with real annotated slide images. We randomly select multiple subsets of 25, 50, 100, 150, 200, 250, and 300 images from the real training set and fine-tune the three detection models under both strategies. Fig.~\ref{fig:effect_RealSlide} and Fig.~\ref{fig:effect_real_slide} presents the mAP results for different models on increasing subsets of training with \textbf{RealSlide} and \textbf{FitVid} data respectively. From the results, we observe that increasing the number of real training slide images consistently and naturally enhances model performance. Additionally, the \textbf{Two Stage} strategy provides a greater performance boost compared to \textbf{Single Stage}, demonstrating that fine-tuning with synthetic slide images before real slide images leads to a marginal yet meaningful improvement. 

\paragraph{\textbf{Results Analysis:}}

Table~\ref{tab:result} presents the element-wise mean Average Precision (mAP) scores across models for both single-stage and two-stage fine-tuning under two experimental conditions: (i) a low-resource setting with only 50 real images and (ii) a higher-resource setting with 300 real images. Notably, the two-stage fine-tuning approach leads to a substantial performance boost in low-frequency classes such as \textbf{Natural Image}, \textbf{Table Caption}, and \textbf{Code}. This improvement underscores the efficacy of synthetic data in enhancing model generalization under data-scarce conditions, as further illustrated in Fig.~\ref{fig:dataset_stat}.

The confusion matrix in Fig.~\ref{fig:effect_real_slide} (left) reveals a high frequency of misclassifications among semantically similar classes, particularly between \textit{diagram} and \textit{chart}. Furthermore, several text-based categories --- including \textit{URL}, \textit{heading}, \textit{footer element}, \textit{figure caption}, \textit{table caption}, and \textit{code} --- are frequently misclassified as the generic \textit{"Text"} class. This phenomenon can be attributed to both intra-class variability (e.g., differing styles of diagrams and charts) and inter-class feature overlap (e.g., structural and contextual similarities between URLs, headings, and table captions).

Fig.~\ref{fig:det_visual_results} provides qualitative examples of slide element detection, further illustrating the strengths and limitations of the proposed approach. These findings highlight the importance of leveraging synthetic data and multi-stage fine-tuning to mitigate class imbalance and enhance recognition performance in real-world document analysis tasks. 

\begin{table}[!t]
\begingroup  
\setlength{\cellspacetoplimit}{4pt}  
\setlength{\cellspacebottomlimit}{4pt}  
\renewcommand{\arraystretch}{1.3} 

\centering
\scriptsize
\caption{Element-wise mAP @ IoU [0.50:0.95] for three slide element detection models under two fine-tune strategies: (i) \textbf{Single Stage} (\textbf{SS}) and (ii) \textbf{Two Stage} (\textbf{TS}) on the test set (750 images) of the \textbf{RealSlide} dataset.}\label{tab:result}
\begin{tabular}{||l|c c||c c||c c||c c||c c||c c||}\hline 
\multirow{2}{*}{\textbf{Element}}
& \multicolumn{6}{c||}{\textbf{Fine-tuning using 50 Real Images}} 
& \multicolumn{6}{c||}{\textbf{Fine-tuning using 300 Real Images}} \\
\cline{2-13}
& \multicolumn{2}{c||}{\textbf{YOLOv9}} 
& \multicolumn{2}{c||}{\textbf{LayoutLMV3}} 
& \multicolumn{2}{c||}{\textbf{DETR}}  
& \multicolumn{2}{c||}{\textbf{YOLOv9}} 
& \multicolumn{2}{c||}{\textbf{LayoutLMV3}} 
& \multicolumn{2}{c||}{\textbf{DETR}} \\
\hline
& \textbf{SS} & \textbf{TS} 
& \textbf{SS} & \textbf{TS}  
& \textbf{SS} & \textbf{TS}  
& \textbf{SS} & \textbf{TS} 
& \textbf{SS} & \textbf{TS}  
& \textbf{SS} & \textbf{TS}  \\
\hline
Title & 69.1 & 75.3  & 66.2 & 71.9  & 57.8 & 67.2  & 70.9  & 75.4  & 71.8  & 77.0  & 59.5  & 69.7  \\
Text  & 06.5 & 17.4  & 11.3 & 13.2  & 07.4 & 13.8  & 15.6  & 21.5  & 13.8  & 18.1  & 09.1   & 15.3  \\
Enumeration  & 57.5& 66.8  & 60.7 & 67.8  & 67.1 & 72.4  & 70.9  & 76.7  & 72.0  & 79.9  & 68.6  & 74.9  \\
URL              & 00.0 & 03.4  & 00.0& 00.8  & 00.8 & 01.5  & 03.4   & 01.3   & 02.4   & 02.1   & 01.8   & 02.7   \\
Equation         & 09.2 & 28.3  & 00.9 & 09.8  & 16.6 & 20.2  & 23.0  & 27.4  & 16.5  & 24.2  & 18.3  & 22.1  \\
Table            & 57.8 & 60.1  & 35.7 & 43.3  & 48.5 & 50.6  & 82.7  & 56.2  & 59.0  & 63.1  & 56.5  & 51.3  \\
Diagram          & 25.8 & 46.3  & 26.6 & 39.9  & 33.7 & 41.0  & 53.8  & 58.6  & 46.0  & 50.4  & 38.8  & 44.7  \\
Chart            & 12.8 & 33.4  & 08.5 & 17.4 & 08.9 & 14.4  & 31.7  & 32.1  & 14.9  & 23.1  & 11.7  & 18.9  \\
Heading          & 06.1 & 09.2  & 03.8 & 16.1  & 10.6 & 18.8  & 18.6  & 22.8  & 24.6  & 35.2  & 13.1  & 20.3  \\
Slide Number     & 25.0 & 27.3  & 33.4 & 29.3  & 20.8 & 24.1  & 27.7  & 25.9  & 28.2  & 26.7  & 22.5  & 25.6  \\
Footer Element   & 48.7 & 42.2 & 47.2 & 48.0  & 36.6 & 42.0  & 51.5  & 47.7  & 43.0  & 48.9  & 40.0  & 45.1  \\
Figure caption   & 02.7 & 05.7  & 00.3 & 10.9  & 01.8 & 06.7  & 15.8  & 14.2  & 07.6   & 09.8   & 04.4   & 08.1   \\
Table caption    & 00.0 & 11.8  & 00.0 & 02.0  & 01.3 &  06.9 & 19.2  & 21.6  & 00.0   & 02.2   & 02.1   & 08.7   \\
Logo             & 48.3 & 46.1  & 03.7 & 28.1 & 18.8 & 26.2  & 67.9  & 69.4  & 26.0  & 42.9  & 22.6  & 34.7  \\
Code             & 02.1 & 34.6  & 00.0& 05.0 & 08.0 & 14.5  & 23.8  & 42.5  & 10.5  & 18.6  & 12.1  & 17.8  \\
Natural Image    & 00.7 & 20.9  & 00.0 & 12.0  & 00.4 & 11.7  & 11.8  & 27.9  & 10.4  & 18.3  & 09.3   & 14.6  \\
\hline
\textbf{Macro avg}  
& 23.3 & 33.0  & 18.6 & 26.1  & 21.2 & 27.0  & 36.8  & 38.8  & 27.9  & 33.8  & 26.8  & 30.2  \\\hline
\end{tabular}
\label{tab:50_vs_300}
\endgroup  
\end{table}

\begin{figure}[!t]
\centerline{
\fbox{\includegraphics[width=0.225\textwidth]{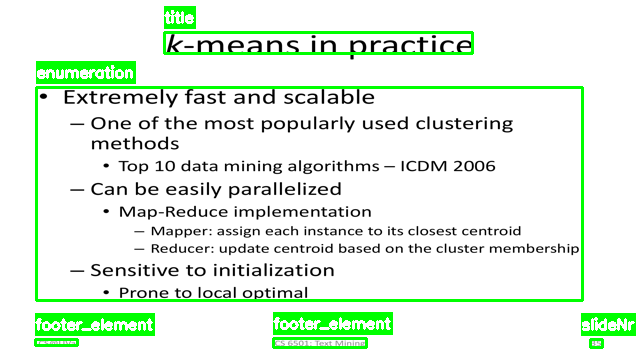}} \hspace{-0.01\textwidth}
\fbox{\includegraphics[width=0.225\textwidth]{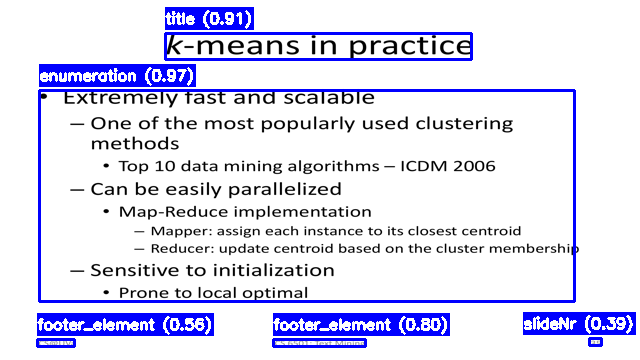}}
\hspace{-0.01\textwidth}
\fbox{\includegraphics[width=0.225\textwidth]{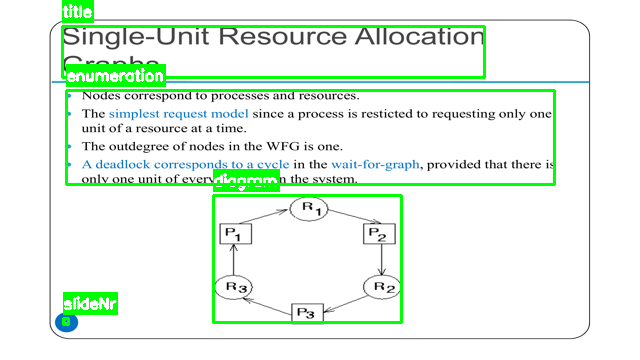}} \hspace{-0.01\textwidth}
\fbox{\includegraphics[width=0.225\textwidth]{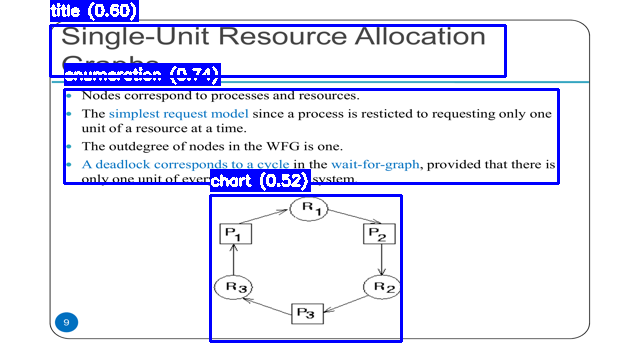}}} 
\caption{Illustration of a selection of visual results from YOLOv9 (Two-Stage), where green denotes ground truth bounding boxes and blue indicates predicted bounding boxes.}\label{fig:det_visual_results} 
\end{figure}

\subsection{Text-based Slide Retrieval} 

\paragraph{\textbf{Baselines and Implementation Details:}}

We consider CLIP~\cite{radford2021learning} as our baseline for retrieval tasks. We define the task as retrieving lecture slide images based on slide summary captions. 
We implement ViT-B/32 CLIP model and use 128 batch size keeping all other parameters standard. While finetuning, we set all parameters to be trainable. Table~\ref{tab:clipRes} demonstrates experiments comparing publicly available slide image-caption pairs along with SynRet and RealSlide train data evaluated on LecSD and RealSlide test sets.

\paragraph{\textbf{Results Analysis:}}

\begin{figure}[!t]
\textit{Query: A slide on Pseudo relevance feedback with a diagram and enumeration}
\centerline{
\subfloat[1st]{
\fbox{\includegraphics[width=0.22\textwidth]{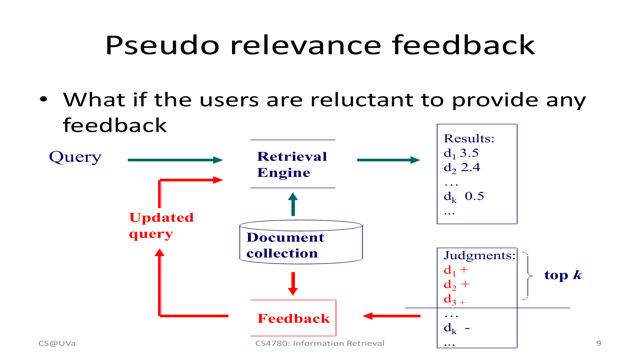}}}
\subfloat[2nd]{
\fbox{\includegraphics[width=0.22\textwidth]{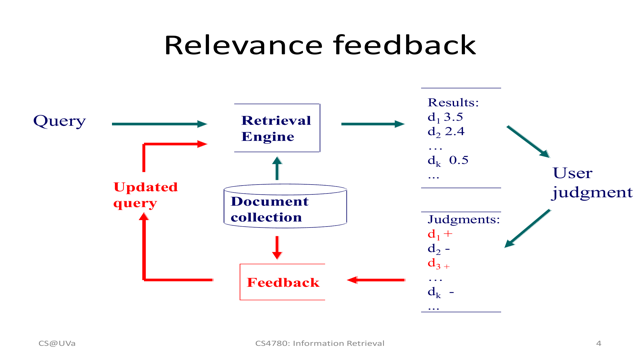}}}
\subfloat[3rd]{
\fbox{\includegraphics[width=0.22\textwidth]{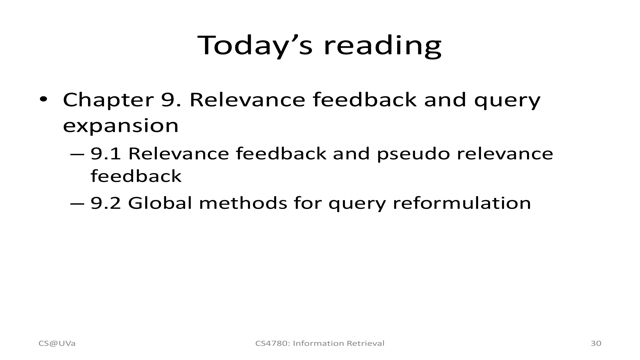}}}
\subfloat[4th]{
\fbox{\includegraphics[width=0.22\textwidth]{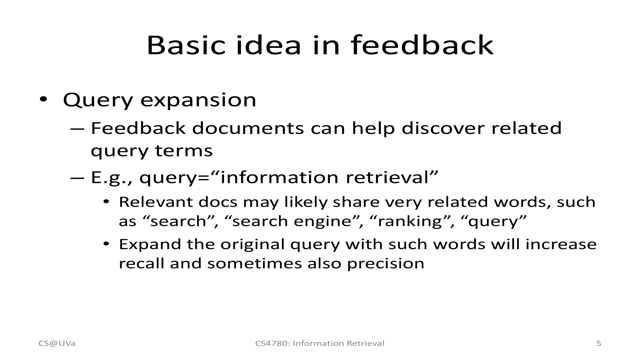}}}}
\caption{Visual example of result from the CLIP model fine-tuned using SynRet data } \label{fig:ret_visual_results}
\end{figure}

We present text-to-lecture slide retrieval results (summarized in Table~\ref{tab:clipRes}) on two benchmarks: LecSD-Test, which comprises 10,000 lecture slides, and our newly annotated RealSlide dataset containing 750 images. Our results show that the synthetic dataset is especially useful when in-domain real slide annotations are unavailable. For instance, it achieves an R@1 of 43 on the RealSlide set, outperforming fine-tuning with out-of-domain data like LecSD-Train. Additionally, our synthetic data provides a slight improvement in R@1 compared to other synthetic datasets, such as DreamStruct~\cite{peng2024dreamstruct}, highlighting the effectiveness of our approach in generating useful training data for lecture slide retrieval. We also train the CLIP model using the same two-stage fine-tuning strategy defined in the Slide Element Detection task; however, minimal improvement is observed in the performance over SynRet alone. Fig.~\ref{fig:ret_visual_results} shows a qualitative result from the RealSlide (750) test set, where the fine-tuned model using SynRet correctly retrieves the relevant slide, with the top-3 results also closely matching the query.\footnote{More generated sample slide images and qualitative results are given in supplementary material.}.

\begin{table}[!t]
\centering
\caption{Text-based lecture slide retrieval using CLIP. We report Recall@1 and Recall@10.}
\scriptsize
\begin{tabular}{|ccc|r|r|}
\hline
\multicolumn{3}{|c|}{\textbf{Dataset}}                                                                                                  & \textbf{R@1} & \textbf{R@10} \\ \hline
\multicolumn{1}{|c|}{\textbf{Finetuning dataset (\# Samples)}} & \multicolumn{1}{c|}{\textbf{Test Dataset (\# Samples)}}     & \textbf{In-domain} &     &      \\ \hline
\multicolumn{1}{|l|}{None (zero-shot)}                  & \multicolumn{1}{l|}{\multirow{5}{*}{LecSD-Test (10,000)}} & NA        & 16  & 44   \\ \cline{1-1} \cline{3-5} 
\multicolumn{1}{|l|}{LecSD-Train (31,475)}              & \multicolumn{1}{l|}{}                                     & Yes       & 45  & 78   \\ \cline{1-1} \cline{3-5} 
\multicolumn{1}{|l|}{DreamStruct (3,183)}               & \multicolumn{1}{l|}{}                                     & No        & 26  & 59   \\ \cline{1-1} \cline{3-5} 
\multicolumn{1}{|l|}{SynRet (2,200)}                    & \multicolumn{1}{l|}{}                                     & No        & 26  & 60   \\ \cline{1-1} \cline{3-5} 
\multicolumn{1}{|l|}{RealSlide (300)}                  & \multicolumn{1}{l|}{}                                     & No        & 20  & 49   \\ \hline
\multicolumn{1}{|l|}{None (zero-shot)}                  & \multicolumn{1}{l|}{\multirow{5}{*}{RealSlide (750)}}     & NA        & 33  & 63   \\ \cline{1-1} \cline{3-5} 
\multicolumn{1}{|l|}{LecSD-Train (31,475)}              & \multicolumn{1}{l|}{}                                     & No        & 31  & 57   \\ \cline{1-1} \cline{3-5} 
\multicolumn{1}{|l|}{DreamStruct (3,183)}               & \multicolumn{1}{l|}{}                                     & No        & 42  & 67   \\ \cline{1-1} \cline{3-5} 
\multicolumn{1}{|l|}{SynRet (2,200)}                    & \multicolumn{1}{l|}{}                                     & No        & 43  & 69   \\ \cline{1-1} \cline{3-5} 
\multicolumn{1}{|l|}{RealSlide (300)}                  & \multicolumn{1}{l|}{}                                     & No        & 40  & 69   \\ \cline{1-1} \cline{3-5} 
\multicolumn{1}{|l|}{SynRet(2200) + RealSlide (300)}                 & \multicolumn{1}{l|}{}                                     & No        & 43  & 70   \\ \hline
\end{tabular}\label{tab:clipRes}
\end{table}

\section{Conclusion}

We proposed an open-source LLM-driven pipeline to generate realistic, coherent, and copyright-free lecture slides with automatic annotations, resulting in the \textbf{SynSlide} dataset for slide understanding. We also introduce \textbf{RealSlide}, an evaluation set of 1,050 manually annotated real slides for benchmarking slide element detection and text-based lecture slide image retrieval. Experiments show that few-shot transfer learning on real slides using models pre-trained on synthetic slides outperforms training on real data alone, emphasizing the value of synthetic slides in low-supervision settings.

Our analysis reveals that existing vision models struggle with fine-grained document tasks, exhibiting high-class confusion and limited gains from more training data. While our synthetic slides help improve performance across multiple tasks with no manual annotation, increasing synthetic data does not lead to proportional gains due to limited data variability. Future work may explore diffusion-based layout generation, context integration, and better generalization. 

Overall, our work demonstrates the promise of synthetic data for lecture slide understanding and establishes a foundation for scalable and versatile benchmarking in this domain.

\section*{Acknowledgments:}
This work is supported by the MeitY Government of India, through the NLTM Bhashini (\url{https://bhashini.gov.in/}) project.

\bibliographystyle{splncs04}
\bibliography{reference}

\newpage

\title{Supplementary Material: AI-Generated Lecture Slides for Improving Slide Element Detection and Retrieval}
\titlerunning{Supplementary Material}

\author{
Suyash Maniyar\thanks{These authors contributed equally.}\inst{1}\orcidID{0009-0000-5882-4377} \and
Vishvesh Trivedi$^{*}$\inst{2}\orcidID{0009-0004-5043-6766} \and
Ajoy Mondal\inst{3}\orcidID{0000-0002-4808-8860} \and
Anand Mishra\inst{1}\orcidID{0000-0002-7806-2557} \and
C.~V.~Jawahar\inst{3}\orcidID{0000-0001-6767-7057}
}
\authorrunning{Maniyar et al.}

\institute{
Indian Institute of Technology, Jodhpur, India \\
\email{suyash.1@alumni.iitj.ac.in, mishra@iitj.ac.in}
\and 
Sardar Vallabhbhai National Institute of Technology, Surat, India \\
\email{u20cs130@coed.svnit.ac.in} 
\and
CVIT, International Institute of Information Technology, Hyderabad, India \\
\email{\{ajoy.mondal, jawahar\}@iiit.ac.in}
}
\maketitle
\appendix

\begin{figure}[!h]
\includegraphics[width=1\textwidth]{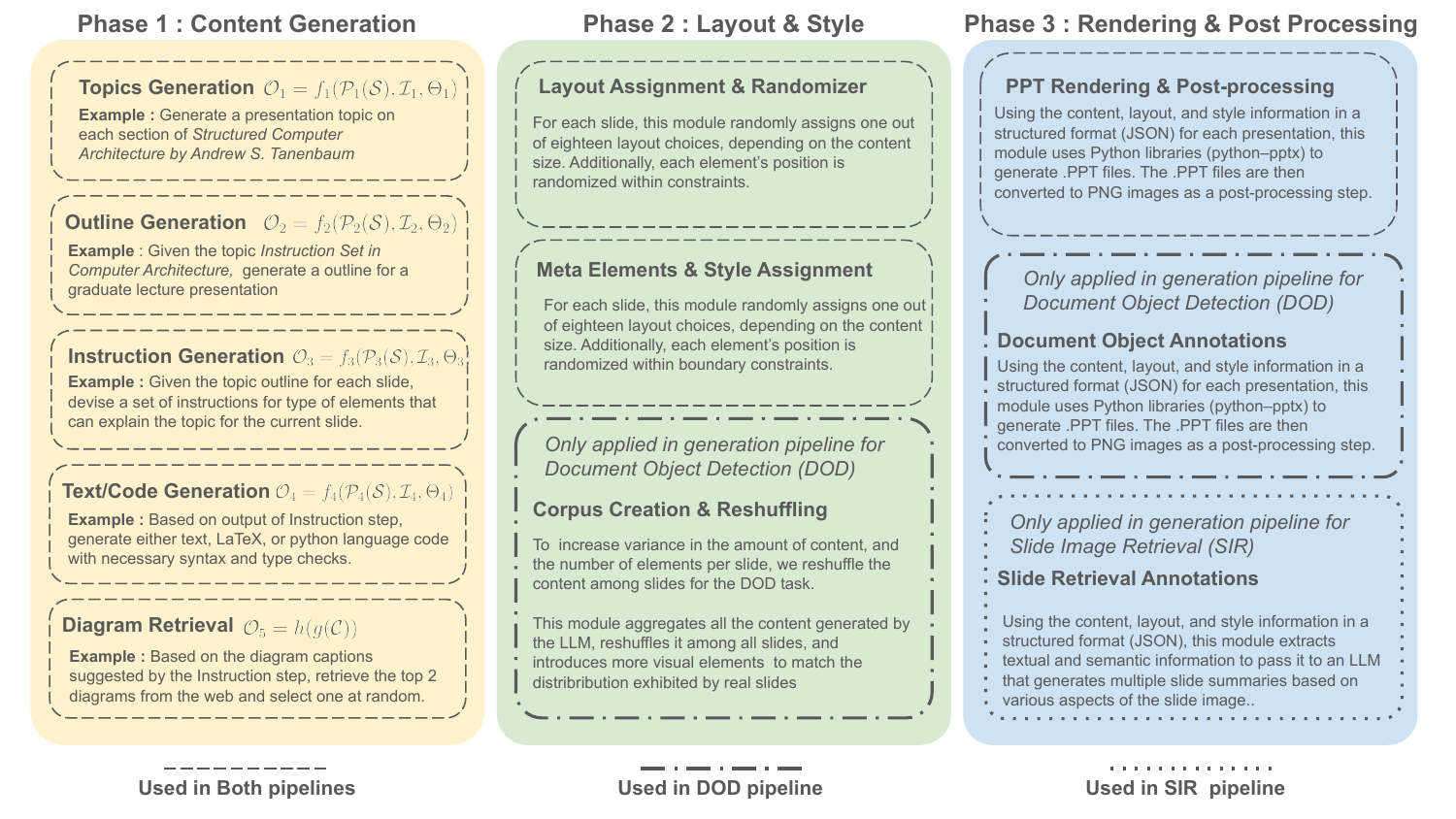}
\caption{Detailed overview of each module in the \textbf{SynLecSlideGen} pipeline.}
\label{fig:detailed_overview} 
\end{figure}

\begin{center}

The goal of this appendix is to provide an in-depth description of the \textbf{SynLecSlideGen} Pipeline and provide visual results for both the experimented tasks. A detailed overview of the pipeline can be obtained from Figure \ref{fig:detailed_overview}
\end{center}

\section{Detailed Prompting Techniques used for Content Generation}\label{appendix:syn-gen-method}

\subsection{Model Configurations and Settings}
Table~\ref{tab:model-config} summarizes the LLM configurations used across the four stages of the generation pipeline. We utilized state-of-the-art models such as GPT-4 and GPT-3.5, selected for their ability to handle complex, multi-modal content generation tasks.

\begin{table}[h!]
    \centering
    \begin{tabular}{|l|l|c|c|c|}
        \hline
        \textbf{Stage} & \textbf{Model} & \textbf{Temperature (\(T\))} & \textbf{Top-p (\(p\))} & \textbf{Max Tokens (\(L\))} \\ \hline
        Topic Generation & GPT-4 & 0.7 & 0.9 & 256 \\ \hline
        Outline Generation & GPT-4 & 0.6 & 0.95 & 512 \\ \hline
        Instruction Generation & GPT-3.5 & 0.7 & 0.9 & 1024 \\ \hline
        Text/Code Generation & GPT-3.5 & 0.5 & 0.8 & 2048 \\ \hline
    \end{tabular}
    \caption{Model configurations and hyperparameter settings for each stage.}
    \label{tab:model-config}
\end{table}

\subsection{Detailed Prompting Techniques}
We provide examples of prompts, instructions, and outputs for each stage of the pipeline to illustrate our approach. The outputs demonstrate the progression from initial topic generation to fully realized slide content.

\subsubsection{Topic Generation}
In this stage, the model generates a list of potential topics based on a seed input, such as a book title or a course name.

\paragraph{Prompt Provided:}
\begin{verbatim}
System Prompt:
You are a helpful assistant to a professor. 
You have access to all resources on the web.

User Prompt:
I am a {subject} professor and I want to create 
lecture slides for various topics in the book 
: {book} by {author}.
Give me a list of topics from the book I provided 
in the form of a Python dict.
Also, I will use these topics to make presentations 
and hence augment the presentation titles if they are generic
For example: for titles like Introduction, Applications 
convert them to Introduction to Deep Learning, 
Applications of Deep Learning, respectively.
Do not provide any converstation.
\end{verbatim}

\paragraph{One Shot Example:}
\begin{verbatim}
Here's an example
     
Book Name : Understanding Deep Learning
Subject : CS
Expected Output:
{{
    "CS": [
        "Math for Deep Learning Basics",
        "Intro to Supervised Learning",
        "Shallow Neural Networks",
        "Activation Functions",
        "Composing Neural Networks",
        "Gradient Descent Optimization",
        "Stochastic Gradient Descent",
        "Adam Optimization Algorithm",
        "Backpropagation in Toy Model",
        "Initialization Techniques",
        "MNIST-1D Performance Analysis",
        "Bias-Variance Trade-off",
        "Double Descent Phenomenon",
        "L2 Regularization Techniques",
        "Implicit Regularization Methods",
        "Model Ensembling Techniques",
        "Bayesian Methods in ML",
        "Data Augmentation Techniques",
        "1D Convolution Basics",
        "Convolution for MNIST-1D",
        "2D Convolution Basics",
        "Downsampling & Upsampling",
        "Shattered Gradients Issue",
        "Residual Networks Introduction",
        "Batch Normalization Role",
        "Self-Attention Mechanisms",
        "Multi-Head Self-Attention",
        "Graph Encoding Techniques",
        "Graph Classification Methods",
        "Neighborhood Sampling",
        "Graph Attention Mechanisms",
        "GAN Toy Example",
        "Wasserstein Distance in GANs",
        "1D Normalizing Flows Intro",
        "Autoregressive Flows Intro",
        "Latent Variable Models Intro",
        "Reparameterization Trick",
        "Importance Sampling Methods",
        "Diffusion Encoder Basics",
        "1D Diffusion Model Basics",
        "Reparameterized Model Intro",
        "Diffusion Models Families",
        "Markov Decision Processes Intro",
        "Dynamic Programming Basics",
        "Monte Carlo Methods Intro",
        "Temporal Difference Methods",
        "Control Variates Methods",
        "Random Data Generation",
        "Full-Batch Gradient Descent",
        "Lottery Tickets Hypothesis",
        "Adversarial Attacks Techniques",
        "Bias Mitigation Strategies",
        "Explainability Techniques"
    ]
}}
""")]
\end{verbatim}

\paragraph{Settings:} GPT-4, \(T=0.7\), \(p=0.9\), \(L=256\).

\subsubsection{Outline Generation}
The topics generated in the previous stage are expanded into a detailed outline.

\paragraph{Prompt Provided:}
\begin{verbatim}
System Prompt:
You are a helpful university professor and you 
are guiding your PhD student to create an outline
for the lecture he will deliver to a class.



User Prompt:
I would like to get help designing a detailed 
Table of Contents for an advanced university
presentation lecture on {topic} based on the book {book}. 
Please help me create the Table of Content in form of a Python dict.
\end{verbatim}

\paragraph{One Shot Example:}
\begin{verbatim}


Example:
Input: Tree Data Structures based on the book
Data Structures and Algorithms made easy. 

Expected Output:
[
"Introduction : Definition & Characteristics",
"Introduction : Example of a Tree",
"Types of Trees",
"Binary Trees : What are they?",
"Binary Trees : Searching an element",
"Tree Traversal",
"Pre-order Traversal",
"In-order Traversal",
"Post-order Traversal",
"Comparing traversal methods",
"Binary Search Trees: Introduction",
"Binary Search Trees: Time Complexity",
"BST v/s Binary Trees",
"Applications of Trees",
"Huffman Algorithm : History",
"Huffman Algorithm : Pseudocode",
"Summary of Trees"
]
 
I want you to provide the output in form of a python
list of strings having slide titles of each slide. 
The length of the list will be the total number of slides
in the presentation. Do not generate more than 15 slide
titles and each title should have maximum of 5 words.       
Just return the output without the conversation.""")
]
\end{verbatim}

\paragraph{Settings:} GPT-4, \(T=0.6\), \(p=0.95\), \(L=512\).

\subsubsection{Instruction Generation}
This stage translates the outline into detailed instructions for creating slide content.

\paragraph{Prompt Provided:}
\begin{verbatim}
User Prompt:

Hello. I want you to help me prepare lecture slides on {topic}. 
I am providing you with the outline of the lecture
in form of a nested dictionary.

Outline :{outline}
Each element in the object is a slide where the value
represent the slide title. I want you to add two keys
namely 'element_type' and 'element_caption' for each
subsection and determine which types of elements would
be most beneficial to understand that subsection.

The elements can be as follows: {elements}.
You should generate exactly two elements per slide. 
Do not generate more of less elements per slide.
Whenever possible generate atleast one text based element
(Description, URL, or Enumeration) and one visual element
(Rest of the elements) per subsection, such that there
is diversity in elements.

As a rule of thumb, make sure the distribution of elements
is nearly same for the entire presentation.
I want you to generate the results within the outline
and only output the revised outline without any conversation.
Do not generate the slide numbers in the output, 
they are just for your reference.
Your output should be in form of a Python Dict.
\end{verbatim}

\paragraph{Expected Output:}
\begin{verbatim}
{
    "Introduction to Gaussian Distributions": [
        {"element_type": "description", "element_caption":
        "Overview of Gaussian distributions and their
        importance in statistics"},
        {"element_type": "graph", "element_caption": "Visual
        introduction to the bell curve shape of Gaussian
        distributions"}
    ],
    
    "Historical Background": [
        {"element_type": "description", "element_caption":
        "Discussion on the origin and development of Gaussian
        distributions"},
    ],
    "The Normal Distribution: Definition": [
        {"element_type": "description", "element_caption":
        "Formal definition of the normal distribution"},
        {"element_type": "equation", "element_caption":
        "Mathematical equation of the normal distribution"}
    ],
    "Properties of Gaussian Distributions": [
        {"element_type": "block-diagram", "element_caption":
        "Diagram showing key properties such as symmetry and
        bell shape"},
        {"element_type": "enumeration", "element_caption":
        "List of statistical properties like mean, variance,
        etc."}
    ],
    ...
}
\end{verbatim}

\paragraph{Settings:} GPT-3.5, \(T=0.7\), \(p=0.9\), \(L=1024\).

\subsubsection{Text/Code Generation}
In the final stage of synthetic data generation, we classify elements based on their type to construct specialized prompts for generating content. Each class type corresponds to a distinct content generation mode:
\begin{itemize}
    \item \textbf{Textual elements:} Captions, headings, descriptions, and enumerations, for which simple text is generated.
    \item \textbf{LaTeX-rendered elements:} Tables and equations, for which LaTeX code is generated.
    \item \textbf{Code-rendered elements:} Plots, charts, and diagrams, for which code in Python (Matplotlib) or DOT language is generated.
\end{itemize}

This approach ensures that the model adheres to specific output requirements for different types of slide content. Below, we describe the process using pseudocode and provide example prompts for a textual "description" element and a LaTeX-rendered "equation" element.

\subsubsection{Pseudocode for Prompt Construction}
The construction of prompts involves iterating through each slide's elements, determining the type of each element, and appending specialized instructions to the respective prompt. The pseudocode for this process is as follows:

\begin{verbatim}
1. Initialize three prompts: 
   a) Text prompt for textual elements.
   b) Structural prompt for LaTeX-rendered elements.
   c) Code prompt for plots and diagrams.

2. For each slide:
   a) Iterate through all elements in the slide.
   b) Identify the element type 
   (e.g., "description", "equation", "plot").
   c) Generate type-specific instructions:
      i. For "textual" types:
         Append context and instructions for generating simple text.
      ii. For "LaTeX-rendered" types:
         Append context and instructions for generating LaTeX code.
      iii. For "code-rendered" types:
         Append context and instructions for generating Matplotlib 
         or DOT code.

3. Add post-generation checks to each prompt:
   a) Ensure syntax correctness.
   b) Verify that the number of generated snippets matches the requests.
\end{verbatim}

\subsubsection{Example Prompts}

\paragraph{Prompt for a Textual Element (Description)}
This prompt instructs the model to generate descriptive text based on the provided caption and context.

\begin{verbatim}
For the section title "Overview of Machine Learning" (Slide Number 1),
generate a short descriptive text for the slide. 
Caption: "Introduction to the basic concepts and applications of 
Machine Learning."
\end{verbatim}

\paragraph{Generated Output:}
\begin{itemize}
    \item Machine Learning is a field of Artificial Intelligence that focuses on using data-driven approaches to make predictions or decisions. Applications include image recognition, natural language processing, and recommendation systems.
\end{itemize}

\paragraph{Prompt for a LaTeX-Rendered Element (Equation)}
This prompt instructs the model to generate LaTeX code for an equation.

\begin{verbatim}
For the section title "Mathematics of Machine Learning" (Slide Number 2),
generate LaTeX code for a simple equation given the caption: 
"Representation of a linear regression model."

Generate LaTeX code as plain text separated by ```latex<content>```. 
Do not include equation numbers.
\end{verbatim}

\paragraph{Generated Output:}
\begin{verbatim}
```latex
y = w_1x_1 + w_2x_2 + \dots + w_nx_n + b
```
\end{verbatim}

\subsubsection{Post-Processing and Validation} After generating all prompts and receiving the outputs, post-processing involves: \begin{itemize} \item \textbf{Syntax Verification:} Ensuring that the LaTeX and code snippets are syntactically correct. \item \textbf{Snippet Count Matching:} Confirming that the number of generated snippets matches the number of requests for each class type. \end{itemize}

This modular approach to prompt construction allows for efficient generation of diverse content types while maintaining high output quality.

\subsection{Summary}
The multi-stage prompting strategy ensures semantic alignment and logical flow in slide content generation. By carefully designing prompts and tuning model settings, we successfully generated a dataset that mimics real-world lecture slide structures while remaining fully annotated for downstream tasks.

\section{Layout and Style Discriminator Module}

\begin{figure}[!h]
\includegraphics[width=1\textwidth]{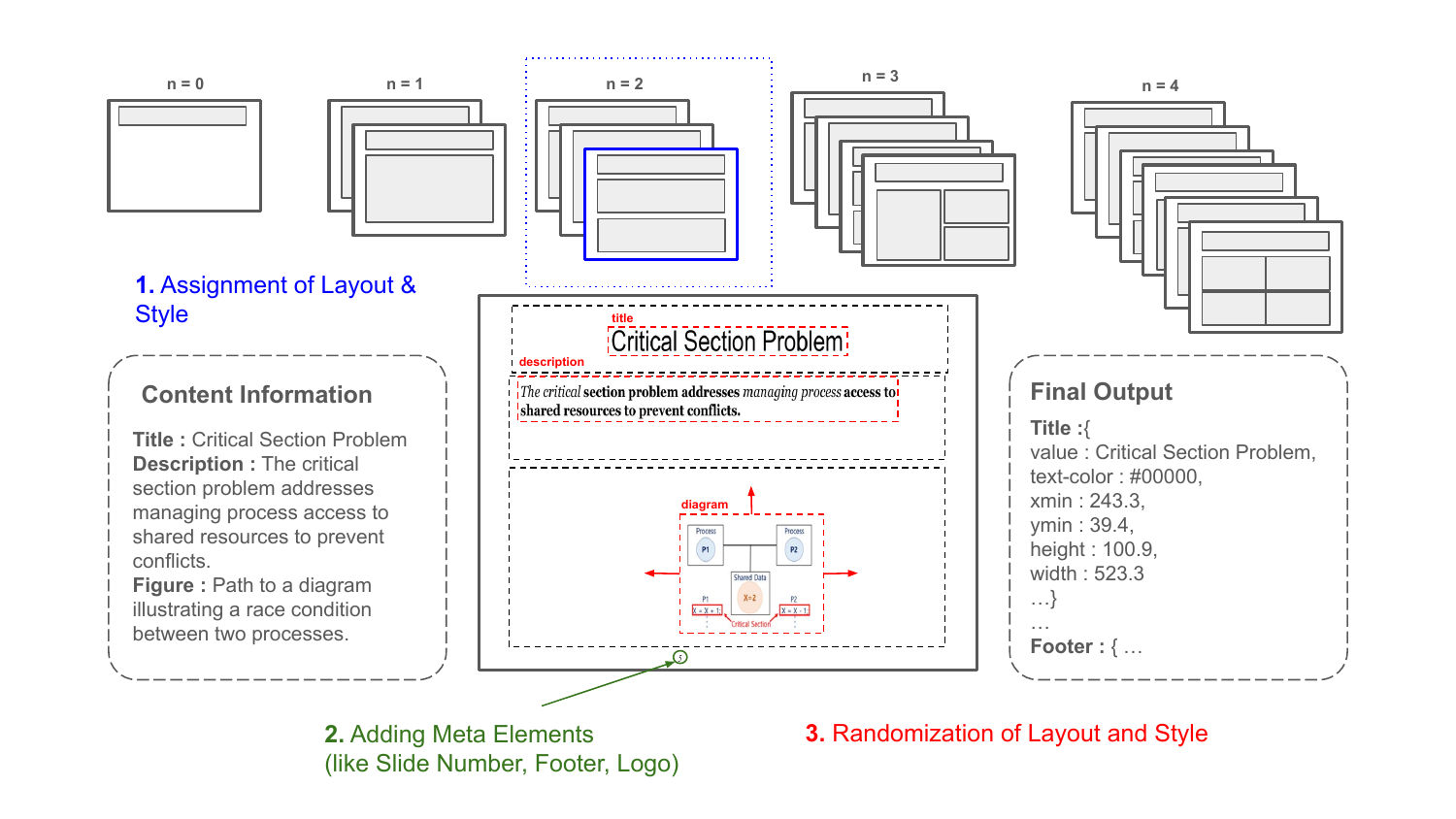}
\caption{Overview of Layout and Style Discriminator module.}
\label{fig:phase2_overview} 
\end{figure}
After generating content as textual descriptions and code snippets in Phase 1, the Layout and Style Discriminator Module is responsible for determining how this content is rendered into slides. This process involves selecting appropriate layouts based on the number of elements, introducing randomness for aesthetic variation, and assigning styles for consistent presentation. \ref{fig:phase2_overview} provides a schematic overview of the process.

\subsection{Layout Assignment Process}
The layout assignment process ensures that the arrangement of elements on the slide follows logical and visually appealing structures. The process is as follows:

\subsubsection{Defining Layout Templates}
Layouts are predefined using a Python class, where each layout specifies the dimensions and positions of the slide elements (e.g., title, body, footer). These layouts are parameterized using a scaling factor \( \text{MUL\_FAC} \) to accommodate different resolutions.

\paragraph{Pseudocode: Layout Definitions}
\begin{verbatim}
1. Initialize a dictionary `dimensions` that stores layouts for:
   a) Title: Single large text area at the top of the slide.
   b) Footer: Three equally spaced text areas at the bottom of the slide.
   c) Body: Configurations for 1 to 3 elements arranged in:
      i. Single column.
      ii. Two-column format.
      iii. Two-row format.
      iv. Grid layout (e.g., 3 columns).

2. Define a method `get_layout_dimensions(layout_id)`:
   a) Match `layout_id` with predefined layouts.
   b) Update the `dimensions` dictionary with corresponding element positions.
   c) Return updated dimensions for rendering.
\end{verbatim}

\subsubsection{Random Layout Selection}
Layouts are assigned based on the number of content elements to be displayed in the slide. Each slide's content is classified by the number of elements, and a random layout is selected from permissible layouts for that count.

\paragraph{Pseudocode: Random Layout Selection}
\begin{verbatim}
1. Create a mapping `layout_mapping`:
   a) Keys: Number of elements on the slide.
   b) Values: List of permissible layout IDs.

2. Function `generate_random_layout(total_body_elements)`:
   a) Input: Total number of elements.
   b) Output: Randomly choose a layout ID from `layout_mapping`.
\end{verbatim}

\subsubsection{Position Perturbation for Randomness}
Given a slide element with dimensions:
\begin{itemize}
    \item $d_{\text{left}}$: Left coordinate.
    \item $d_{\text{top}}$: Top coordinate.
    \item $W$: Width.
    \item $H$: Height.
\end{itemize}

We perturb the position and size of the element to introduce randomness while maintaining structural coherence. Let $\sigma$ represent the standard deviation of the Gaussian noise added to each coordinate. The perturbation is calculated as follows:

Let the perturbed top and left coordinates be $d'_{\text{top}}$ and $d'_{\text{left}}$, respectively.

\begin{align}
    d'_{\text{top}} &= d_{\text{top}} + \frac{H - h_0}{2} + \text{clip}\left(\mathcal{N}(0, \sigma), -\frac{H - h_0}{2}, \frac{H - h_0}{2}\right) \\
    d'_{\text{left}} &= d_{\text{left}} + \frac{W - w_0}{2} + \text{clip}\left(\mathcal{N}(0, \sigma), -\frac{W - w_0}{2}, \frac{W - w_0}{2}\right)
\end{align}

Here:
\begin{itemize}
    \item $h_0$: Adjusted height, scaled based on the element type.
    \item $w_0$: Adjusted width, scaled based on the element type.
    \item $\mathcal{N}(0, \sigma)$: Gaussian noise with mean $0$ and standard deviation $\sigma$.
    \item $\text{clip}(x, a, b)$: A function that restricts $x$ to the range $[a, b]$:
    \[
    \text{clip}(x, a, b) = 
    \begin{cases} 
      a & \text{if } x < a \\
      x & \text{if } a \leq x \leq b \\
      b & \text{if } x > b
    \end{cases}
    \]
\end{itemize}

The adjusted height $h_0$ and width $w_0$ are computed as:
\begin{align}
    h_0 &= \alpha \cdot H \\
    w_0 &= \alpha \cdot W
\end{align}
where $\alpha$ is a scaling factor sampled uniformly:
\[
\alpha \sim \mathcal{U}(\tau, 1)
\]
Here, $\tau$ is a predefined lower bound for the scaling factor.

The standard deviation $\sigma$ for the Gaussian noise and the scaling factor $\alpha$ may vary depending on the type of element:
\begin{itemize}
    \item \textbf{Title}:
    \begin{align*}
        \sigma &= 0.5 \\
        h_0 &= 0.8 \cdot H \\
        w_0 &= 0.8 \cdot W
    \end{align*}
    \item \textbf{Body}:
    \begin{align*}
        \sigma &= 1.0 \\
        h_0 &= \alpha \cdot H \quad \text{with } \alpha \sim \mathcal{U}(\tau, 1) \\
        w_0 &= \alpha \cdot W \quad \text{with } \alpha \sim \mathcal{U}(\tau, 1)
    \end{align*}
    \item \textbf{Footer}:
    \begin{align*}
        \sigma &= 0.2 \\
        h_0 &= H \\
        w_0 &= 0.8 \cdot W
    \end{align*}
\end{itemize}

\paragraph{Pseudocode: Perturbing Layout Positions}
\begin{verbatim}
1. Function `randomize_location(dims, element)`:
   a) Inputs:
      i. `dims`: Dictionary with position and size (top, left, height, width).
      ii. `element`: Type of slide element (title, body, footer).
   b) Process:
      i. Adjust dimensions using scaling factors for height and width.
      ii. Perturb top and left coordinates using Gaussian noise (STD).
   c) Outputs: Updated (left, top, width, height) coordinates.
\end{verbatim}

\subsubsection{Handling Meta-Elements}
Meta-elements such as slide numbers, footers, dates, and presenter names are added at predefined positions (e.g., the bottom of the slide). Their positions are slightly randomized using the perturbation function.

\subsection{Style Assignment}
Styles determine the visual presentation of elements, including fonts, colors, and backgrounds. Randomized styles are assigned to ensure slides appear distinct yet coherent.

\paragraph{Style Assignment Steps:}

\begin{enumerate}
    \item Define a pool of permissible styles for:
    \begin{itemize}
        \item Titles: Font size, color, alignment.
        \item Body text: Font type, spacing, indentation.
        \item Background: Solid colors, gradients, or images.
    \end{itemize}
    \item Randomly select styles for each slide element.
    \item Apply styles consistently across all elements of the same type within a slide.
\end{enumerate}

\subsection{Example Workflow}
Consider a slide with two body elements. The workflow is as follows:
\begin{enumerate}
    \item \textbf{Content Analysis:} Count the number of elements (\( n=2 \)).
    \item \textbf{Random Layout Selection:} Choose a layout ID (e.g., ID=2) from permissible layouts for \( n=2 \).
    \item \textbf{Randomization:} Apply position perturbations to layout elements.
    \item \textbf{Style Assignment:} Randomly assign styles to the title, body, and background.
\end{enumerate}

\subsection{Type of Layouts and Design Templates}

\begin{figure}[htbp]
    \centering
    \subfloat[n = 0]{\fbox{\includegraphics[width=0.3\textwidth]{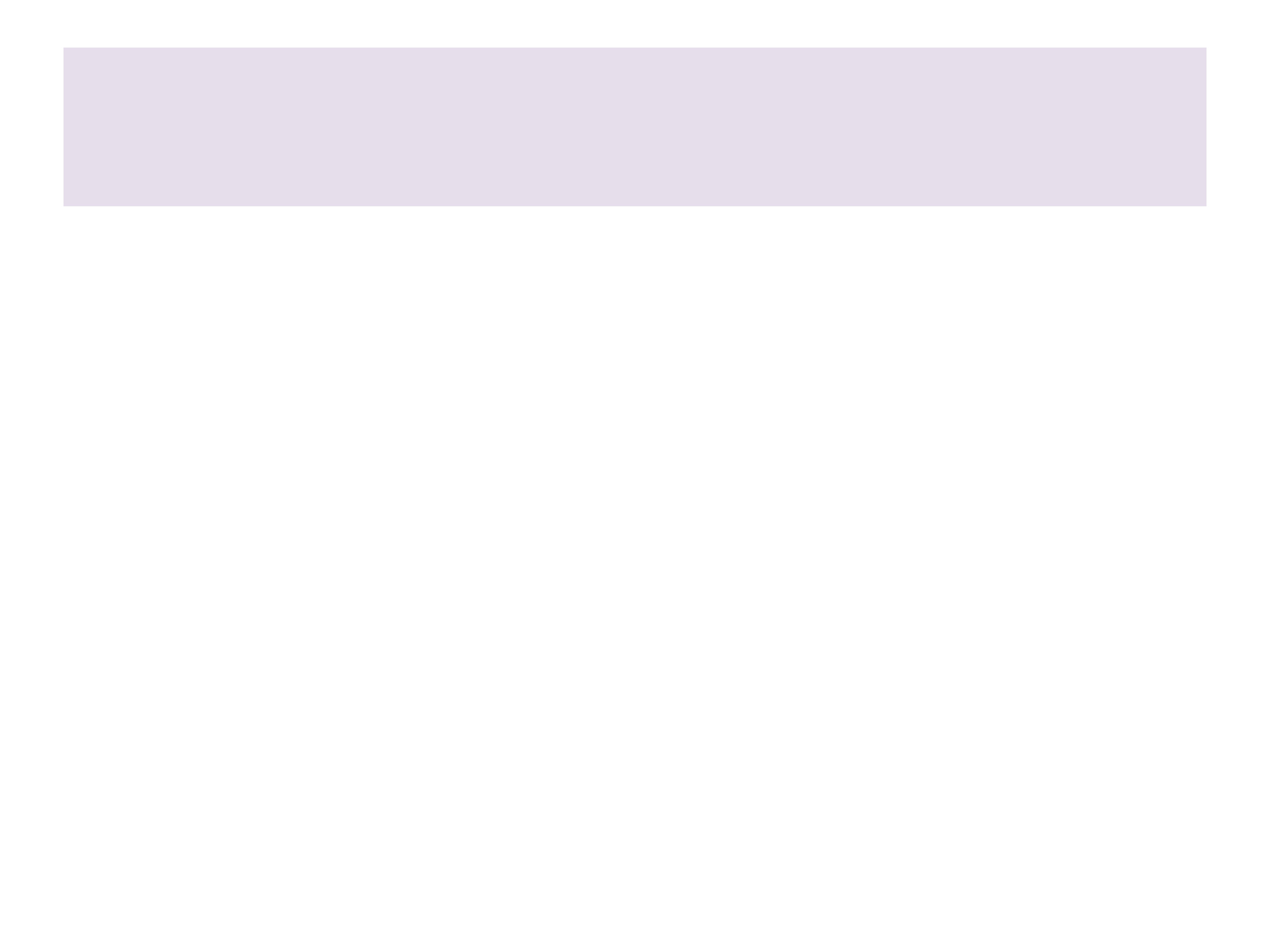}}}
    \subfloat[n = 1]
    {\fbox{\includegraphics[width=0.3\textwidth]{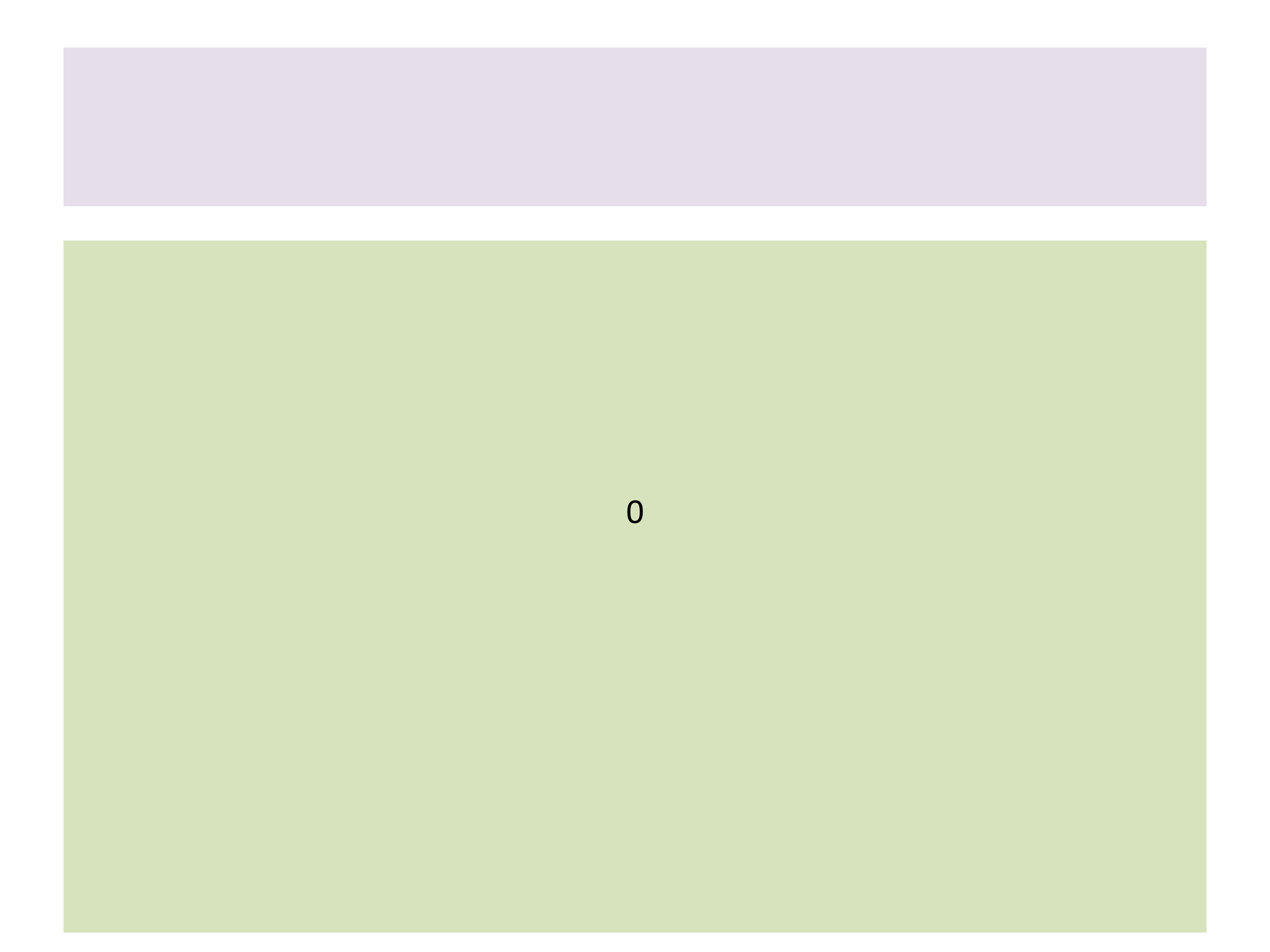}}}
    \subfloat[n = 2]
    {\fbox{\includegraphics[width=0.3\textwidth]{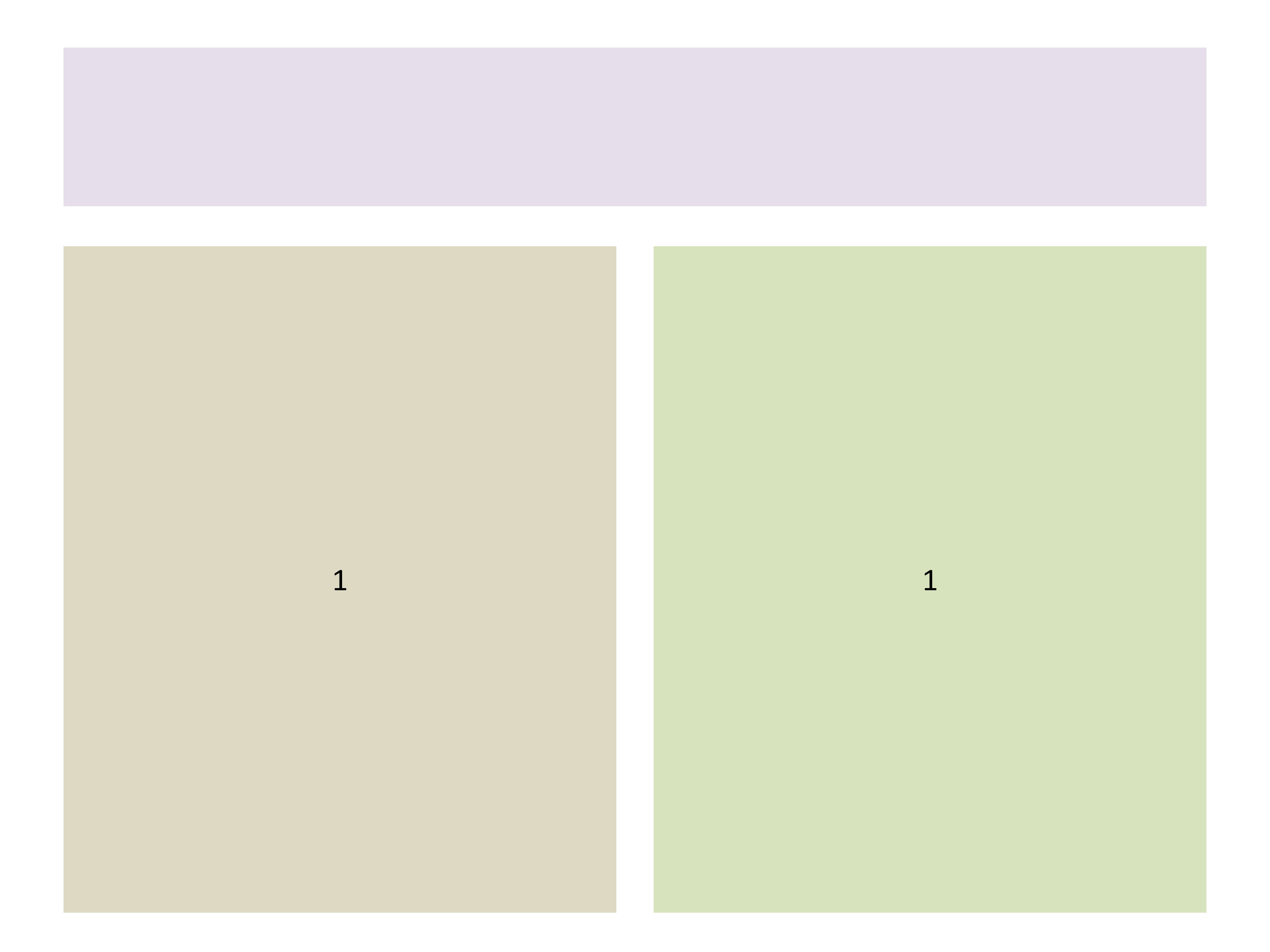}}} \\
    \subfloat[n = 2]
    {\fbox{\includegraphics[width=0.3\textwidth]{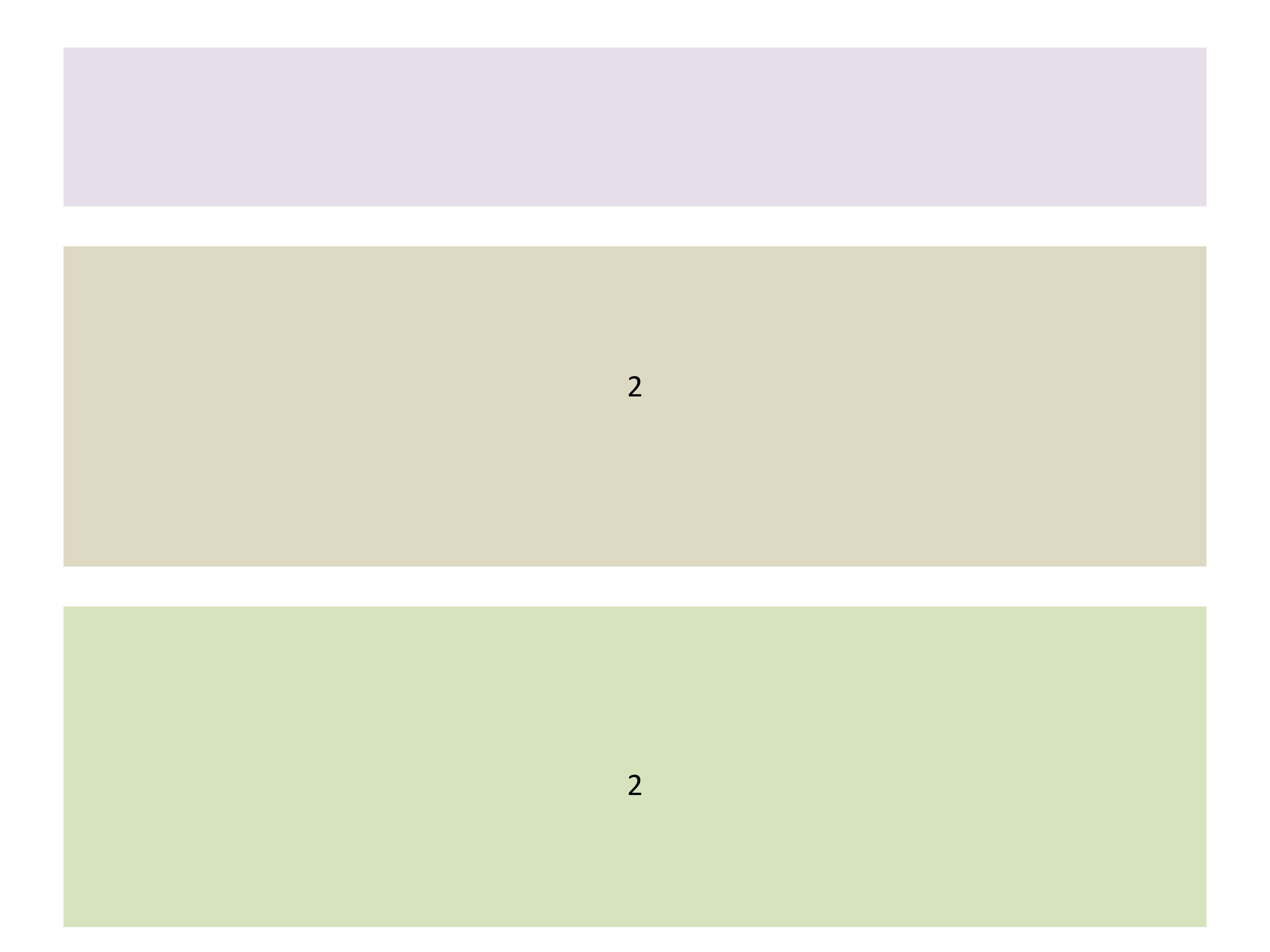}}}
     \subfloat[n = 3]
    {\fbox{\includegraphics[width=0.3\textwidth]{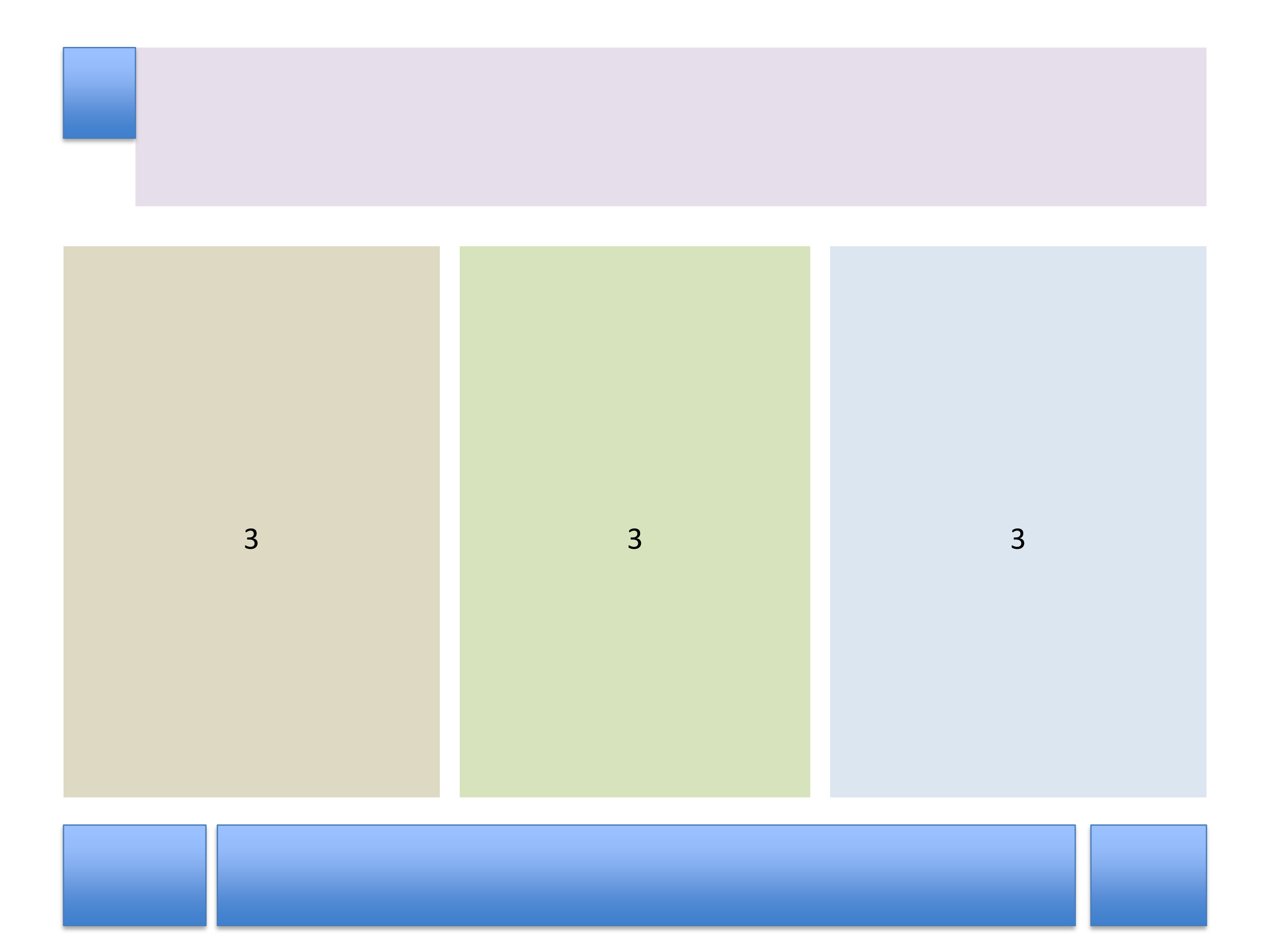}}}
     \subfloat[n = 3]
    {\fbox{\includegraphics[width=0.3\textwidth]{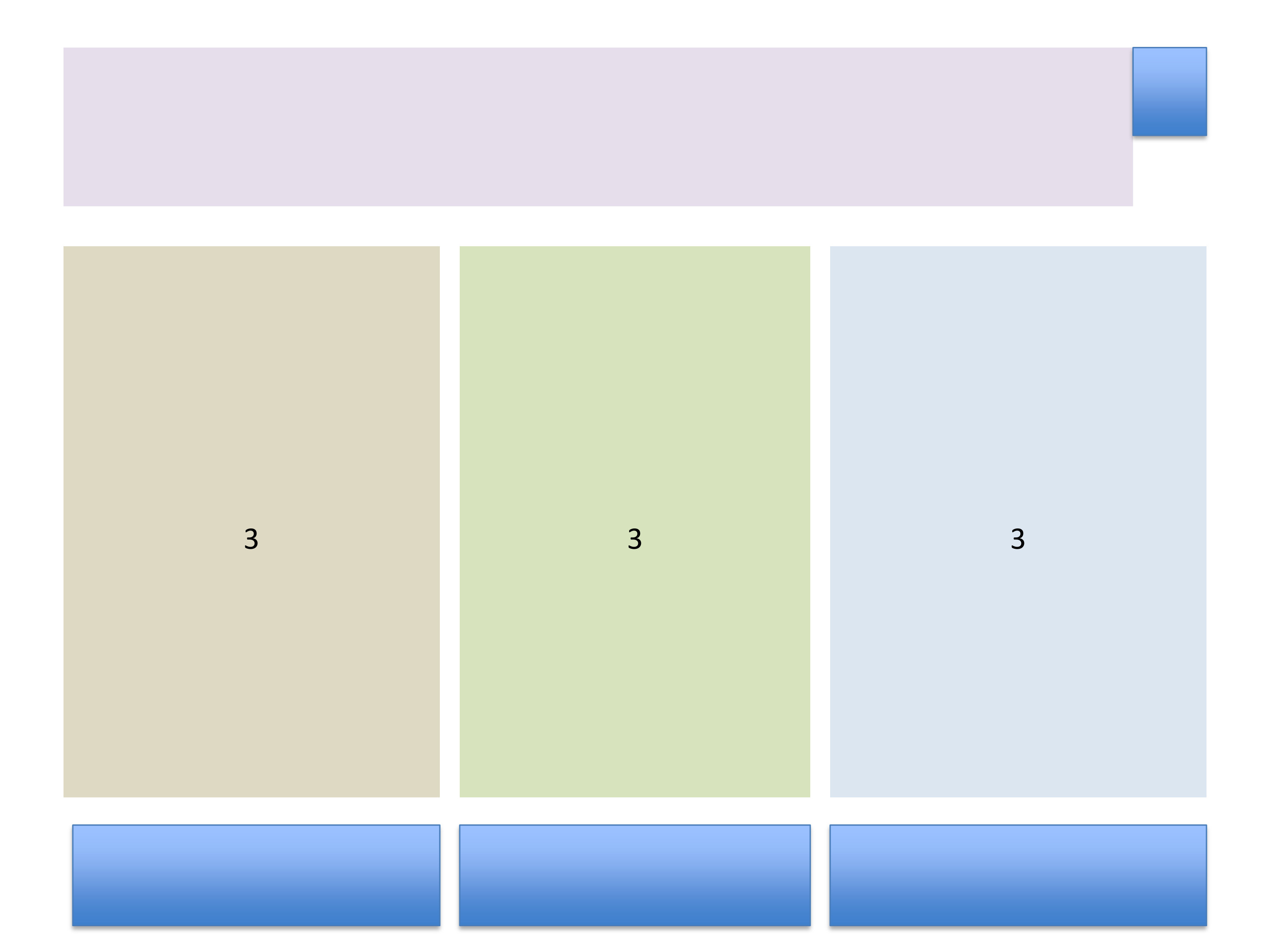}}} \\
     \subfloat[n = 3]
    {\fbox{\includegraphics[width=0.3\textwidth]{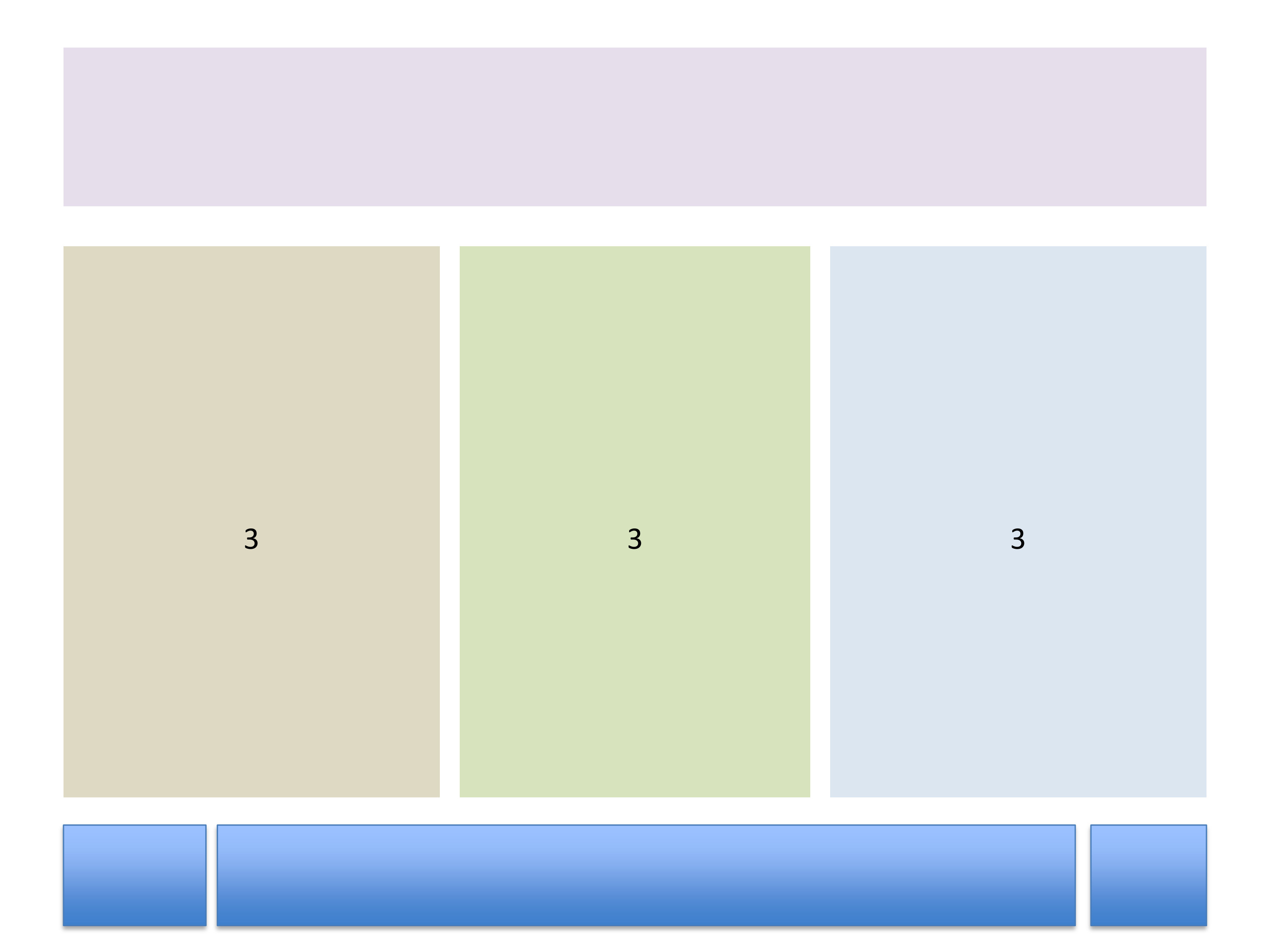}}}
     \subfloat[n = 3]
    {\fbox{\includegraphics[width=0.3\textwidth]{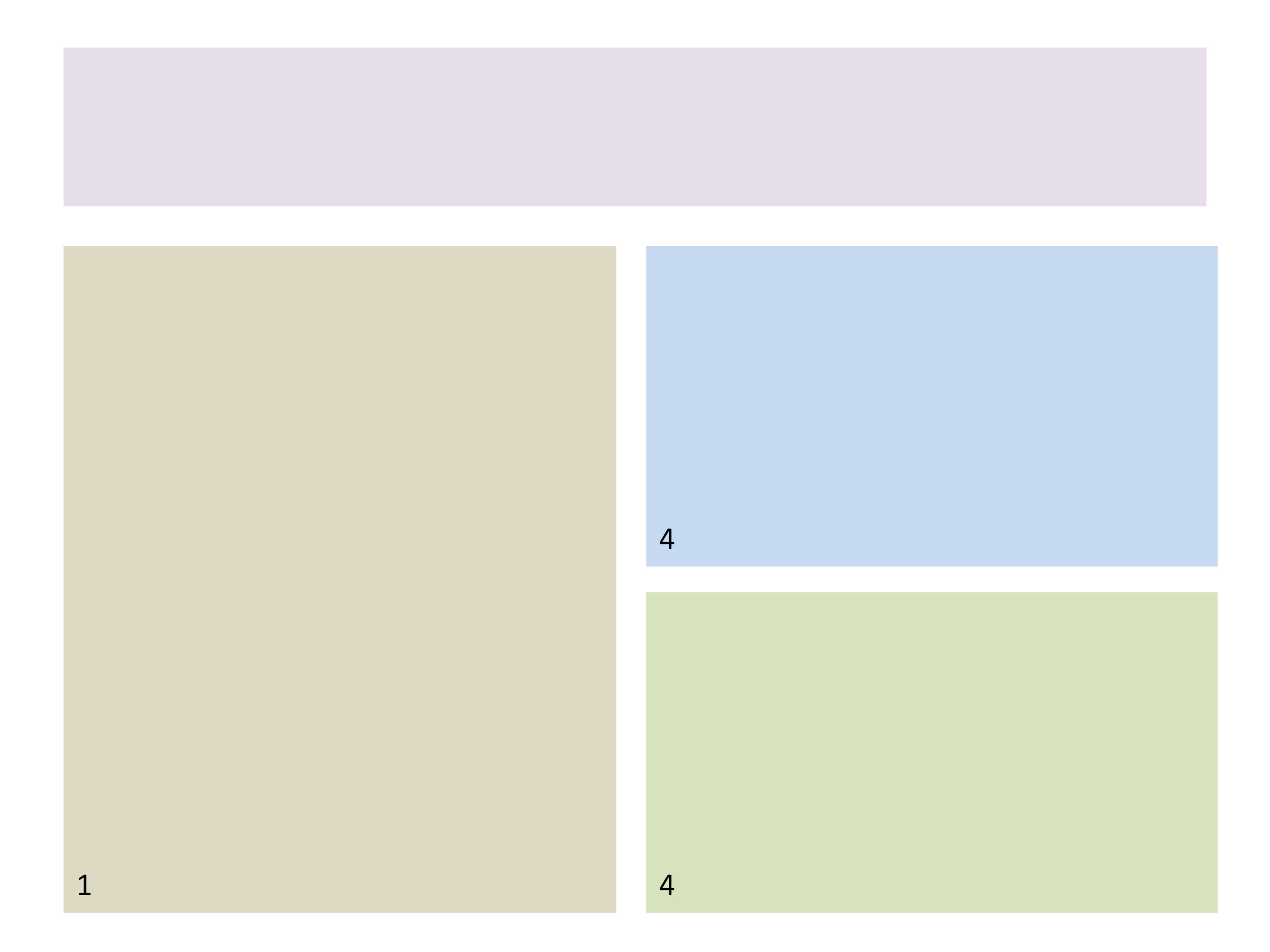}}}
         \subfloat[n = 3]
    {\fbox{\includegraphics[width=0.3\textwidth]{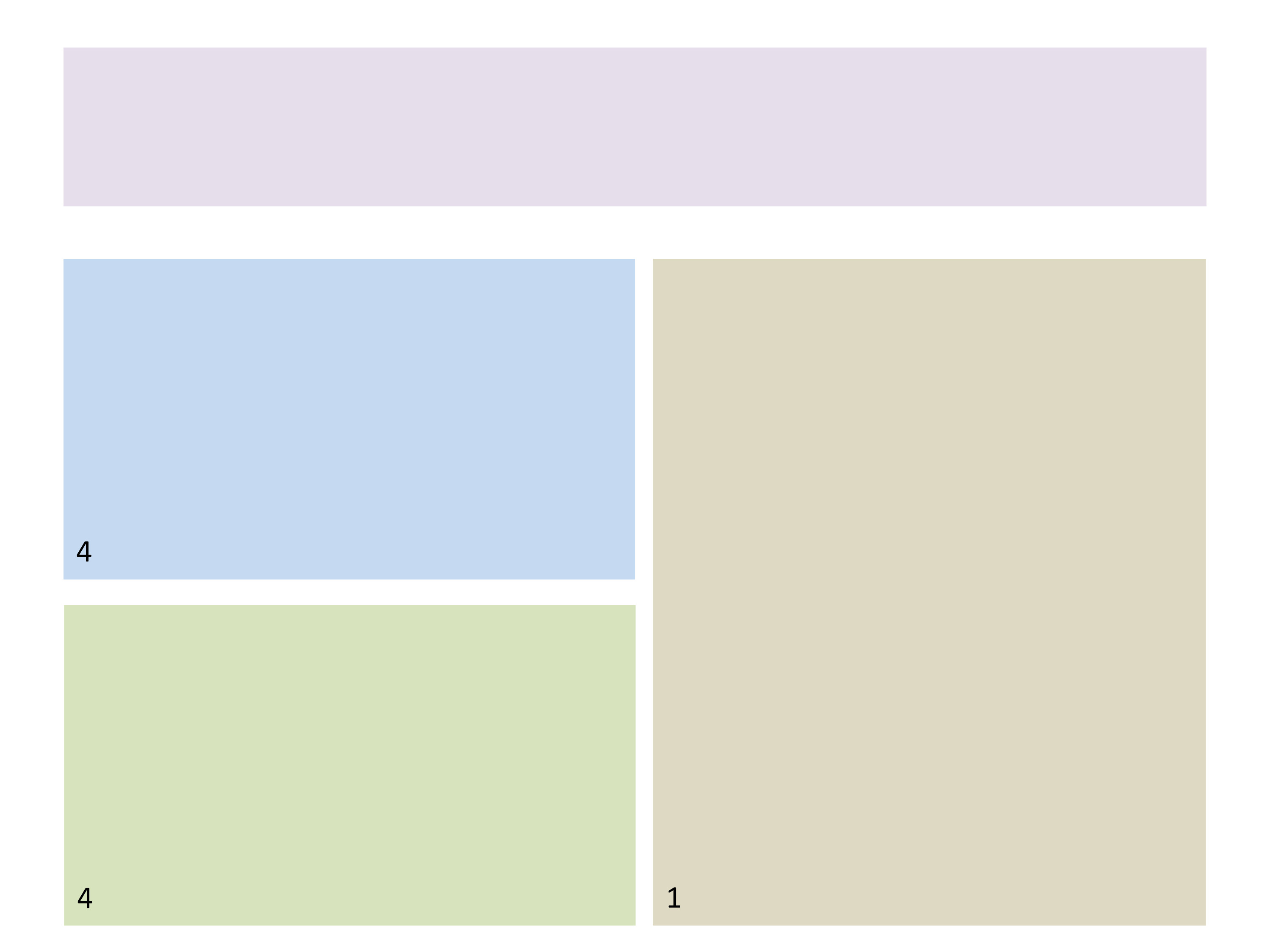}}} \\
         \subfloat[n = 3]
    {\fbox{\includegraphics[width=0.3\textwidth]{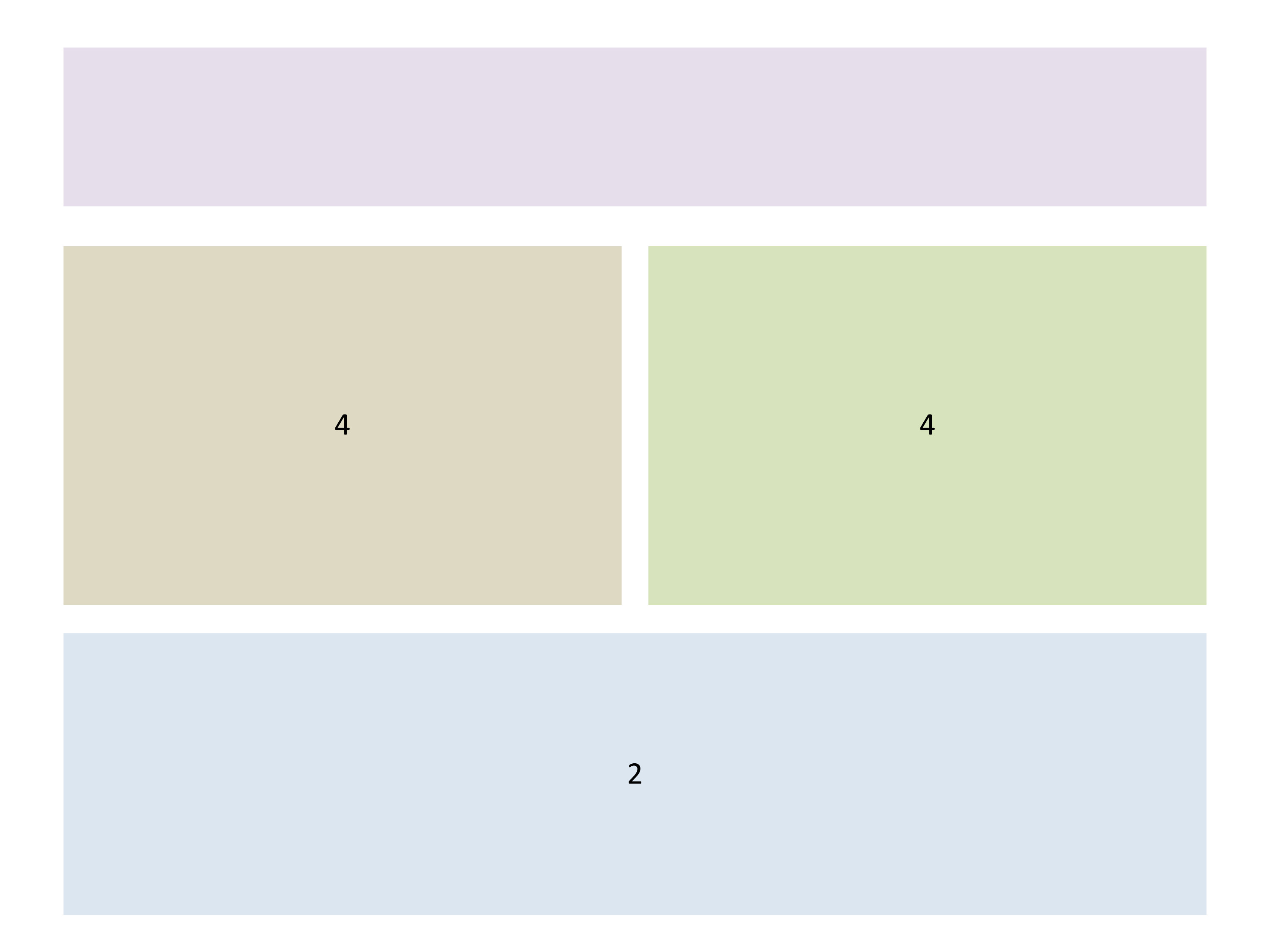}}}
         \subfloat[n = 3]
    {\fbox{\includegraphics[width=0.3\textwidth]{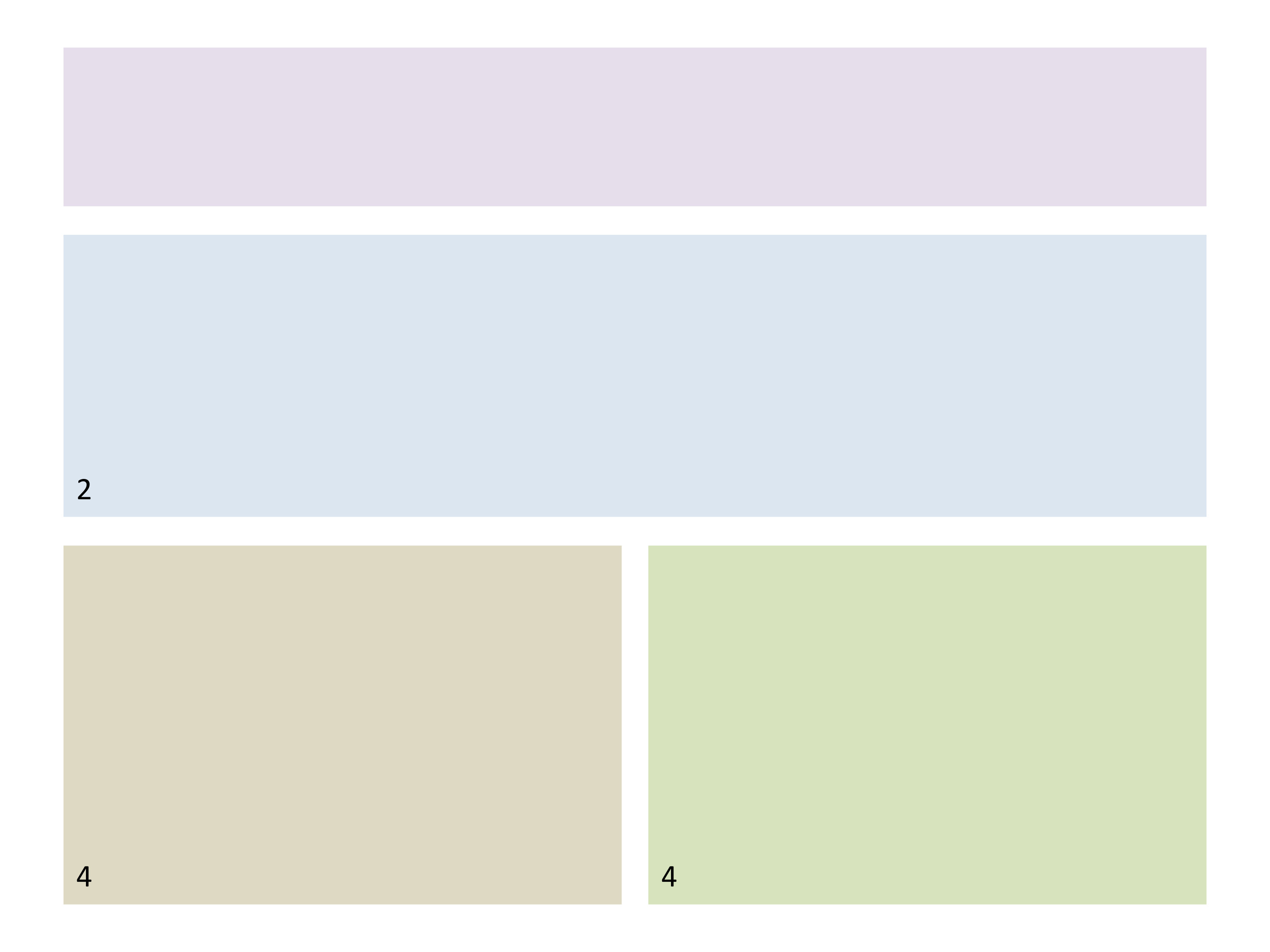}}}
         \subfloat[n = 4]
    {\fbox{\includegraphics[width=0.3\textwidth]{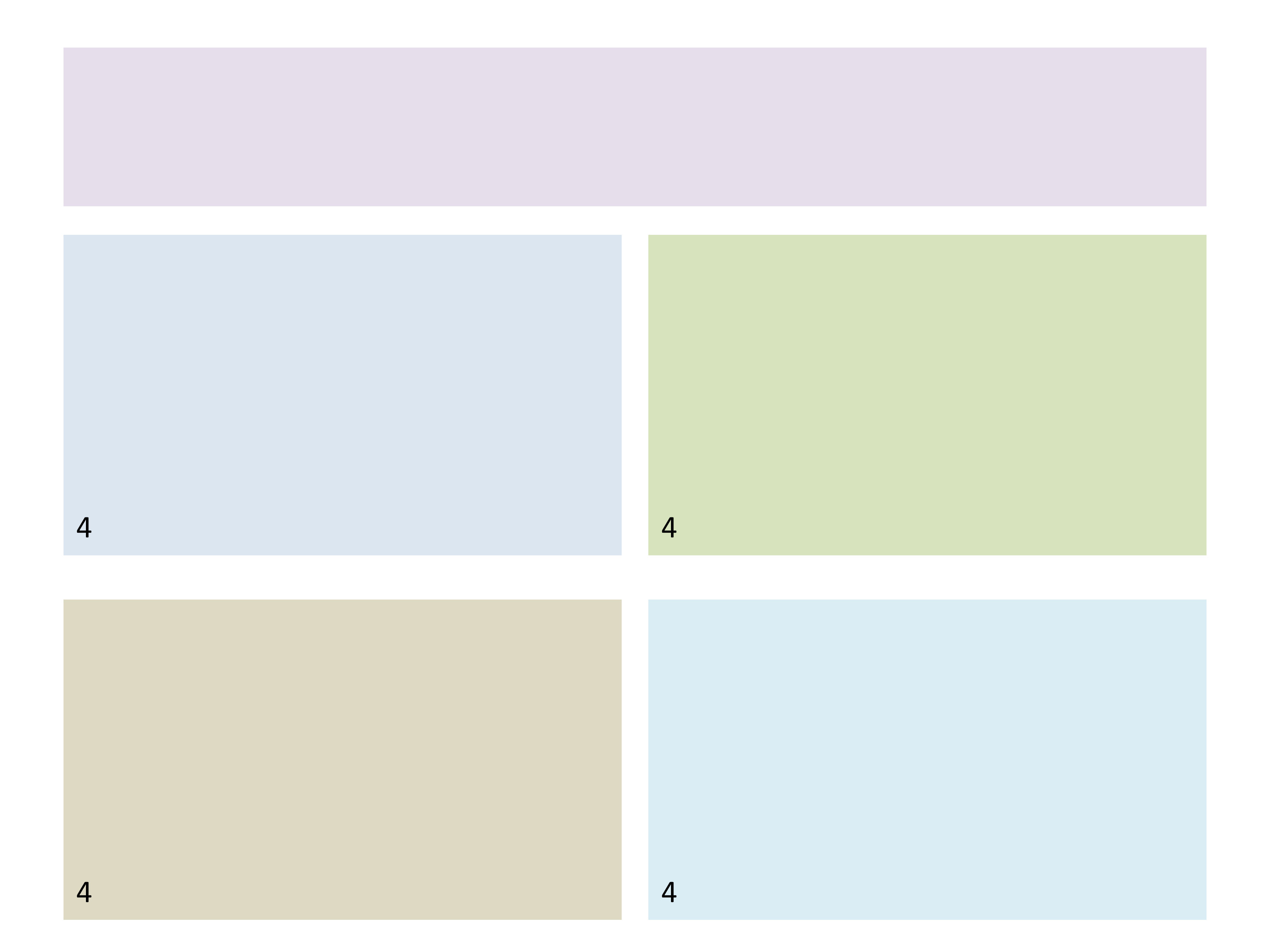}}} \\
         \subfloat[n = 4]
    {\fbox{\includegraphics[width=0.3\textwidth]{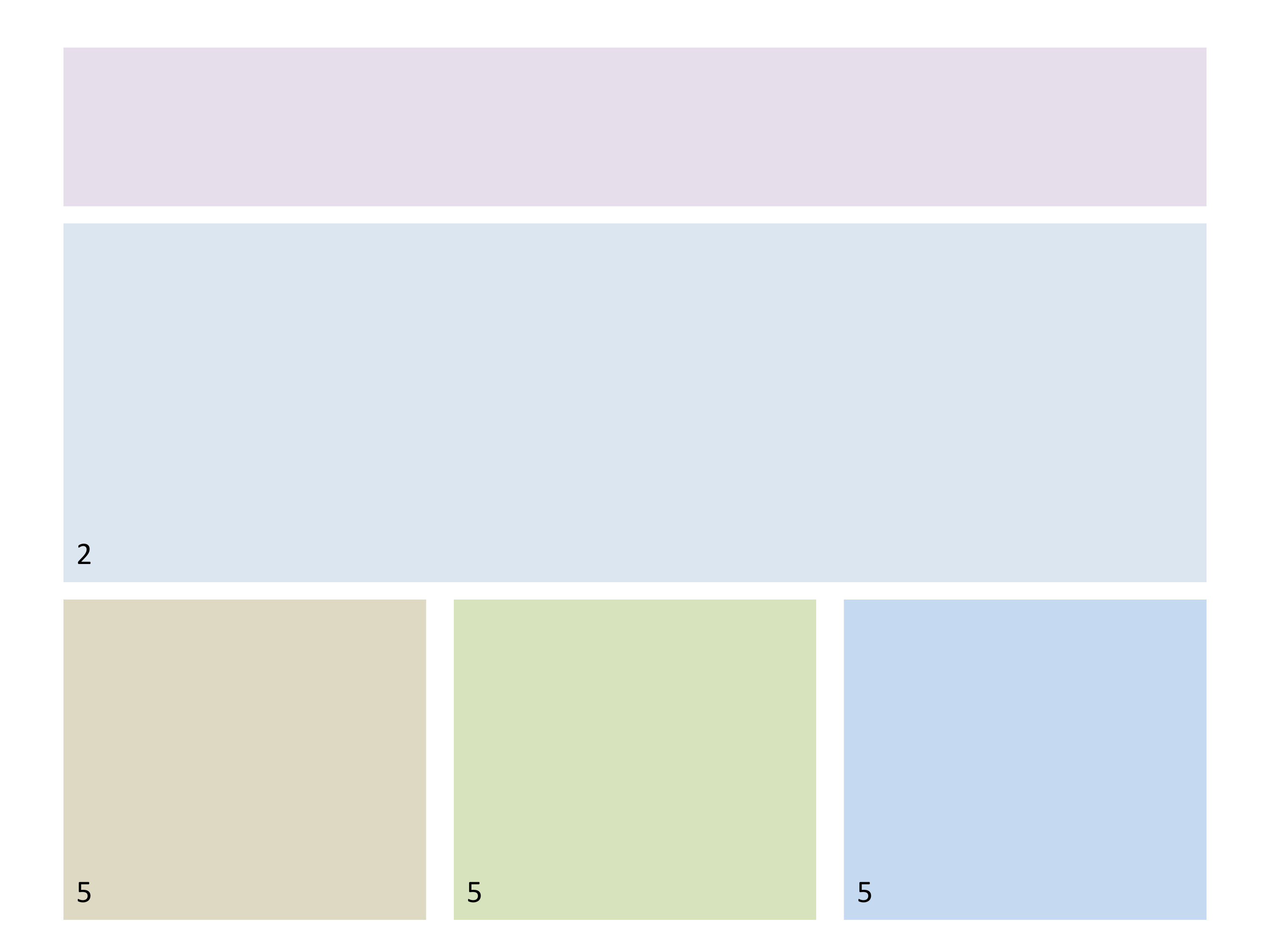}}}
         \subfloat[n = 4]
    {\fbox{\includegraphics[width=0.3\textwidth]{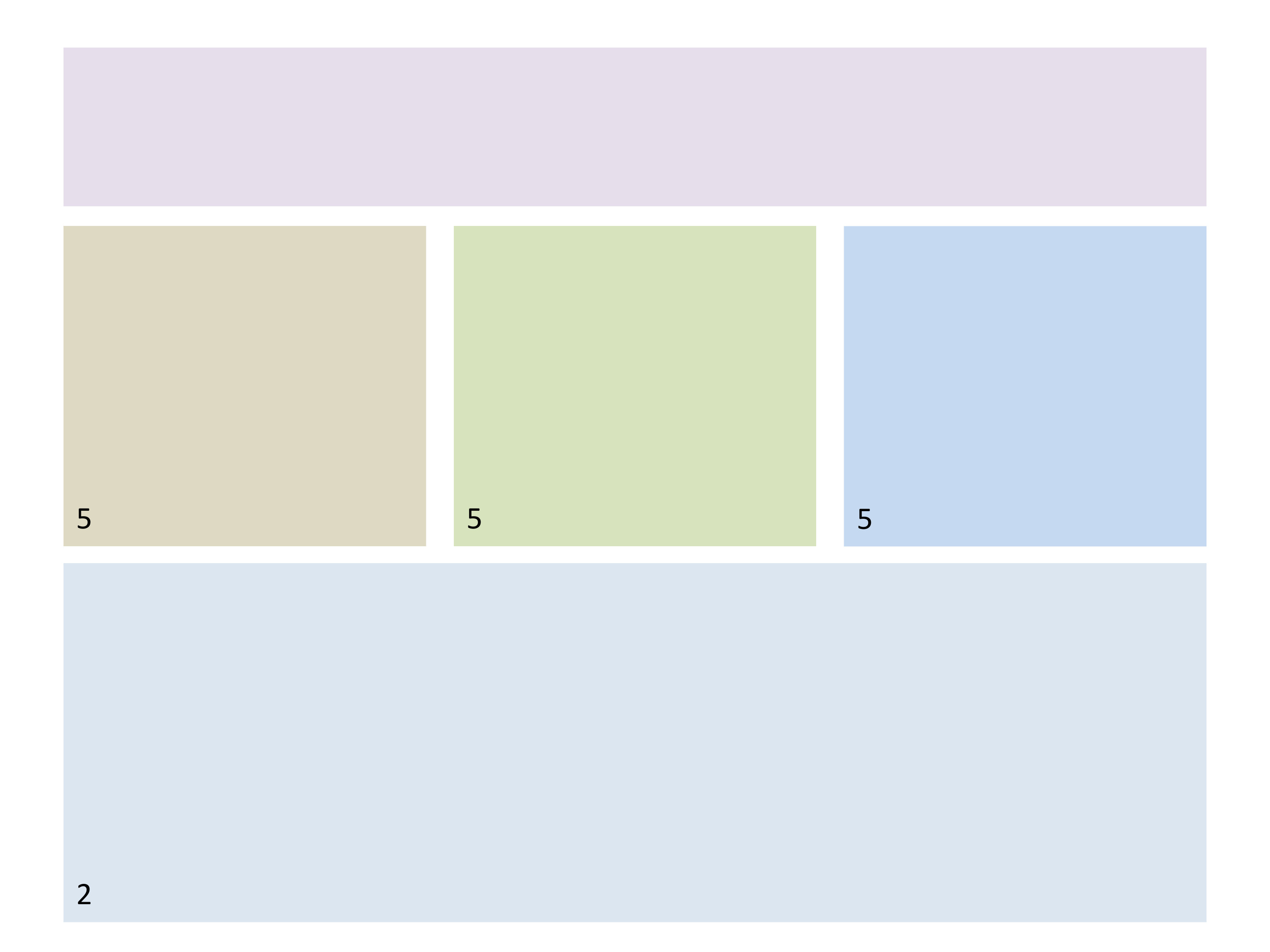}}}

    \caption{Base Layouts used for SynSlide Generation}
    \label{fig:grid1}
\end{figure}

To sum up, the Layout and Style discriminator module achieves the following:
\begin{itemize}
    \item Provides flexibility in slide design by using randomized layouts and styles.
    \item Ensures visual variety while maintaining structural coherence.
    \item Simplifies the transition from raw content to visually rendered presentations.
\end{itemize}

\newpage

\section{Limitations of SynSlideGen}

In this section, we discuss in detail the known limitations related to the \textbf{SynSlideGen} pipeline and the resulting \textbf{SynSlides} generated slides. Moreover, we provide data scaling results of \textbf{SynSlides} and its effect on downstream performance.

\subsection{Effect of static layouts}
As mentioned in Phase 2, we select one out of eighteen predefined layouts for slide generation. Hence, by design, each slide is constrained to have a maximum of 4 body elements apart from title, and other footer/meta elements. While this is not restrictive for business presentations that have limited element density per slide, lecture slides tend to show higher variability in number of body elements. Also, slide elements are significantly moved or resized often for better visibility or aesthetic appeal making them more unstructured than limited randomization performed in synthetic slides. This is evident from comparing element-wise spatial heatmaps of RealSlide benchmark and SynSlides dataset. Moreover, it has been noticed that lecture slides often use proprietary background templates  Hence, synthetic slides might lack generalization to out-of-distribution lecture slides with high density of content or non-traditional layouts. An important future contribution to the pipeline is to implement an efficient method to assign randomized non-static layouts while minimizing overlap between elements to ensure high quality of synthetic slides.

\subsection{Inadequate context while generation}

Current method of context seeding in SynSlideGen is through a series of prompts injected at several stages of generation. While this method provides more opportunity to capture wide context compared to prior methods that use a single prompt, it still falls short of achieving content depth found in real lecture slides. Comparing RealSlide and SynSlides we find that real lecture presentations have, on average, three times the number of slides as compared to SynSlides. This difference directly affects content quality where real lecture presentations are found to be more intricate and advanced lecture content. We restrict synthetic slides to around 12 - 15 per presentation thereby effectively limiting the depth of content that can be covered per topic. Such biases may hinder use of SynSlides for other high-level document tasks like Visual Question Answering.

\subsection{Separate annotation functions}

SynSlideGen provides automatic generation and annotation for two downstream tasks namely Slide Element Detection and Text-based Slide Image Retrieval. However, post generation of slide images, they annotation pipeline is required to be separate for each task. Hence, while the raw JSON content for each synthetic slide can be reused for multiple tasks, there requires additional effort to scale up number of downstream tasks for the same data.

\subsection{Scaling synthetic slides}
Our experiments with increasing synthetic slides for Slide Element Detection show that performance gains are marginal (see Table~\ref{result}) and hence we opt to choose an optimal size set that balances performance and training time.

\begin{table}[h]
\centering
\begin{tabular}{|c|c|c|}\hline
\# Synthetic images &\multicolumn{2}{c|}{Performance on }\\
for Training &\multicolumn{2}{c|}{RealTest (750 images)}\\
\cline{2-3}
 &Synthetic &Synthetic + Real \\\hline 
500 & 27.8 & 35.2 \\
1000 & 28.1 & 36.4 \\
2200* & 29.8 & 38.8 \\
4000 & 30.2 & 38.8 \\
8000 & 30.7 & 39.7 \\
16000 & 30.9 & 40.3 \\
\hline
\end{tabular}
\vspace{0.01\textwidth}
\caption{Synthetic: model trained on synthetic images only. Synthetic + Real: model trained on synthetic, then 300 real images in two-stage finetuning setting.
*Size refers to \textbf{SynDet}.}\label{result}
\end{table}

\begin{figure}[htbp]
    \centering
        \subfloat[Title]
        {\fbox{\includegraphics[width=\linewidth]{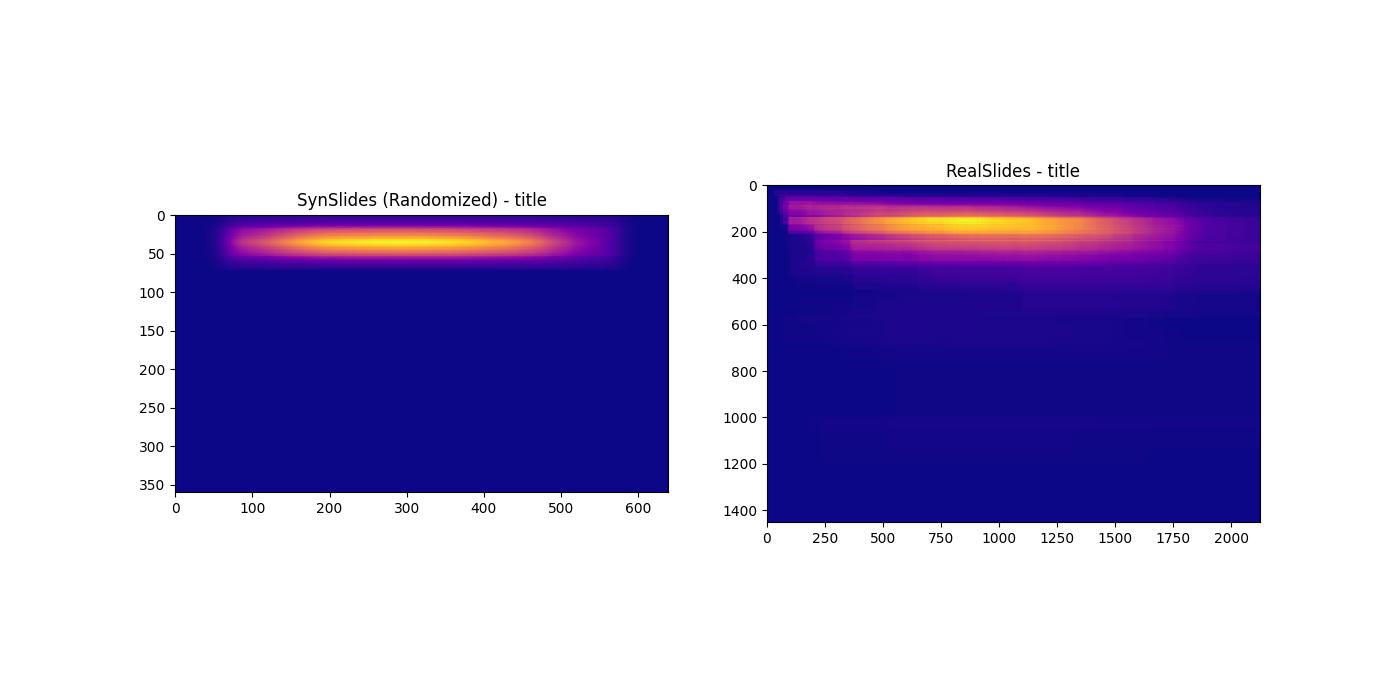}}} \\ 
        \subfloat[Chart]
        {\fbox{\includegraphics[width=\linewidth]{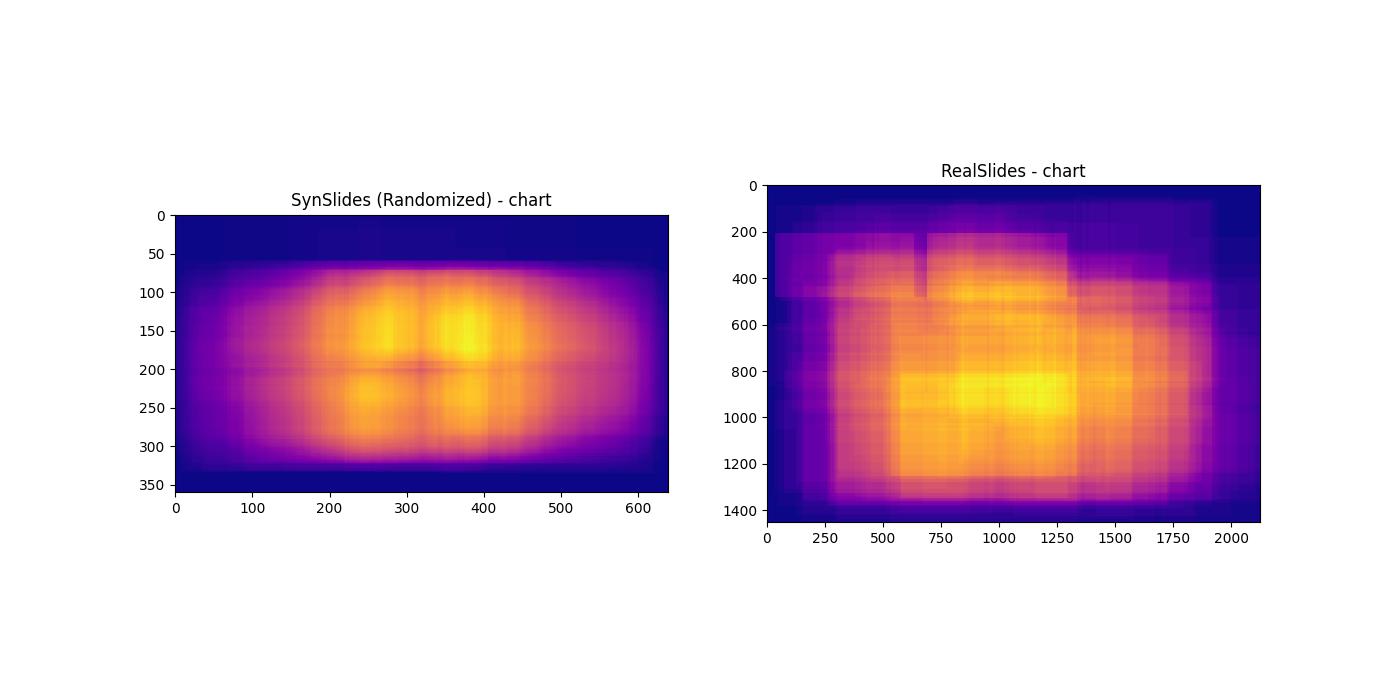}}} \\
       \subfloat[Table Caption]
        {\fbox{\includegraphics[width=\linewidth]{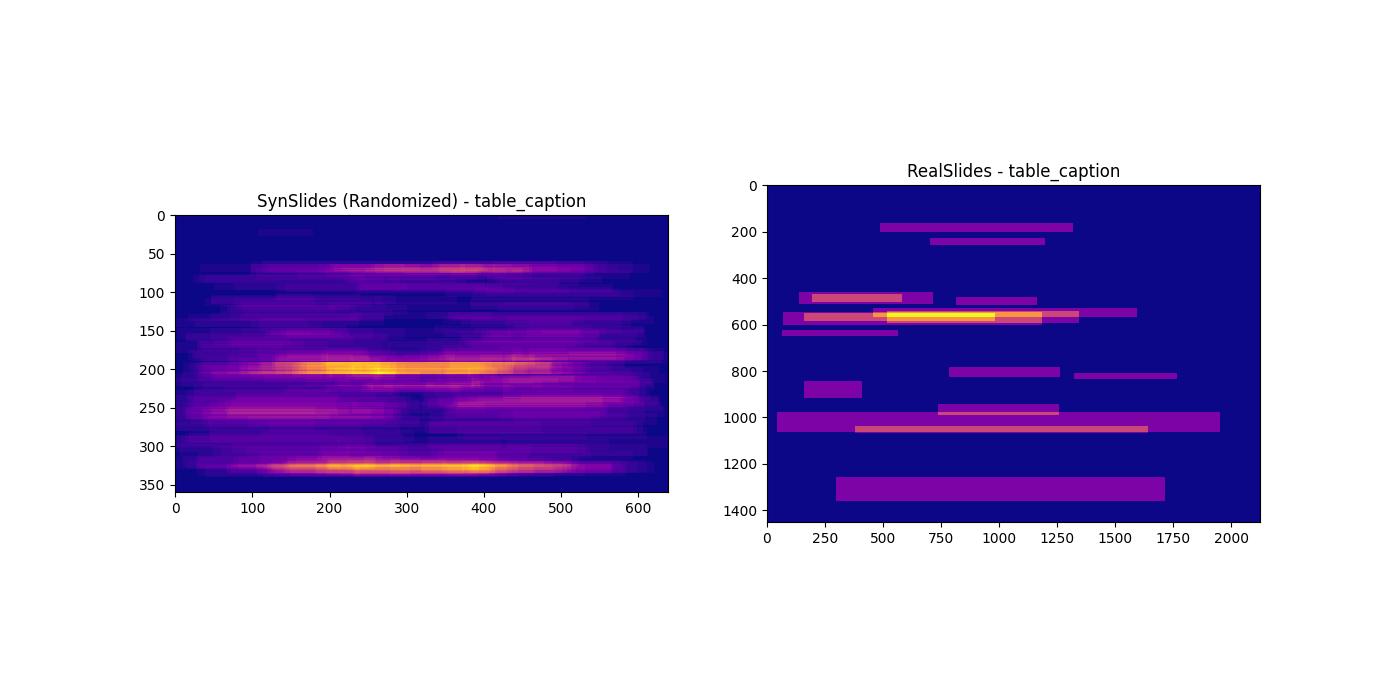}}} 

    \caption{Layout Heatmaps comparing randomized SynSlides (Left) with RealSlide layouts (Right). }
    \label{fig:}
\end{figure}

\begin{figure}[htbp]
    \centering
       \subfloat[Table]
        {\fbox{\includegraphics[width=\linewidth]{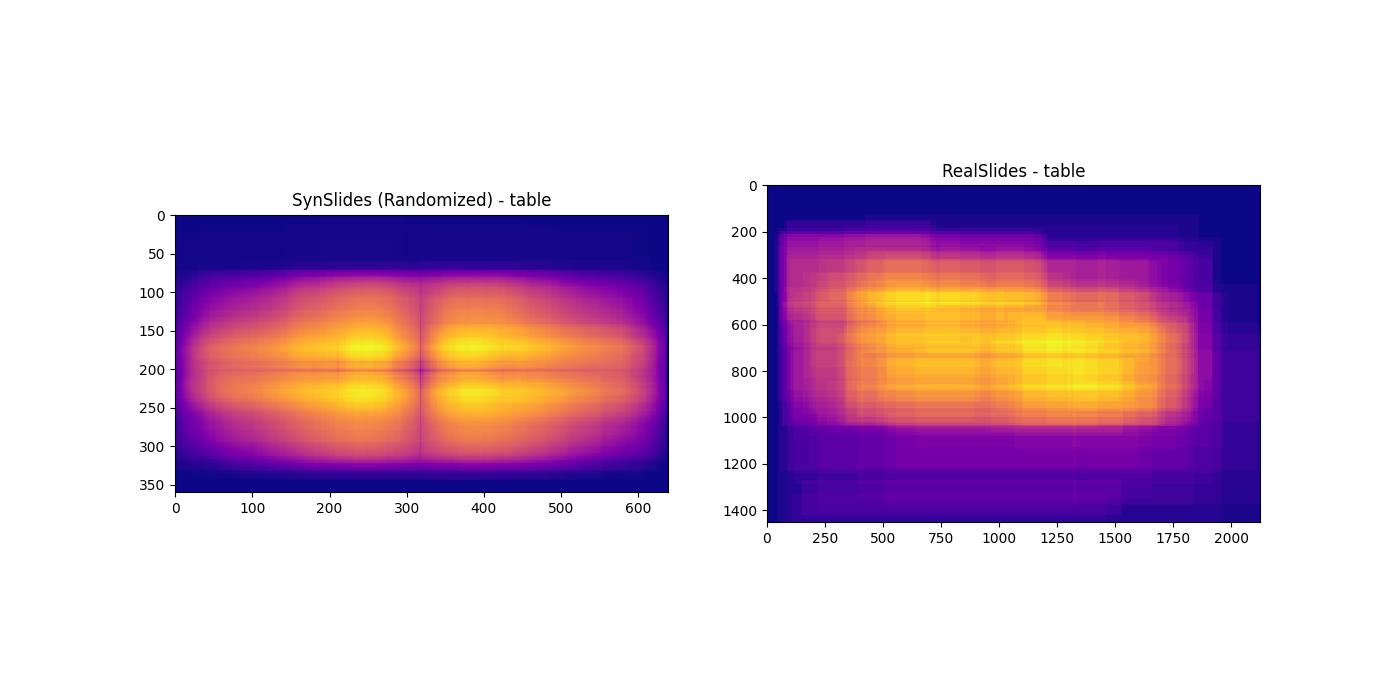}}} \\
       \subfloat[Text]
        {\fbox{\includegraphics[width=\linewidth]{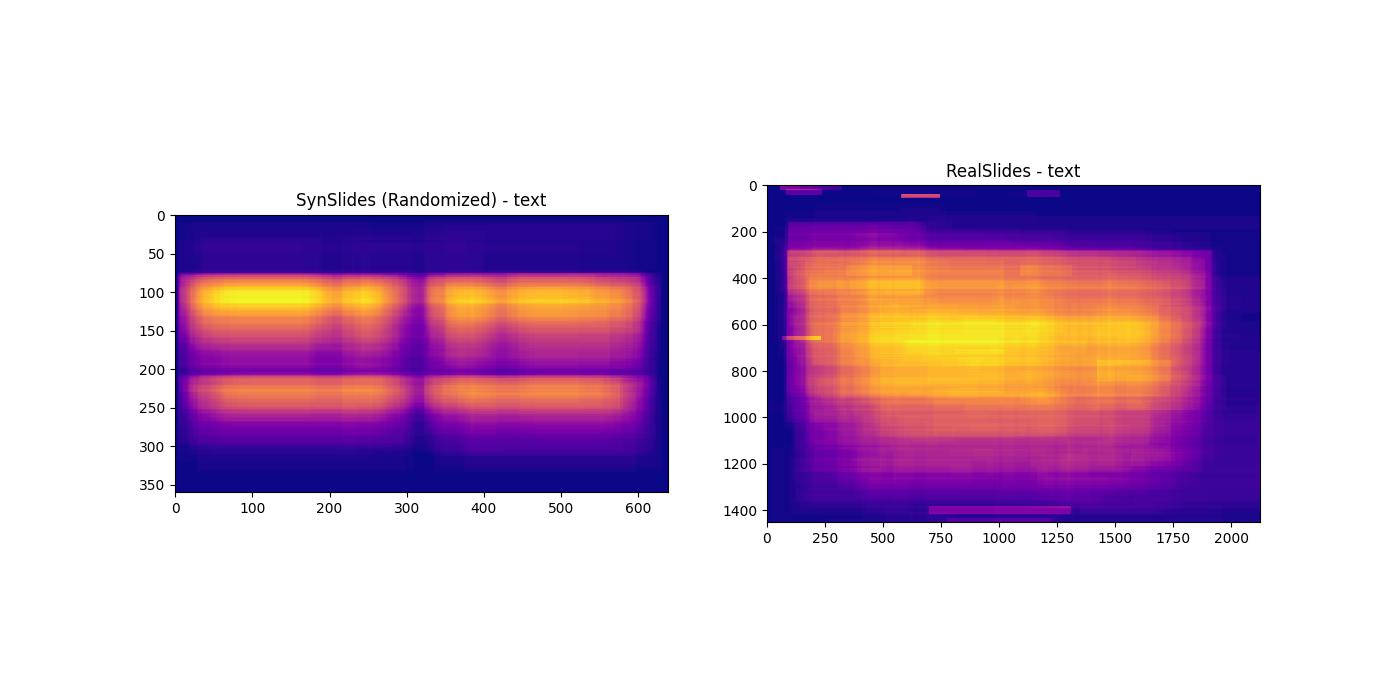}}} \\
       \subfloat[Equation]
        {\fbox{\includegraphics[width=\linewidth]{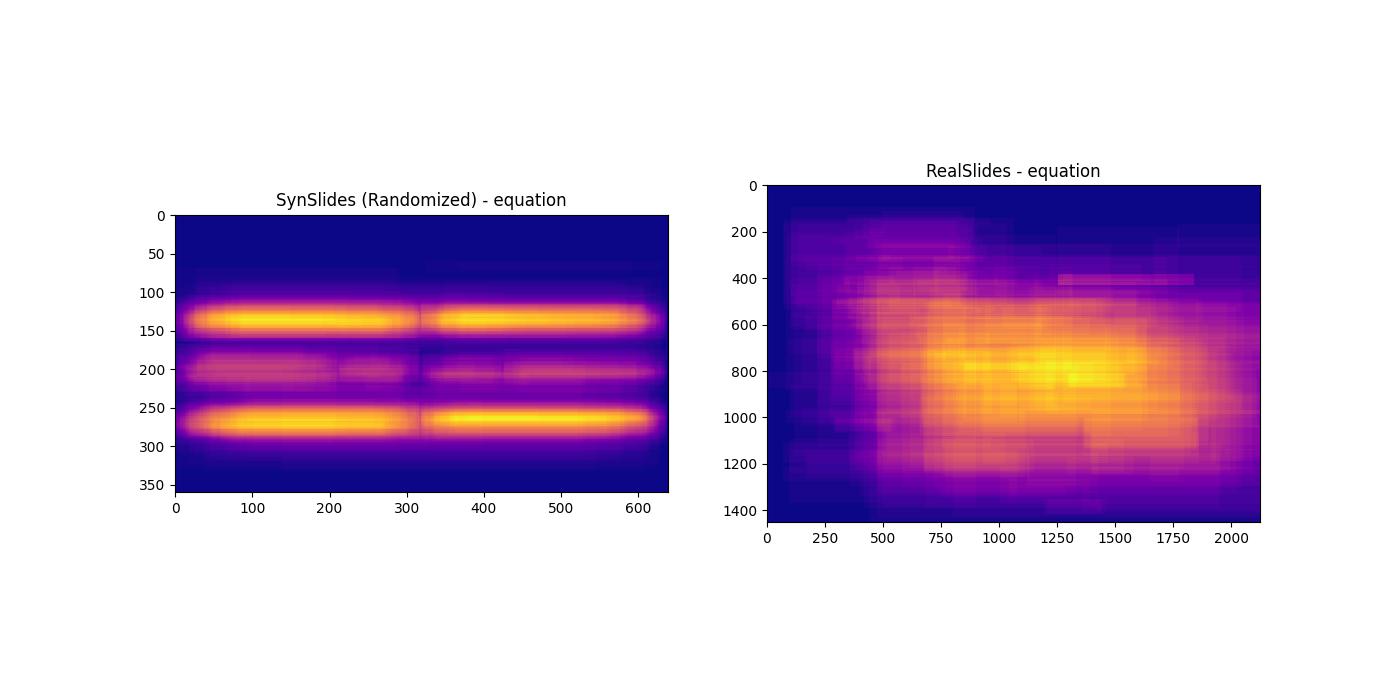}}} 
    \caption{Layout Heatmaps comparing randomized SynSlides (Left) with RealSlide layouts (Right). }
\end{figure}

\newpage
\section{Discussion on Choice of Classes for Slide Element Detection}

We define 16 classes for the task by merging and pruning classes from the 25-class SPaSe dataset. Specifically, we merge: Date, Footnote, Affiliation into Footer-Element; Presentation-Title and Slide-Title into Title; and Screenshot and Realistic Image into Image. We prune Map, Handwritten Equation, Comment, Drawing, and Legend due to insufficient instances in pre-made lecture slides. We determined the 16 classes based on their prevalence in lecture presentations, while also ensuring they are fine-grained enough for tasks such as slide narration systems.

\section{Visual Sample and Results}
Here we present additional figures and tables that supplement the main text. 

\begin{figure}[!h]
\centerline{
\fbox{\includegraphics[width=0.45\textwidth]{1117322.png}}
\hspace{0.001\textwidth}
\fbox{\includegraphics[width=0.45\textwidth]{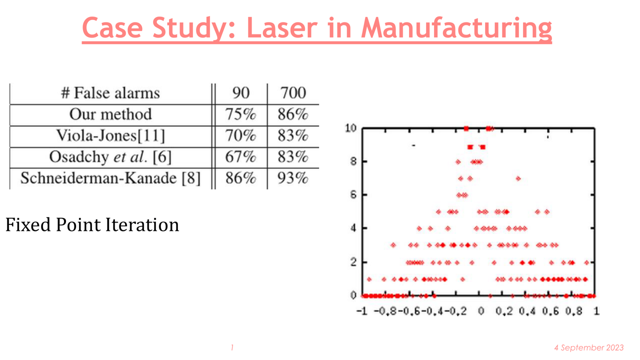}}}
\vspace{0.001\textwidth}
\centerline{
\fbox{\includegraphics[width=0.45\textwidth]{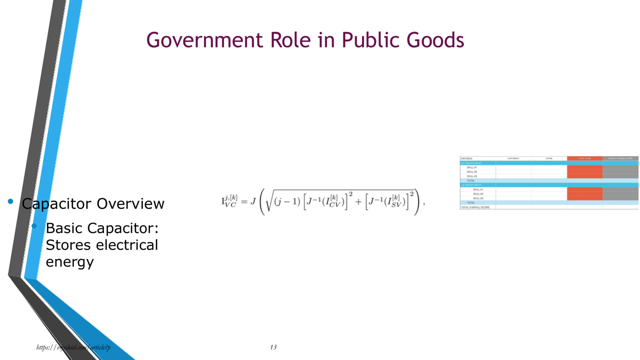}} 
\hspace{0.001\textwidth}
\fbox{\includegraphics[width=0.45\textwidth]{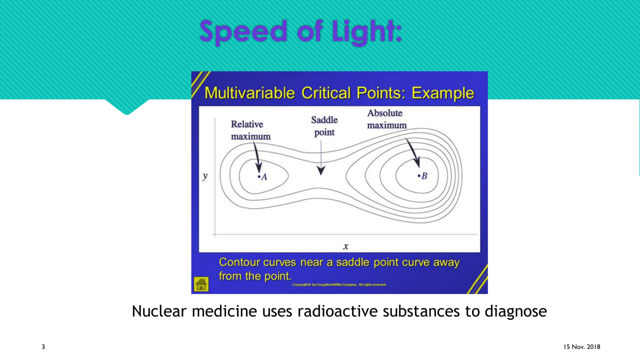}}}
\vspace{0.001\textwidth}
\centerline{
\fbox{\includegraphics[width=0.45\textwidth]{113663165.png}}
\hspace{0.001\textwidth}
\fbox{\includegraphics[width=0.45\textwidth]{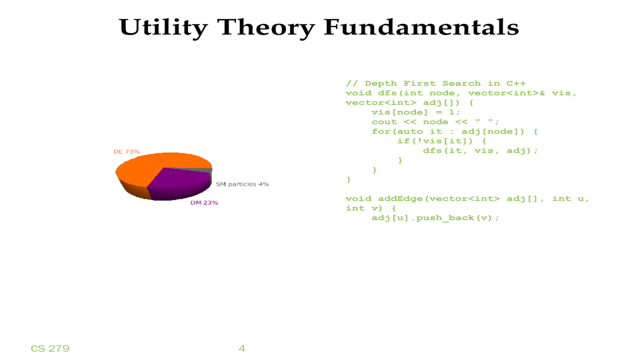}}}
\caption{\textbf{SynDet} Visual Examples (Semantically Non Coherent Slides but Visually closer to Real Slides)}
\end{figure}

\begin{figure}[!h]
\centerline{
\fbox{\includegraphics[width=0.45\textwidth]{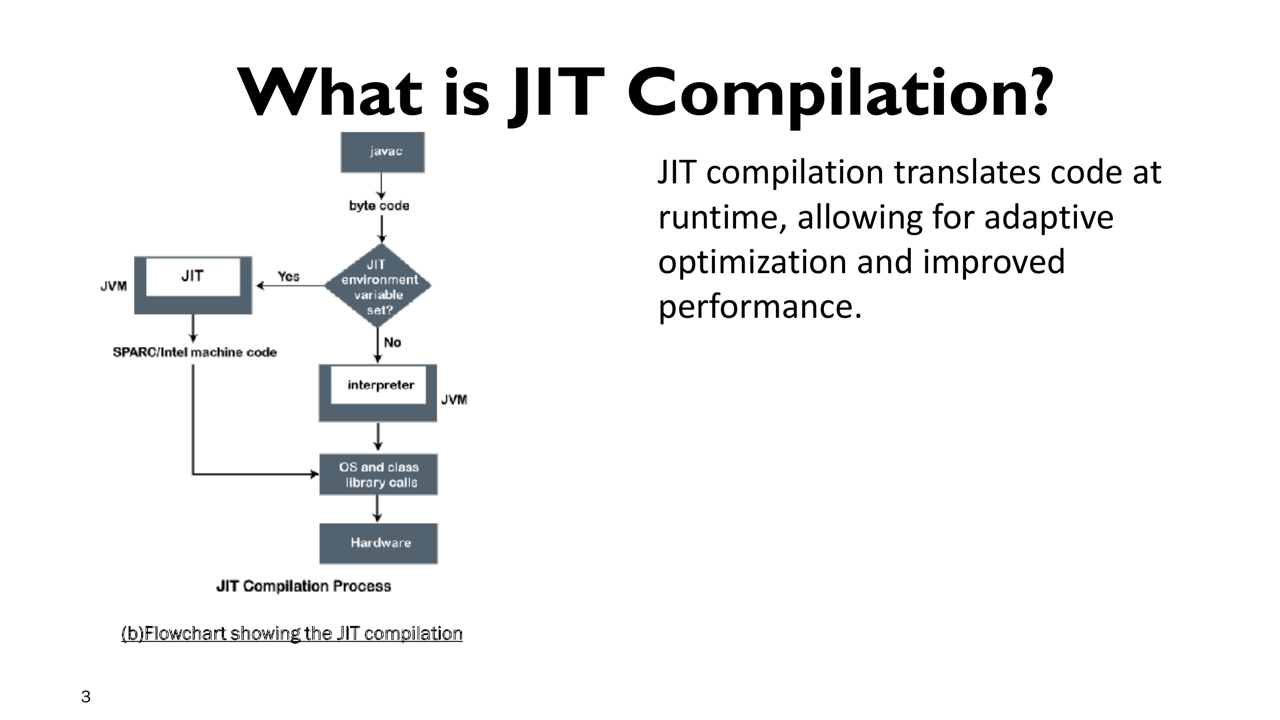}}
\hspace{0.001\textwidth}
\fbox{\includegraphics[width=0.45\textwidth]{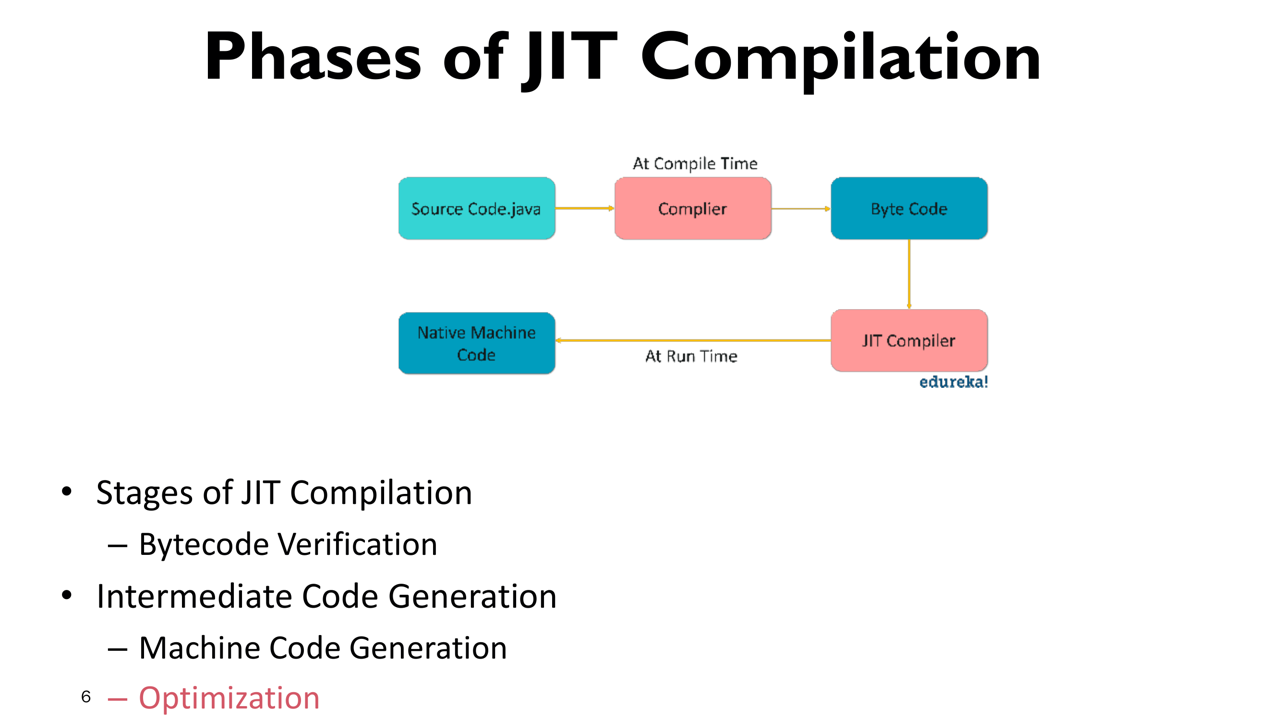}}}
\vspace{0.001\textwidth}
\centerline{
\fbox{\includegraphics[width=0.45\textwidth]{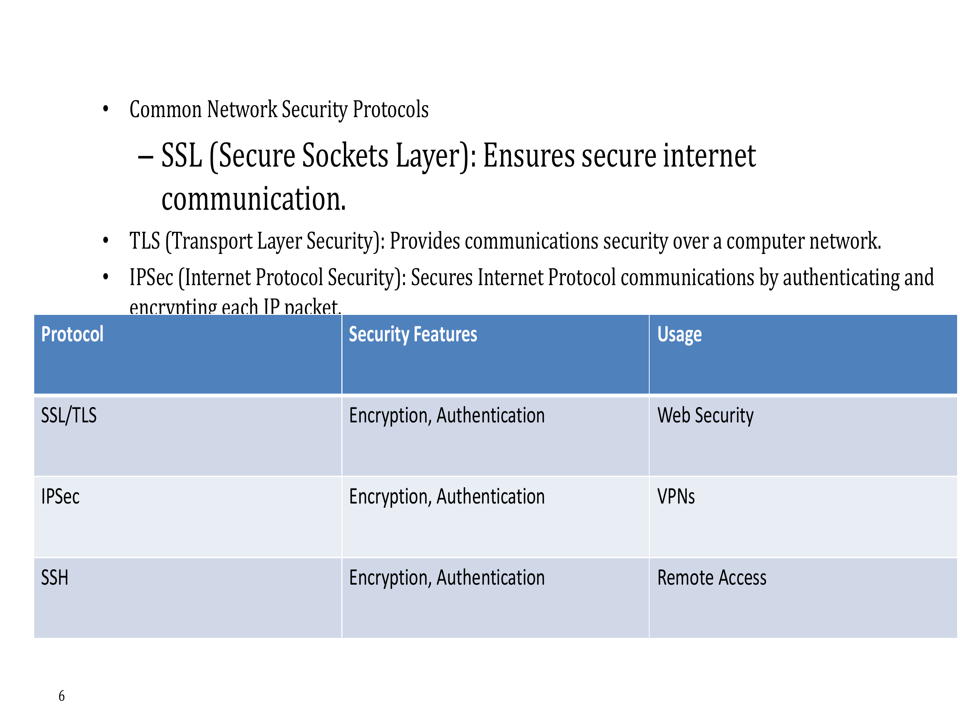}}
\hspace{0.001\textwidth}
\fbox{\includegraphics[width=0.45\textwidth]{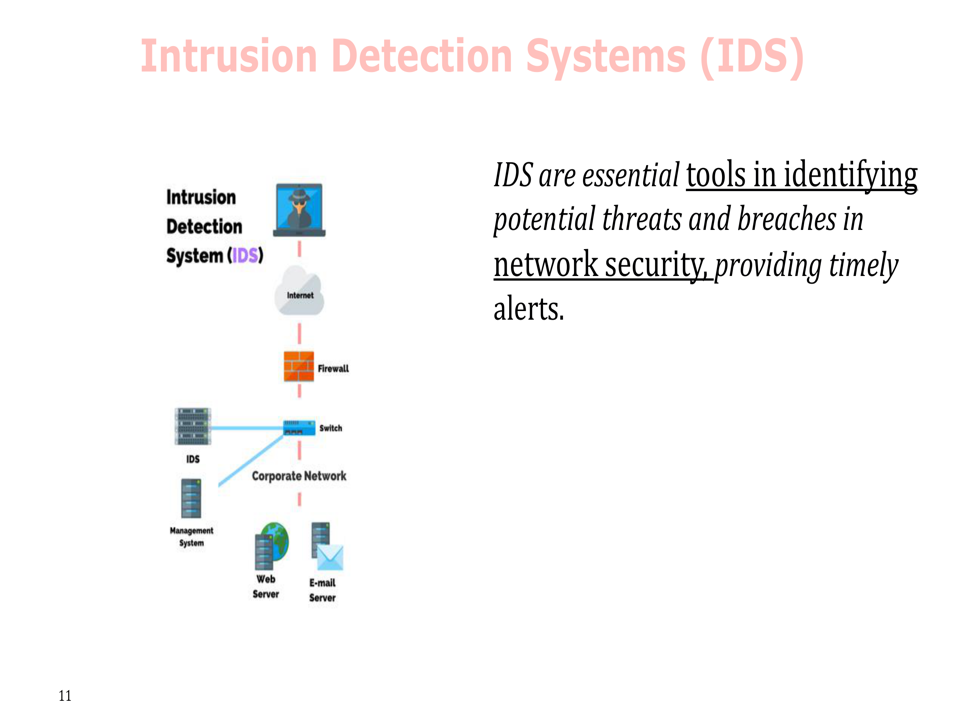}}}
\vspace{0.001\textwidth}
\centerline{
\fbox{\includegraphics[height=0.3\textwidth,width=0.45\textwidth]{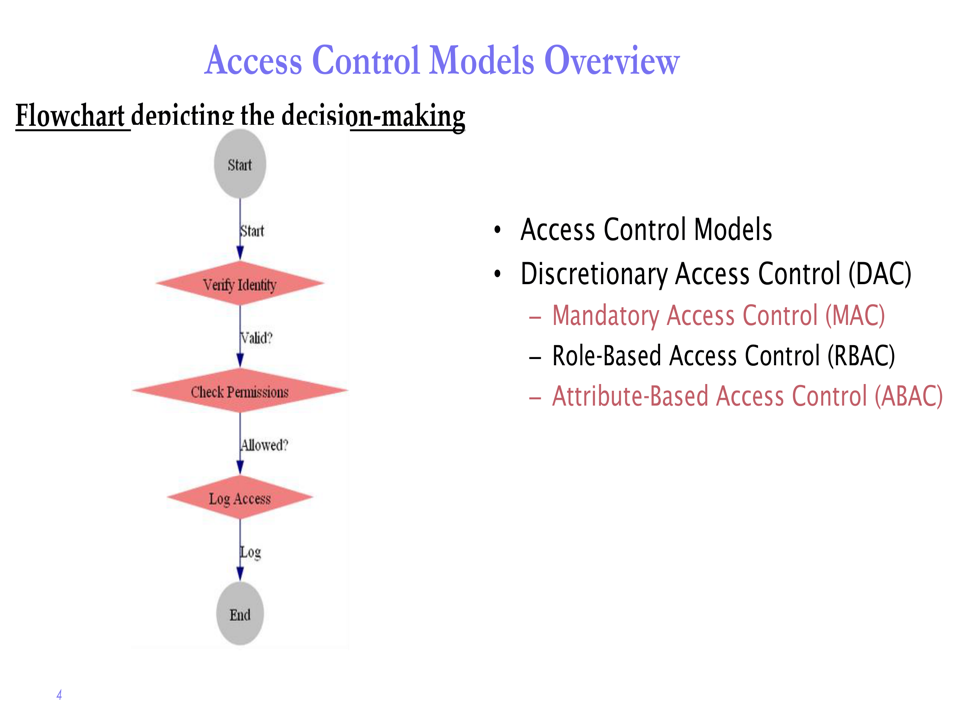}}
\hspace{0.001\textwidth}
\fbox{\includegraphics[height=0.3\textwidth,width=0.45\textwidth]{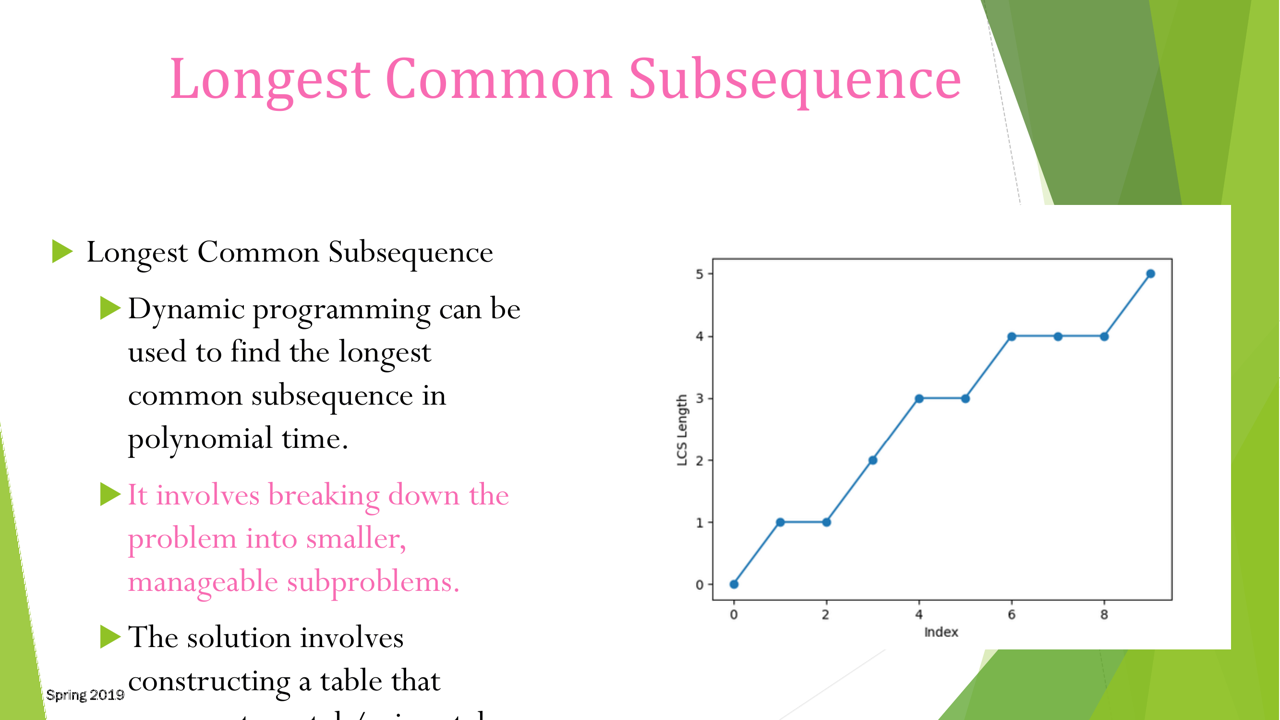}}}
\caption{\textbf{SynRet} Visual Examples (Coherent Slides)}
\end{figure}
\begin{figure}[!h]
\centerline{
\fbox{\includegraphics[width=0.45\textwidth]{213822282.png}} 
\hspace{0.001\textwidth}
\fbox{\includegraphics[width=0.45\textwidth]{22786033.png}}}
\vspace{0.001\textwidth}
\centerline{
\fbox{\includegraphics[width=0.45\textwidth]{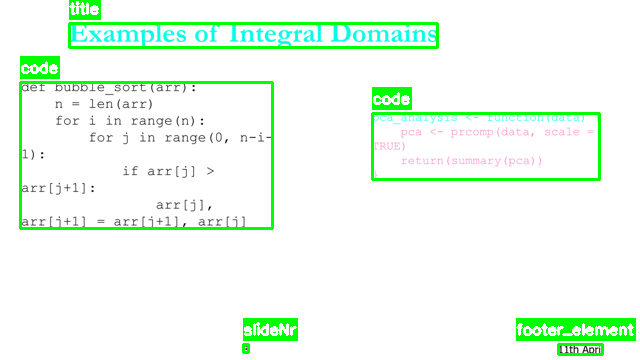}} 
\hspace{0.001\textwidth}
\fbox{\includegraphics[width=0.45\textwidth]{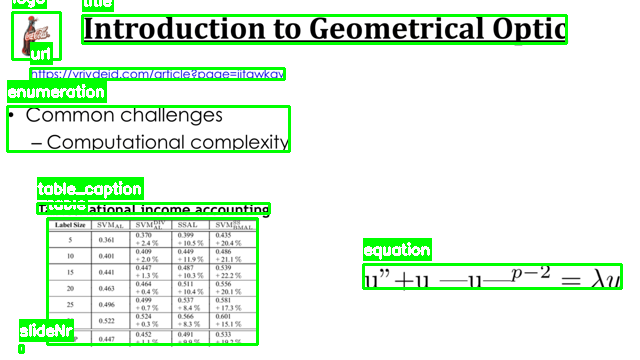}}} 
\caption{Ground truth annotations of sample SynDet slide images}
\end{figure}

\begin{figure}[!h]
\centerline{
\fbox{\includegraphics[width=1\textwidth]{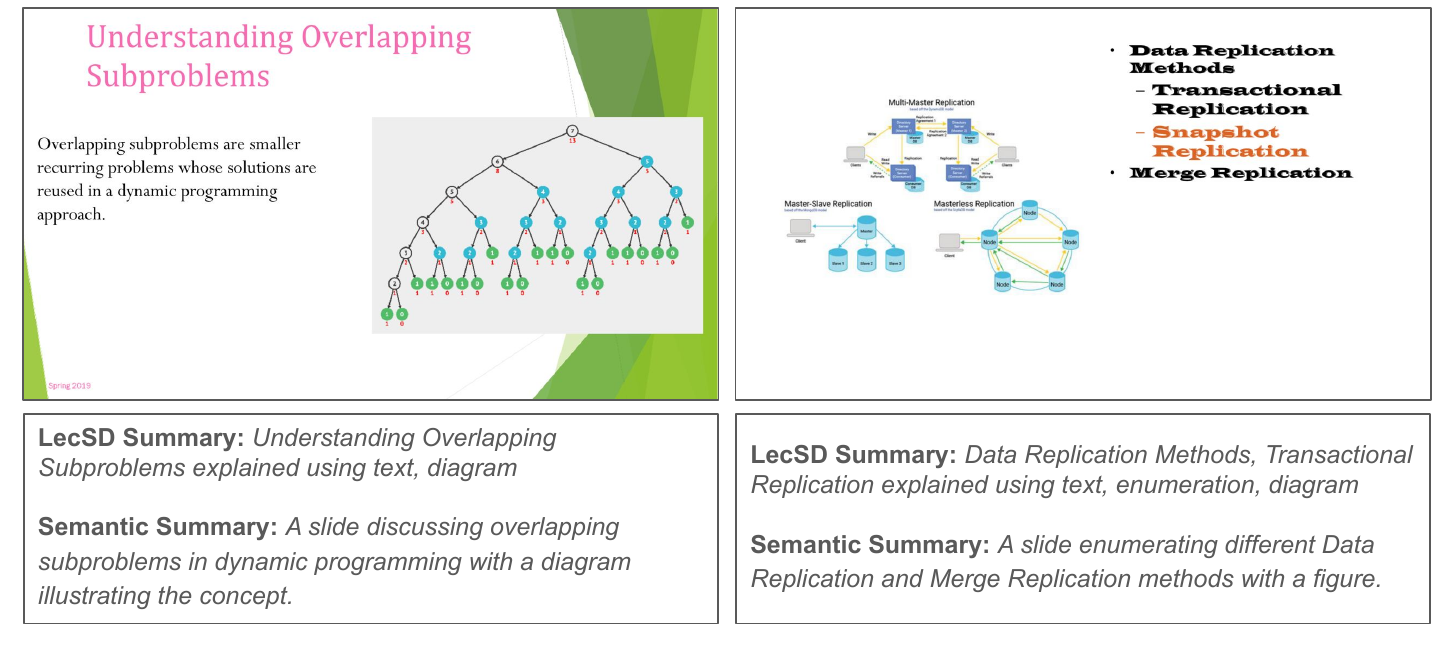}}}
\caption{Illustrate \textit{semantic summary} and \textit{LecSD-style summary} for two randomly selected sampled slides from our \textbf{SynRet}.}\label{fig:sem_summary} 
\end{figure}


\begin{figure}[!t]
\centerline{
\fbox{\includegraphics[width=0.4\textwidth]{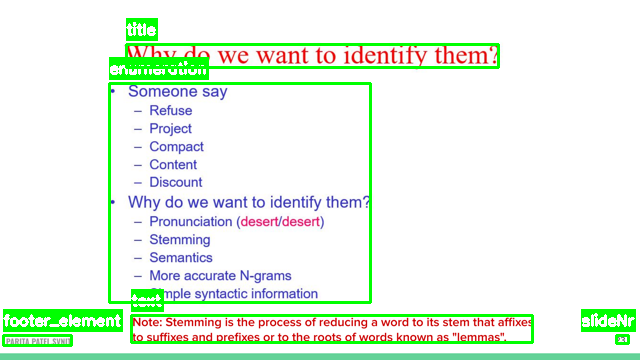}}
\hspace{0.001\textwidth}
\fbox{\includegraphics[width=0.4\textwidth]{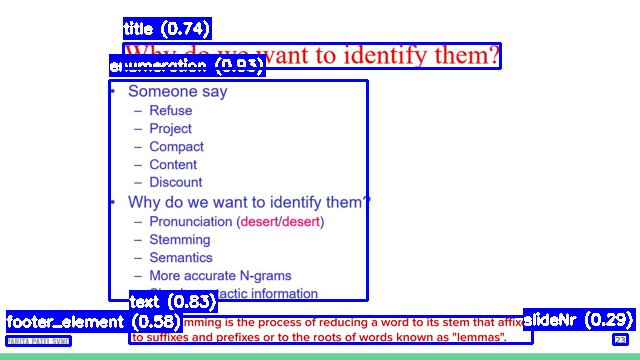}}}
\vspace{0.001\textwidth}
\centerline{    
\fbox{\includegraphics[width=0.4\textwidth]{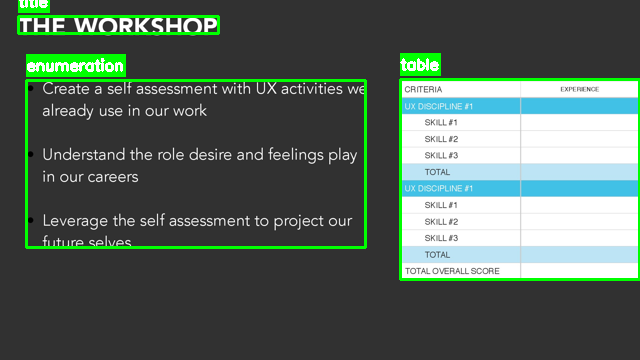}} 
\hspace{0.001\textwidth}
\fbox{\includegraphics[width=0.4\textwidth]{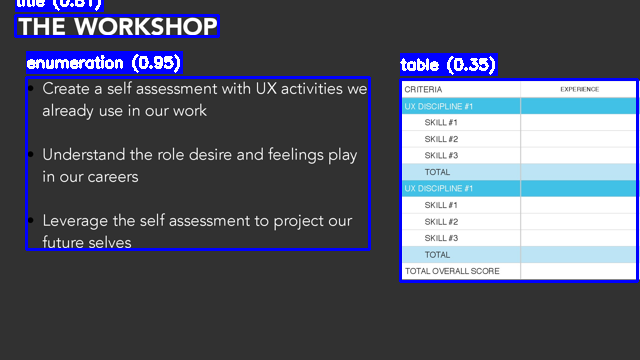}}} 
\vspace{0.001\textwidth}
\centerline{ 
\fbox{\includegraphics[width=0.4\textwidth]{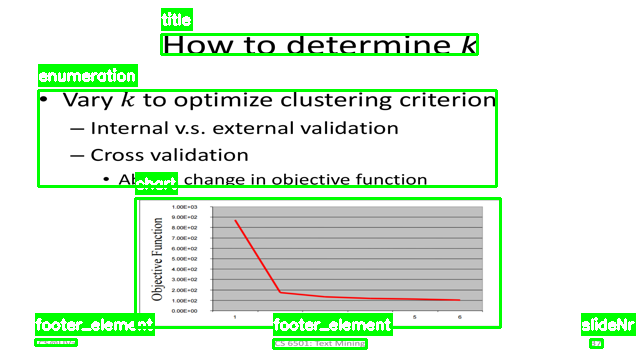}}  
\hspace{0.001\textwidth}
\fbox{\includegraphics[width=0.4\textwidth]{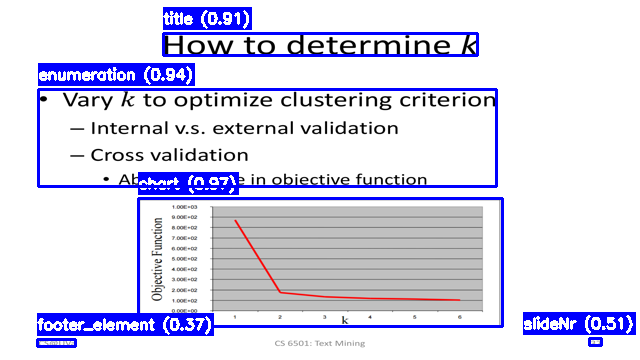}}} 
\vspace{0.001\textwidth}
\centerline{ 
\fbox{\includegraphics[width=0.4\textwidth]{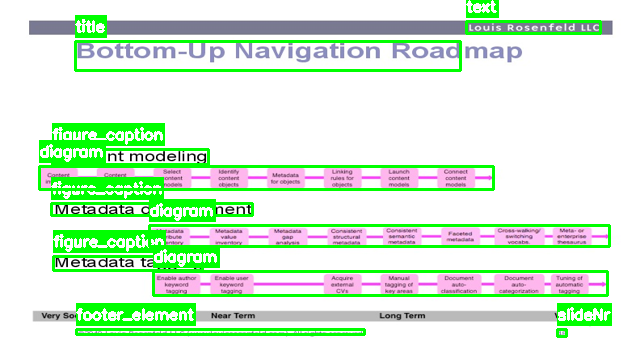}} 
\hspace{0.001\textwidth}
\fbox{\includegraphics[width=0.4\textwidth]{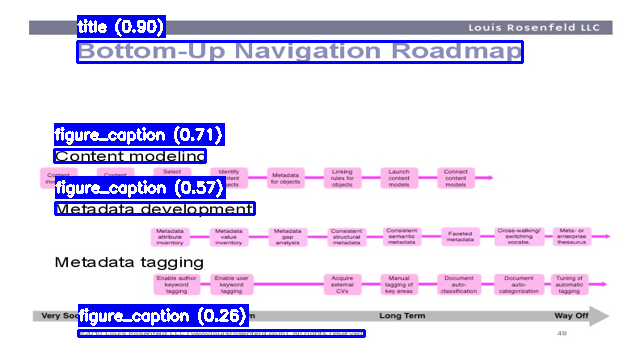}}}
\vspace{0.001\textwidth}
\centerline{ 
\fbox{\includegraphics[width=0.4\textwidth]{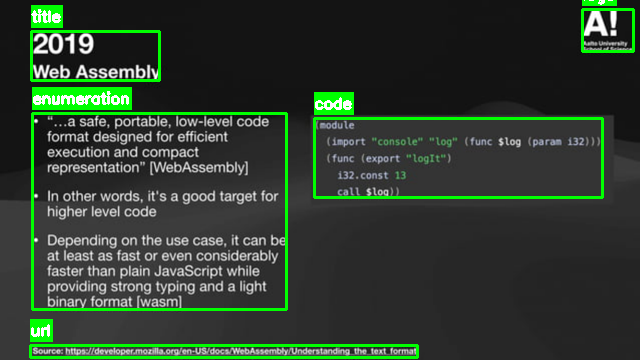}}
\hspace{0.001\textwidth}
\fbox{\includegraphics[width=0.4\textwidth]{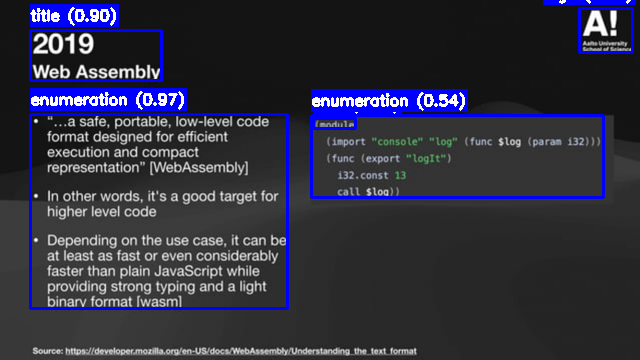}}}
\vspace{0.001\textwidth}
\centerline{
\fbox{\includegraphics[width=0.4\textwidth]{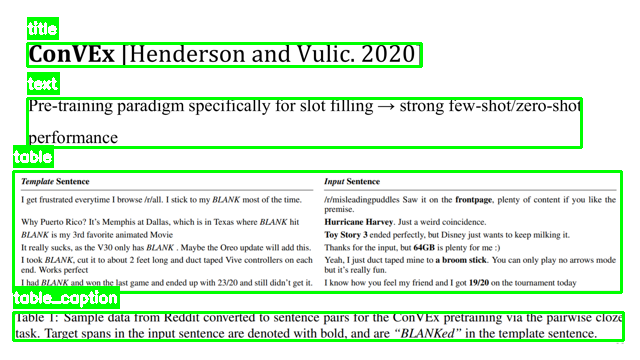}} 
\hspace{0.001\textwidth} 
\fbox{\includegraphics[width=0.4\textwidth]{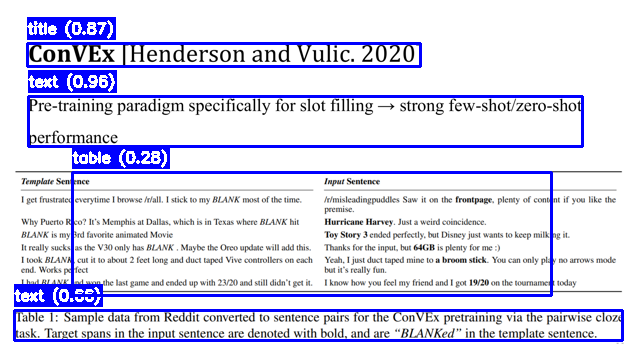}}}
\vspace{0.001\textwidth}
\centerline{
\fbox{\includegraphics[width=0.4\textwidth]{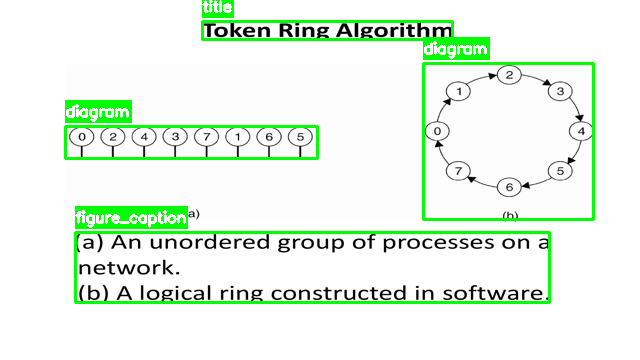}} 
\hspace{0.001\textwidth}  
\fbox{\includegraphics[width=0.4\textwidth]{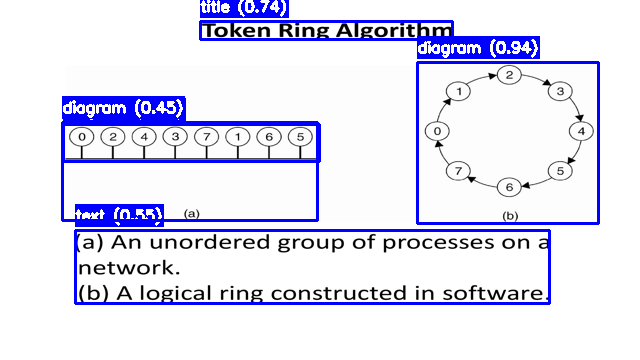}}} 
\caption{Illustration of further selected visual results from YOLOv9 (Two-Stage), where green denotes ground truth bounding boxes and blue indicates predicted bounding boxes.(Later results highlight missed detections and class confusion)}
\label{fig:det_visual_results}
\end{figure}
\begin{figure}[!t]
\centerline{
\includegraphics[width=1\textwidth, height=1\textheight, keepaspectratio]{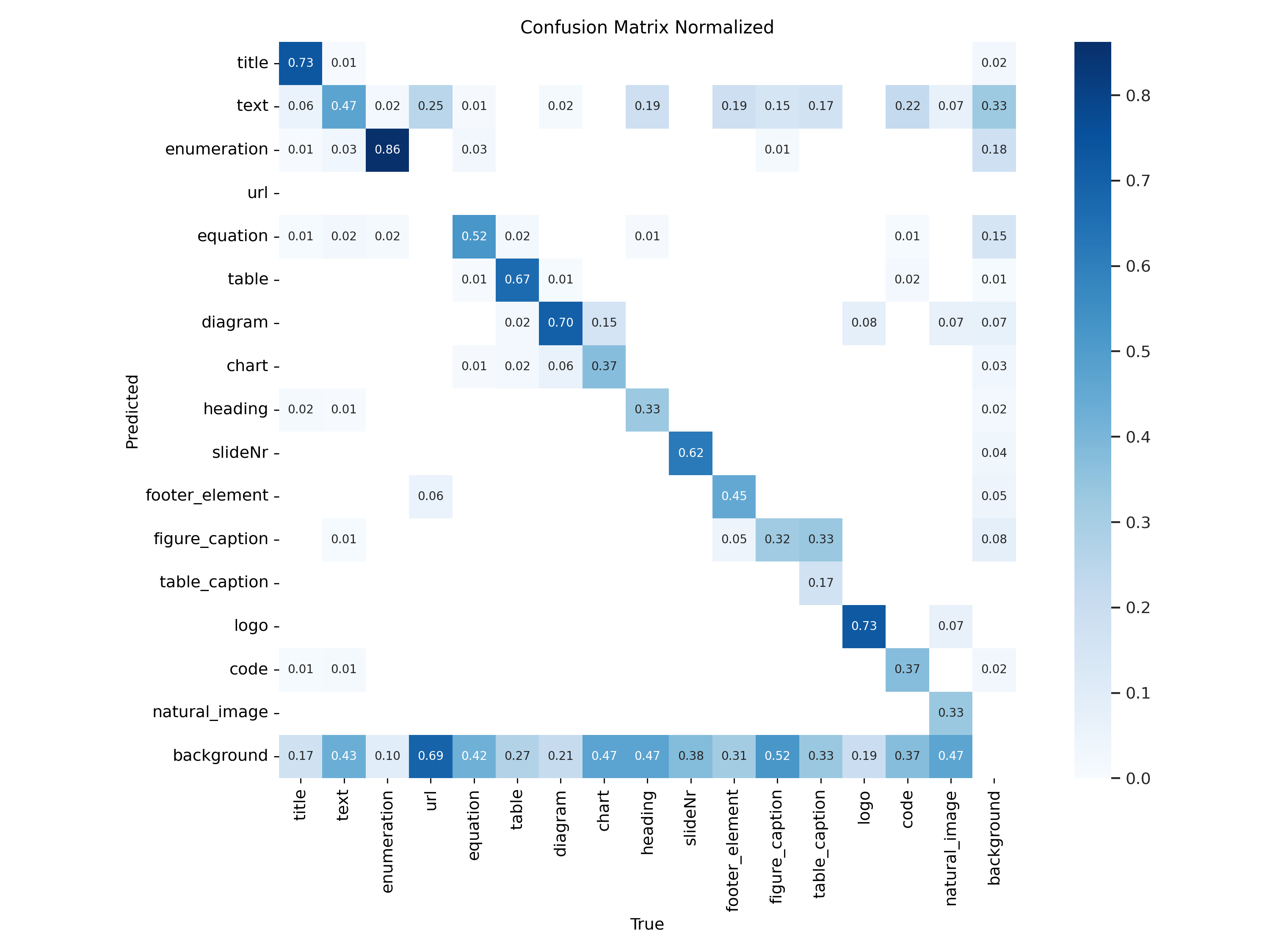}} 
\caption{Confusion Matrix for predictions of YOLOV9 Two stage finetuned model (SynDet+RealSlide Train) on RealSlide Test.}
\label{fig:sample_figure}
\end{figure}

\begin{figure}[h]
    \centering

    \textbf{Query:} Slide with a title and diagram about DynaEval.  
    \\
    \includegraphics[width=0.24\textwidth]{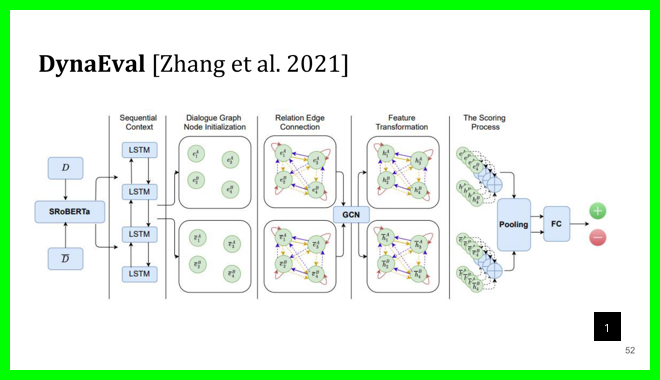}
    \includegraphics[width=0.24\textwidth]{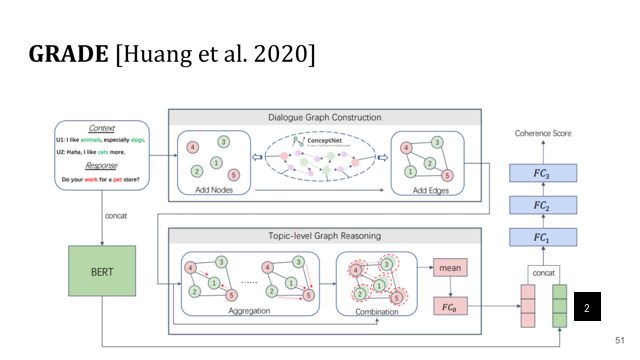}
    \includegraphics[width=0.24\textwidth]{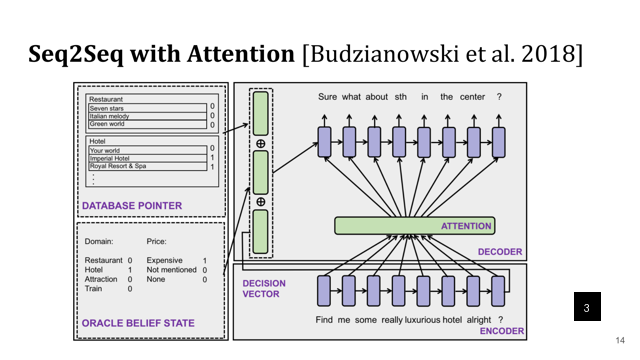}
    \includegraphics[width=0.24\textwidth]{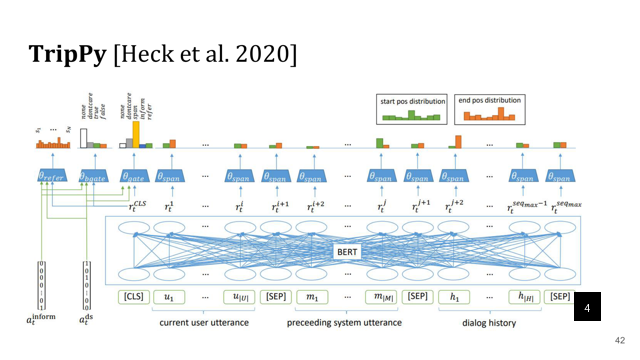}

    \hrule
    \vspace{10pt} 

    \textbf{Query:} Concurrency Control Techniques explained using text, enumeration, flow-chart.  
    \\
    \includegraphics[width=0.24\textwidth]{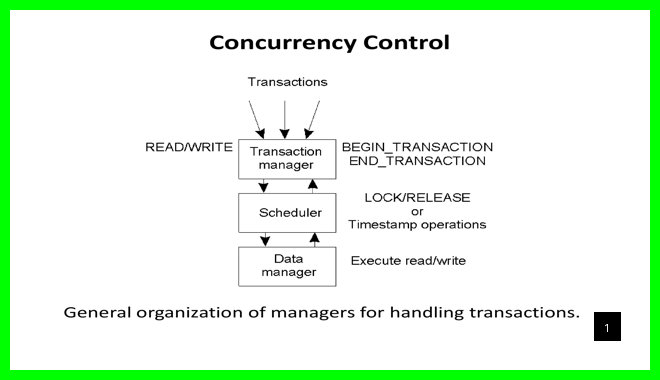}
    \includegraphics[width=0.24\textwidth]{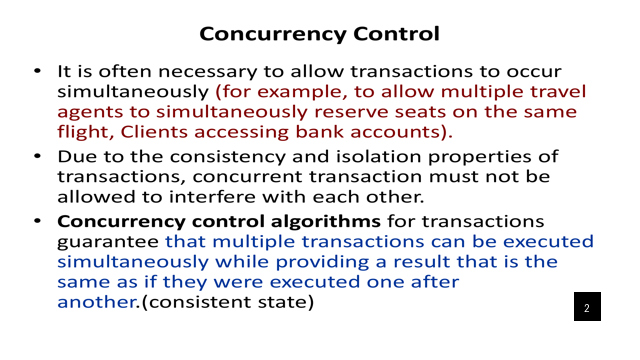}
    \includegraphics[width=0.24\textwidth]{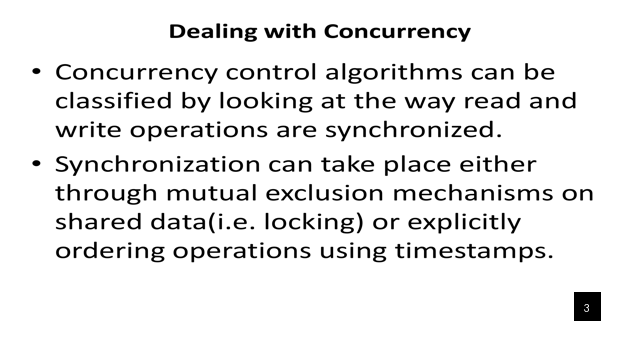}
    \includegraphics[width=0.24\textwidth]{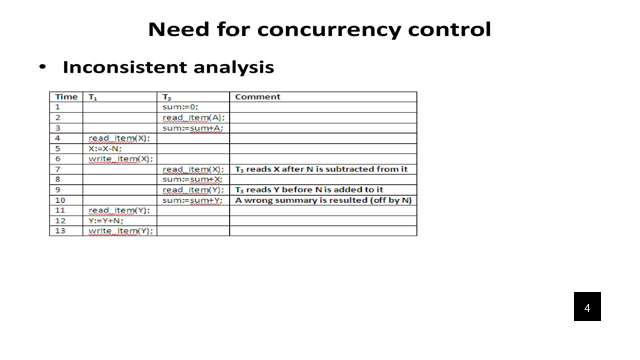}

    \hrule
    \vspace{10pt}

    \textbf{Query:} Slide on steps to run React application.  
    \\
    \includegraphics[width=0.24\textwidth]{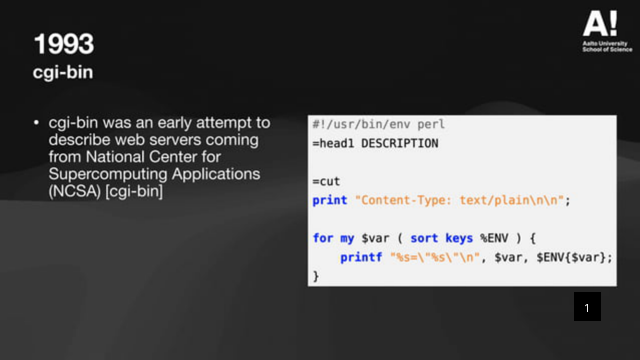}
    \includegraphics[width=0.24\textwidth]{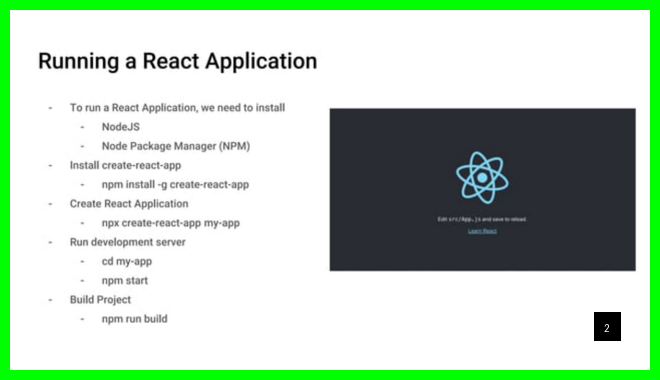}
    \includegraphics[width=0.24\textwidth]{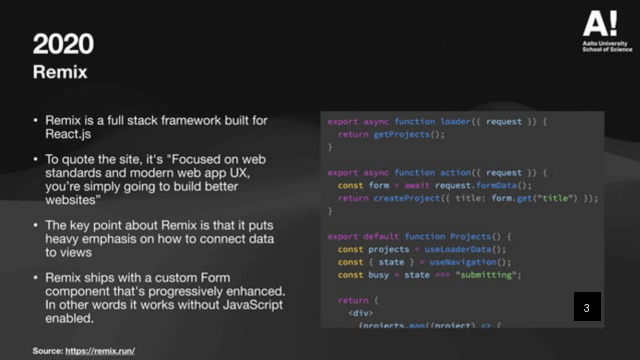}
    \includegraphics[width=0.24\textwidth]{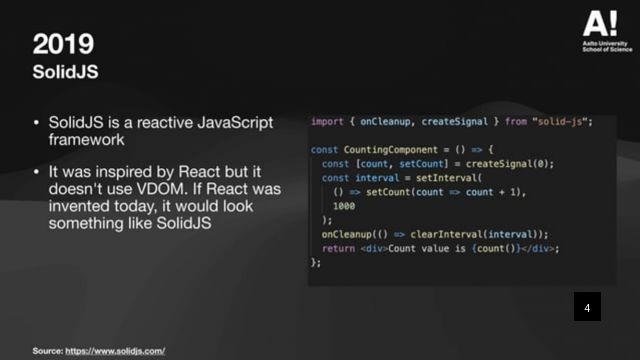}

    \hrule
    \vspace{10pt}

    \textbf{Query:} Lecture slide on k-means clustering and Expectation Maximization algorithm.  
    \\
    \includegraphics[width=0.24\textwidth]{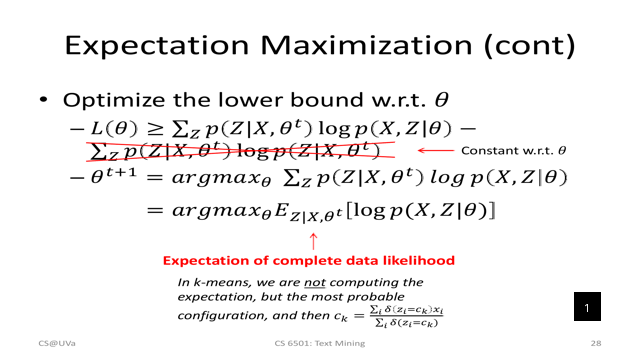}
    \includegraphics[width=0.24\textwidth]{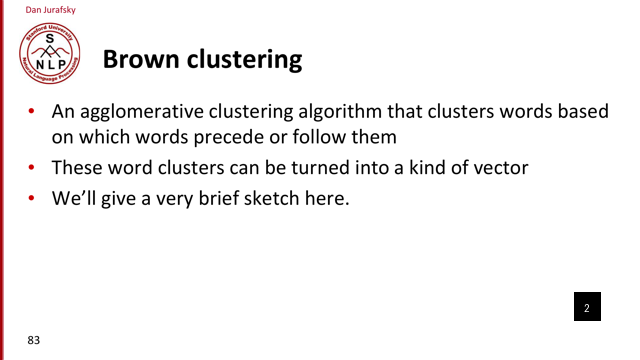}
    \includegraphics[width=0.24\textwidth]{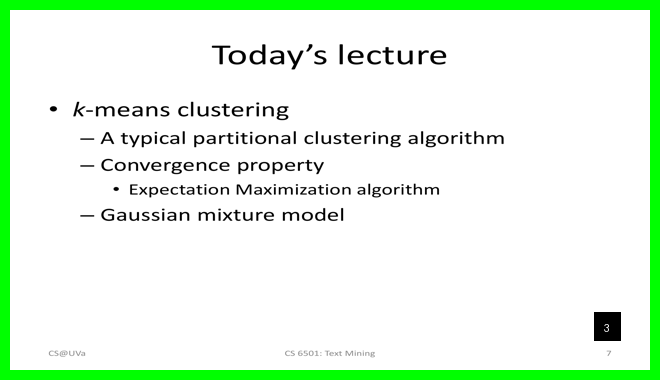}
    \includegraphics[width=0.24\textwidth]{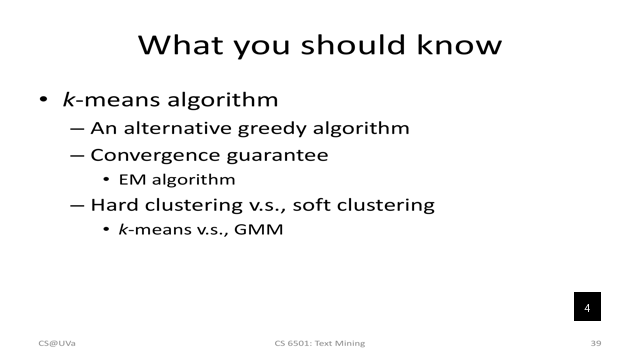}

    \hrule
    \vspace{10pt}

    \caption{Visual Analysis of result of CLIP model finetuned using SynRet data}

\end{figure}
\end{document}